\AddToHook{package/hyperref/before}{\RequirePackage{float}}
\documentclass[pdflatex,sn-nature]{sn-jnl}%

\usepackage{graphicx}%
\usepackage{multirow}%
\usepackage{amsmath,amssymb,amsfonts}%
\usepackage{mathrsfs}%
\usepackage[title]{appendix}%
\usepackage{xcolor}%
\usepackage{textcomp}%
\usepackage{manyfoot}%
\usepackage{booktabs}%
\usepackage{algorithm}%
\usepackage{algorithmic}
\usepackage{listings}%
\usepackage{verbatim}

\definecolor{mygreen}{rgb}{0,0.6,0}
\definecolor{mygray}{rgb}{0.5,0.5,0.5}
\definecolor{mymauve}{rgb}{0.58,0,0.82}

\usepackage{placeins}
\usepackage{array}
\usepackage{longtable}
\usepackage{pdflscape}
\providecommand{\MouseGPTFigureDir}{figures_final/PDF}
\providecommand{\theHfigure}{\arabic{figure}}

\begin{document}

\title[MouseGPT]{MouseGPT: A Large-scale Vision-Language Model for Mouse Behavior Analysis}

\author[1,2]{\fnm{Teng} \sur{Xu}}\email{xt@shanghaitech.edu.cn}
\equalcont{These authors contributed equally to this work.}

\author[1,2]{\fnm{Taotao} \sur{Zhou}}\email{zhoutt2023@shanghaitech.edu.cn}
\equalcont{These authors contributed equally to this work.}

\author[1,2]{\fnm{Youjia} \sur{Wang}}\email{wangyj2@shanghaitech.edu.cn}
\equalcont{These authors contributed equally to this work.}

\author[3]{\fnm{Peng} \sur{Yang}}\email{yp@shanghaitech.edu.cn}
\equalcont{These authors contributed equally to this work.}

\author[3]{\fnm{Simin} \sur{Tang}}\email{tangsm2023@shanghaitech.edu.cn}

\author[1]{\fnm{Kuixiang} \sur{Shao}}\email{shaokx@shanghaitech.edu.cn}

\author[3]{\fnm{Zifeng} \sur{Tang}}\email{tangzf@shanghaitech.edu.cn}

\author[1]{\fnm{Yifei} \sur{Liu}}\email{arnoliu@shanghaitech.edu.cn}

\author[3]{\fnm{Xinyuan} \sur{Chen}}\email{chenxy7@alumni.shanghaitech.edu.cn}

\author[1]{\fnm{Yihan} \sur{Liu}}\email{liuyh2023@shanghaitech.edu.cn}
\author[1]{\fnm{Zhengning} \sur{Xu}}\email{xuzhn2024@shanghaitech.edu.cn}
\author[1]{\fnm{Boyang} \sur{Xia}}\email{xiaby2023@shanghaitech.edu.cn}

\author[4]{\fnm{Hongshuang} \sur{Wang}}\email{hongshuang.wang@ciac.ac.cn}

\author[5]{\fnm{Xiaohui} \sur{Wang}}\email{xiaohui.wang@ciac.ac.cn}

\author[3]{\fnm{Huoqing} \sur{Luo}}\email{luohq@shanghaitech.edu.cn}

\author[1]{\fnm{Jingya} \sur{Wang}}\email{wangjingya@shanghaitech.edu.cn}

\author*[3]{\fnm{Ji} \sur{Hu}}\email{huji@shanghaitech.edu.cn}

\author*[1]{\fnm{Jingyi} \sur{Yu}}\email{yujingyi@shanghaitech.edu.cn}

\affil*[1]{\orgdiv{School of Information Science and Technology}, \orgname{ShanghaiTech University}, \orgaddress{\postcode{201210}, \state{Shanghai}, \country{China}}}

\affil[2]{\orgname{LumiAni Technology}, \orgaddress{\postcode{201210}, \state{Shanghai}, \country{China}}}

\affil[3]{\orgdiv{School of Life Science and Technology}, \orgname{ShanghaiTech University}, \orgaddress{\postcode{201210}, \state{Shanghai}, \country{China}}}

\affil[4]{\orgdiv{Laboratory of Chemical Biology}, \orgdiv{Changchun Institute of Applied Chemistry}, \orgname{Chinese Academy of Sciences}, \orgaddress{\city{Changchun}, \postcode{130022}, \state{Jilin}, \country{China}}}

\affil[5]{\orgdiv{School of Applied Chemistry and Engineering}, \orgname{University of Science and Technology of China}, \orgaddress{\city{Hefei}, \postcode{230026}, \state{Anhui}, \country{China}}}

\abstract{Analyzing animal behavior is crucial in neuroscience, yet deciphering its intricate dynamics remains challenging. Traditional machine vision approaches detect spontaneous behaviors but are constrained by limited interpretability and reliance on manual labeling, restricting exploration of the full behavioral spectrum. Here, we introduce MouseGPT, an integrated vision-language model and analytical framework that translates raw video of animal activity into natural language to automate open-vocabulary behavior analysis. Built upon a comprehensive dataset of over 42 million frames spanning diverse psychiatric conditions, MouseGPT maps multi-view or single-camera recordings directly into computable semantic representations. An agentic framework leverages these representations for detailed behavior profiling, cross-view stable clustering, and zero-shot novel behavior discovery without requiring labor-intensive annotation. Surpassing conventional kinematics-based pipelines and generalist vision-language models in behavioral precision and adaptability, MouseGPT establishes continuous semantic representations as a powerful tool for deciphering complex ethological dynamics in animal models.}


\maketitle

\section{Introduction}\label{sec1}
Animal behavior studies have long been a cornerstone in systems neuroscience~\cite{JohnNeuron2017,Calhoun2019, Datta2019} and psychiatric research~\cite{Eric2010}. They provide a controlled, manipulable framework for modeling human neurological and psychiatric conditions in animals, enabling the investigation of neural and molecular mechanisms that may offer insights applicable to humans. Through genetic, chemical, and environmental manipulations, scientists can probe causative relationships, identify potential targets for psychiatric therapies, and generate valuable translational insights for drug testing. For example, the classic Tail Suspension Test (TST)~\cite{Hiroshi2022,Slattery2012} and Forced Swim Test (FST)~\cite{Slattery2012} help quantify behaviors in rodents~\cite{Anand2019} suggestive of depressive states by measuring immobility. Although simple and reproducible, such tests are short-term and do not reflect the complexity and chronic nature of depression. In reality, the behaviors of rodents are often subtle, fluctuating, and interrelated, spanning locomotion, grooming, social interactions, exploratory behavior, and responses to novel stimuli. Each of these behaviors can provide unique insights critical for modeling complex experimental conditions. 

Capturing these behaviors across diverse experimental conditions typically relies on video recordings. In many settings, however, behavioral scoring still depends on human observers who watch long recordings and manually annotate selected behaviors to derive statistical measurements~\cite{JohnF2005}. This process is labor-intensive, prone to fatigue, bias, and inconsistency, and becomes especially challenging in advanced scenarios such as free-moving or socially interacting mice. The immense volume and complexity of behavioral data further exacerbate these difficulties, often leading to selective recording that overlooks subtle but critical behaviors~\cite{Alex2014}. These limitations, compounded by subjective interpretation and the difficulty of documenting behavior consistently in real time, significantly undermine the accuracy and reliability of collected data~\cite{LindsayP2021,Linda2019,KSCHMACK2021}.

Advancements in artificial intelligence have driven the development of automatic tools that mitigate the challenges of manual observation, with the goal of making behavior analysis more efficient and scalable. Supervised learning algorithms~\cite{Fabrice2019, Miwa2024} have shown success in automating the identification of specific, predefined behaviors by using expert-annotated datasets, but they remain limited to behaviors represented in the training data and depend on substantial manual labeling. Unsupervised approaches offer an important alternative by discovering behavioral structure without predefined labels, using features such as 3D keypoint positions~\cite{YaNing2024, Tanmay2019, Lauer2022, A-SOiD, B-SOiD, KangHuang2021}, movement speed~\cite{kabra2013jaaba, Luxem2022,Pereira2019}, and joint angles~\cite{B-SOiD, A-SOiD}. However, these methods are fundamentally grounded in kinematic similarity~\cite{YaNing2024, Gosztolai2021, DunnDANNCE2021, Weinreb2024}: they group behaviors by geometric and temporal resemblance rather than by ethological meaning. As a result, the resulting clusters may remain semantically heterogeneous, subtle or rare behaviors may be merged or overlooked~\cite{Pereira2022}, and extensive post-hoc human review is often still required to assign biological relevance~\cite{Alexander2020,Pereira2020}. To fully capture the complexity of mouse behavior, it is therefore essential to establish a representational space that links raw visual observations directly to human-interpretable ethological concepts.

Vision-Language Models (VLMs) provide a viable path to deliver the ``universal feature'' required to construct this representational space. By intertwining visual and textual data, VLMs offer a richer, more nuanced understanding of complex multi-modal tasks and have already proven effective across diverse research domains~\cite{Christensen2024, Cui2024, Peng2024, chu20245, lu2024multimodal, shen2024unbiased}. Within computational ethology, the pioneering framework AmadeusGPT~\cite{ye2023amadeusGPT} introduced an agentic paradigm to cohesively orchestrate existing computational modules, providing the community with the crucial ability to act upon behavioral data. Here, we introduce MouseGPT, a foundational VLM for open-vocabulary interpretation of mouse behavior that supports quantitative behavioral analysis. The dataset, fine-tuned models and native kinematic pipeline were developed primarily for individual mice. By mapping visual cues, such as postures or movements, into a continuous semantic feature space via natural language, MouseGPT interprets animal behaviors beyond the limitations of predefined categories. Trained on large and diverse datasets, MouseGPT supports open-vocabulary descriptions of subtle, compound, and previously unseen behaviors, providing context-rich annotations that enrich behavioral analysis.

MouseGPT advances mouse behavior research through three key contributions. First, it enables a comprehensive interpretation of mouse behaviors by integrating visual and textual information to generate detailed, context-sensitive descriptions. This approach captures behavioral nuances that traditional methods often overlook, providing a more robust and automated solution for studying complex behaviors. We also established an expert-verified benchmark with automated rubric-based grading to compare behavioral descriptions across general-purpose VLMs. Second, we deliver a first-of-its-kind open-vocabulary dataset of over 42 million frames of multi-view video recordings, covering mice under various psychiatric conditions, including depression, hallucination, and schizophrenia. This multi-view setup was used to construct a rich training resource and 3D reference; applying MouseGPT to new data, however, is not restricted to that acquisition setting and can extend to standard laboratory videos, including single-camera recordings, while seamlessly integrating with community-standard pose estimators such as DeepLabCut~\cite{Lauer2022} and SLEAP~\cite{Pereira2022}. Unlike prior datasets that often lack sufficient scale and diversity, our dataset ensures a broad representation of behaviors, enhancing MouseGPT's ability to deliver accurate and insightful behavioral analysis while also serving as a valuable resource for future model development. Finally, as a downstream application to demonstrate the utility of this new semantic modality, we designed an integrated behavioral analysis framework that offers advanced tools for behavioral clustering, classification, and statistical assessment. By mapping textual descriptions to high-dimensional feature embeddings, this framework allows researchers to quantify and explore mouse behaviors with precision, accommodating both high-level behavioral trends and fine-grained analysis. Together, these components position MouseGPT as a transformative tool, providing researchers with a flexible, content-rich approach to deciphering the complexities of mouse behavior and facilitating novel discoveries in neuropsychiatric research.

To demonstrate the capabilities of MouseGPT, we evaluated its performance on several datasets, including public single-camera benchmarks and datasets focused on hallucination and other psychiatric conditions in mice. We applied MouseGPT to these datasets to perform a range of functions such as behavior profiling, clustering, fine-grained behavior analysis, and new behavior discovery. These tasks highlight MouseGPT’s ability to interpret and categorize complex behaviors even in challenging scenarios involving nuanced behavioral patterns. For instance, MouseGPT effectively distinguished between different degrees of depression in mice, identified characteristic behaviors associated with hallucination, and provided valuable insights into the effects of various psychiatric conditions.

In our evaluations, MouseGPT was compared with other open-source and proprietary vision-language models, including GPT-4o, InternVL2~\cite{chen2023internvl}, and MiniCPM-2.6~\cite{yao2024minicpm}. Across expert assessment and downstream analyses, MouseGPT outperformed these models in both accuracy and depth of behavioral understanding, particularly in its ability to generalize across different contexts and generate contextually rich descriptions. These findings underscore the robustness and versatility of MouseGPT as an essential tool for ethological research.

\section{Results}\label{sec2}

\subsection{Overview of MouseGPT: A Behavior Understanding Model and An Analytical Framework}

We present MouseGPT, a foundational vision-language model and analytical framework that converts mouse video into open-vocabulary, ethologically interpretable descriptions and quantitatively analyzable behavioral representations. MouseGPT processes visual observations from standard laboratory videos, ranging from single-camera recordings to synchronized multi-view videos, alongside optional pose-derived kinematic information (Fig.~\ref{fig:fig1}). It interprets mouse behavior using an open-vocabulary approach to capture both common and subtle actions, including previously undefined behavioral patterns. By representing behavior through natural-language descriptions and corresponding semantic features (Methods), MouseGPT enables the interpretation and analysis of complex behaviors across diverse experimental conditions.

\begin{figure}[!htbp]
\setcounter{figure}{0}
\renewcommand{\thefigure}{\arabic{figure}}
\renewcommand{\theHfigure}{main.\arabic{figure}}
\renewcommand{\figurename}{Fig.}
\centering
\includegraphics[width=\textwidth,height=0.8\textheight,keepaspectratio]{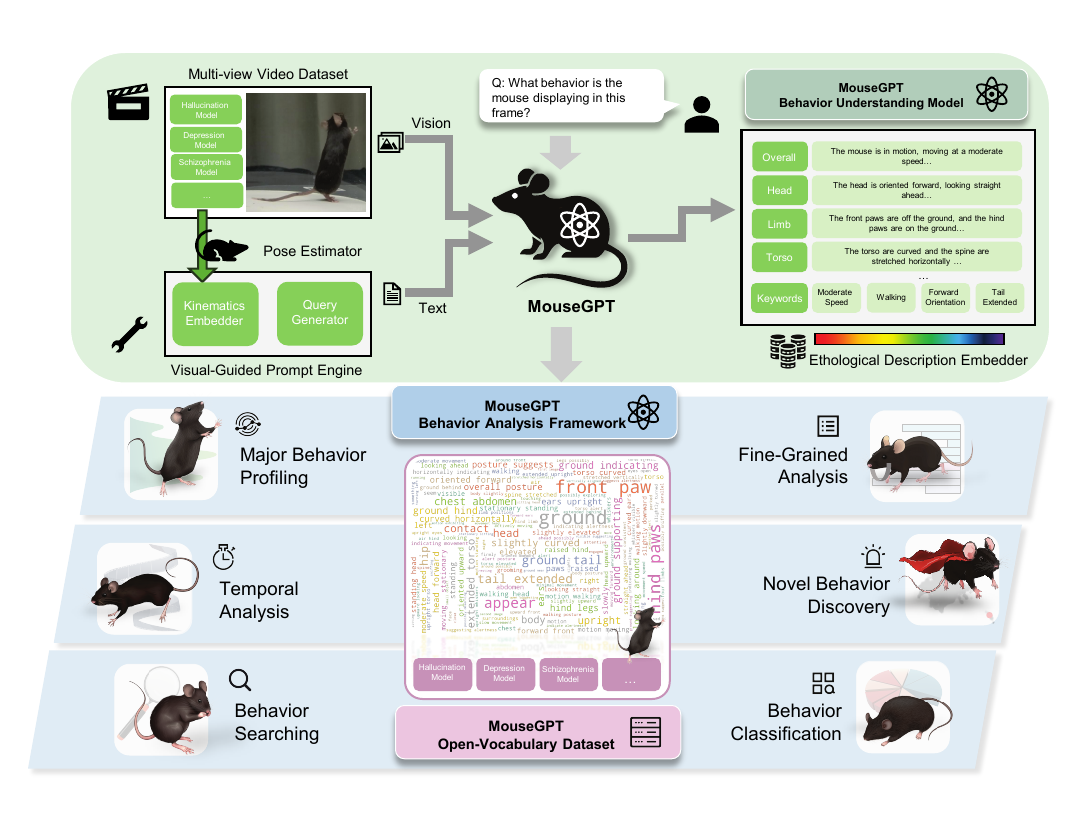}
\caption{\textbf{MouseGPT, a comprehensive mouse behavior understanding model and analysis framework.} By integrating single-camera or multi-view video data, pose estimation, and kinematic embeddings, MouseGPT generates open-vocabulary ethological descriptions of mouse behaviors. It leverages a visual-guided prompt engine and a pioneering open-vocabulary behavioral dataset, featuring disease-specific behavior phenotypes (e.g., hallucination, depression, schizophrenia). The system supports behavior classification, temporal analysis, fine-grained behavior profiling, and novel behavior discovery, delivering transformative capabilities for studying and analyzing complex animal behaviors.}
\label{fig:fig1}
\end{figure}

The MouseGPT behavior understanding model generates natural-language descriptions from video observations of mouse behavior (Sec.~\ref{sec:totext}). To support this capability, we constructed the MouseGPT dataset, which integrates large-scale multi-view recordings with pose estimations, kinematic information, and open-vocabulary behavior annotations, providing rich supervision for mouse behavior understanding (Sec.~\ref{sec:totext}). When kinematic information is incorporated at inference time, it can be obtained either from our native pipeline or from community-standard pose-estimation tools such as DeepLabCut and SLEAP. This design couples rich visual evidence with semantically structured behavioral descriptions that can be directly used for downstream analysis.

The MouseGPT Behavior Analysis Framework builds on the outputs of the behavior understanding model, providing researchers with a comprehensive suite of tools for detailed behavior profiling and analysis (Sec.~\ref{sec:tobehavior}). By embedding generated descriptions in a high-dimensional semantic space (Methods), MouseGPT captures fine-grained relationships among behaviors beyond fixed label categories. These representations support functionalities such as major behavior profiling, temporal analysis, fine-grained behavior analysis, behavior searching, classification, and novel behavior discovery (Fig.~\ref{fig:fig1}).
To demonstrate the utility of the platform in neuroscience applications, we applied MouseGPT to datasets involving psychoactive substances and related neuropsychiatric conditions (Sec.~\ref{sec:drugs}). This application setting illustrates how open-vocabulary behavioral representations can be used to summarize continuous activity, compare behavioral structure across conditions, and examine sub-patterns within broad behavioral categories at higher resolution.

To enhance usability, we developed a Natural Language User Interface (NLUI) that allows researchers to interact with MouseGPT through natural-language queries (Sec.~\ref{sec:nlui}). This interface supports tasks such as retrieving ethogram reports, behavior searching, and behavior classification, reducing the need for extensive reprogramming and making MouseGPT more accessible for routine behavioral analysis.

\subsection{MouseGPT Enhances Mouse Behavior Understanding through Vision-Language Integration}
\label{sec:totext}
MouseGPT addresses the challenge of comprehensive mouse behavior analysis by transforming visual signals from standard laboratory videos, including single-camera and synchronized multi-view recordings, into rich textual descriptions of mouse behaviors. To achieve this, we developed and curated a comprehensive dataset spanning diverse behavioral scenarios, paired with pose-derived kinematic statistics and open-vocabulary annotations (Extended Fig.~\ref{fig:ef1}), to enable effective supervised fine-tuning (SFT) and ensure robust model generalization.

The dataset was collected using a custom-built 3D video capture system comprising eight synchronized cameras capturing footage at 4K resolution and 60 frames per second (Fig.~\ref{fig:fig2}a-left, Methods). This setup provided precise multi-angle recordings of mouse behaviors, which were then used to reconstruct accurate 3D key-point positions (Fig.~\ref{fig:fig2}a-middle, Methods). These data were critical for capturing naturalistic and stimulus-induced behaviors, including conditions such as depression, hallucination, and schizophrenia, providing a diverse foundation for MouseGPT's training. To further enhance dataset quality, we used GPT-4o to generate candidate open-vocabulary descriptions and employed an automated rating procedure to retain high-quality annotations (Fig.~\ref{fig:fig2}a-right, Methods). Because these GPT-4o-generated descriptions were used as supervisory signals during data construction, we explicitly validated them through blinded expert review. An independent panel of 15 senior researchers in animal behavior judged them to be factually correct in 78.6\% of cases on average (Fig.~\ref{fig:fig2}c, Methods), supporting the reliability of the training annotations.

To train and infer with MouseGPT, we employed a kinematics-aware prompt-engineering strategy that augments visual inputs with pose-derived statistics (Fig.~\ref{fig:fig2}b-left, Methods). These enriched prompts were used during supervised fine-tuning of the MouseGPT VLM, allowing visual observations to be interpreted together with complementary kinematic context. We trained model variants optimized for either high-accuracy applications (MouseGPT-Large, Methods) or computationally efficient deployment (MouseGPT-Lite, Methods). 

The trained model outputs natural-language descriptions of behaviors (Fig.~\ref{fig:fig2}b-right; Supplementary Fig.~4), which are then vectorized into high-dimensional embeddings~\cite{openai2024embedding} for downstream analysis (Sec.~\ref{sec:tobehavior}). To visualize the differences in behavior distribution across experimental groups, we used Uniform Manifold Approximation and Projection (UMAP)~\cite{mcinnes2018umap} on MouseGPT embeddings (Fig.~\ref{fig:fig2}e, Methods). The resulting semantic space exhibits clear organization across behavioral conditions, illustrating the model's ability to distinguish behaviors between different experimental groups.

\begin{figure}[!htbp]
\setcounter{figure}{1}
\renewcommand{\thefigure}{\arabic{figure}}
\renewcommand{\theHfigure}{main.\arabic{figure}}
\renewcommand{\figurename}{Fig.}
\centering
\includegraphics[width=\textwidth,height=0.8\textheight,keepaspectratio]{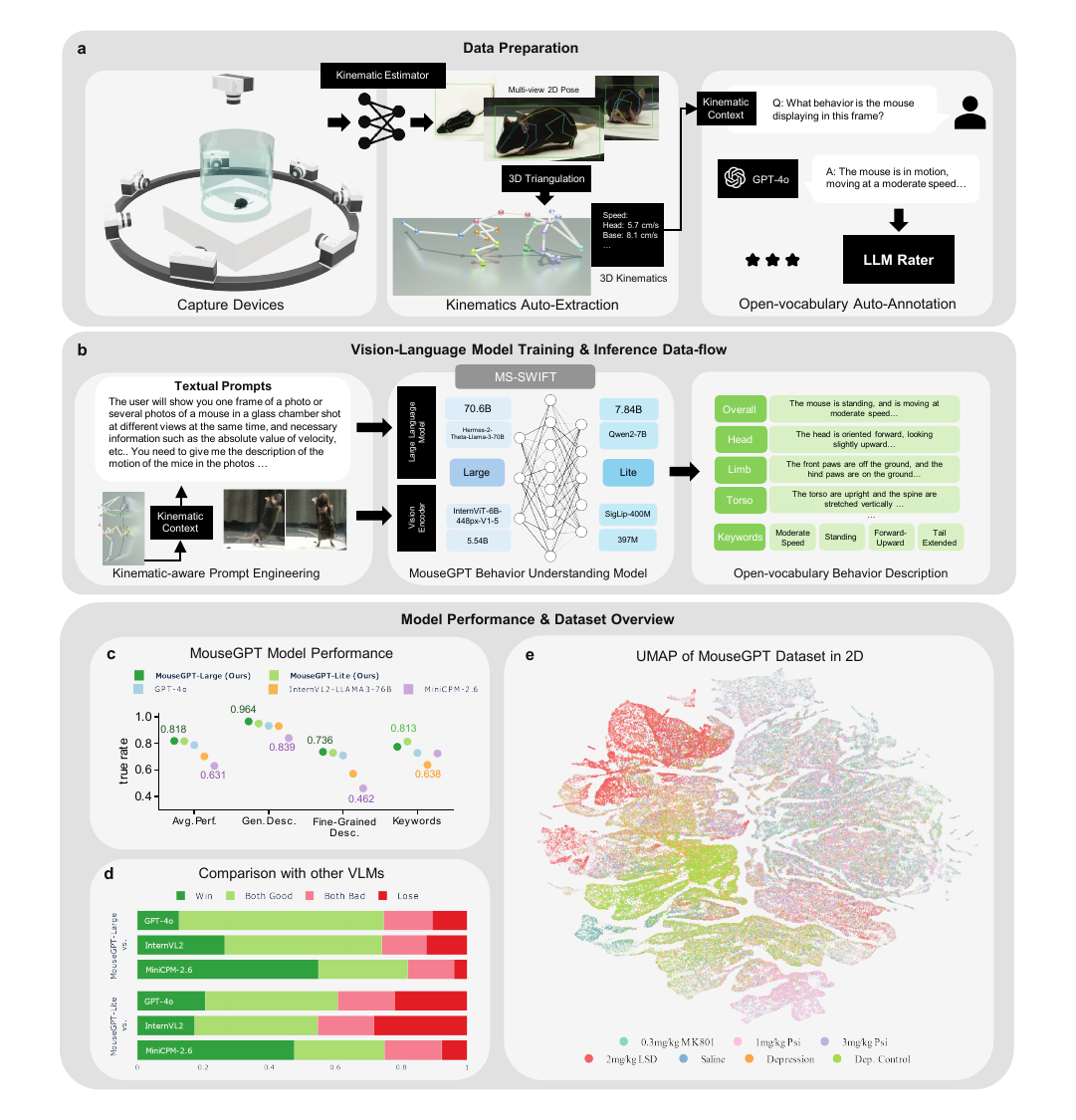}
\caption{\textbf{MouseGPT Behavior Understanding Model.}
\textbf{a}, Data Preparation: A multi-view 3D capture system generates keypoints and limb velocities for each frame, with training data produced by GPT-4o and refined through automated selection processes.
\textbf{b}, Vision-Language Model Training and Inference Workflow: MouseGPT-Large is optimized for detailed behavior analysis, while MouseGPT-Lite provides a lightweight alternative for streamlined tasks.
\textbf{c}, Performance Evaluation: A blind panel of 15 independent senior researchers evaluated model outputs for factual correctness. The assessment confirms the rigorous quality of the GPT-4o training annotations (78.6\% baseline true rate) and highlights MouseGPT's superior performance against both open-source models of comparable size and GPT-4o.
\textbf{d}, Comparison with Other VLMs: Experts performed blinded, paired comparisons of descriptions generated by MouseGPT and other vision-language models for the same input. The plot reports the proportions of judgments in four categories: Win, Both Good, Both Bad and Lose.
\textbf{e}, Dataset Visualization: UMAP projection of the MouseGPT dataset in 2D, with color-coded clusters representing distinct drug treatments. }
\label{fig:fig2}
\end{figure} 

In addition to evaluating vision-language models, we examined the contribution of input information and the compatibility of its sources. Kinematic estimates from community-standard pose-estimation tools, such as DeepLabCut and SLEAP, closely matched those from our native estimator, and the resulting behavioral descriptions remained semantically consistent (Extended Fig.~\ref{fig:ef2}a--c, Methods). A formal ablation study further showed that the kinematic prompt provides a practical grounding signal for behavior interpretation, improving the model's ability to resolve speed- and action-related ambiguities relative to vision-only prompting (Extended Fig.~\ref{fig:ef2}d,e, Methods).

We benchmarked MouseGPT against other open-source and proprietary models, including InternVL2~\cite{chen2023internvl}, MiniCPM-2.6~\cite{yao2024minicpm}, and GPT-4o~\cite{openai2024gpt4o}, and found that MouseGPT demonstrated superior performance in interpreting subtle and novel behaviors, as validated by blinded domain experts (Fig.~\ref{fig:fig2}c,d). MouseGPT-Large achieved the highest accuracy for detailed behavior interpretation, while MouseGPT-Lite provided an efficient solution for computationally constrained environments.  On an expert-verified benchmark of 435 frames evaluated across eight general-purpose VLMs and 14 settings, the mean MouseGPT Benchmark Index ranged from 50.44 for GPT-4o to 89.96 for GPT-6 Astra at maximum requested reasoning effort (Extended Fig.~\ref{fig:ef3}; Methods), illustrating the potential to improve description quality by updating the framework's external generation model. These evaluations highlight MouseGPT’s precision, versatility, and practical generalizability as a framework for mouse behavior research.

\subsection{Advanced Behavior Analysis Framework of MouseGPT}
\label{sec:tobehavior}

MouseGPT's advanced framework begins by converting natural-language descriptions of mouse behaviors into computable semantic representations (Fig.~\ref{fig:fig3}a), enabling a broad range of analytical techniques. By leveraging these open-vocabulary descriptions, MouseGPT offers a versatile approach to comprehensively analyze mouse behaviors, facilitating clustering of behavioral patterns (Fig.~\ref{fig:fig3}b), analysis of fine-grained behaviors (Fig.~\ref{fig:fig3}c), identification of novel activities (Fig.~\ref{fig:fig3}d), and prediction of behavioral phenotypes (Fig.~\ref{fig:fig3}e). This transformation from descriptive language to quantitative analysis allows researchers to explore behavioral data with greater flexibility and interpretability than traditional label-based or purely kinematics-based pipelines.

\begin{figure}[!htbp]
\setcounter{figure}{2}
\renewcommand{\thefigure}{\arabic{figure}}
\renewcommand{\theHfigure}{main.\arabic{figure}}
\renewcommand{\figurename}{Fig.}
\centering
\includegraphics[width=\textwidth,height=0.8\textheight,keepaspectratio]{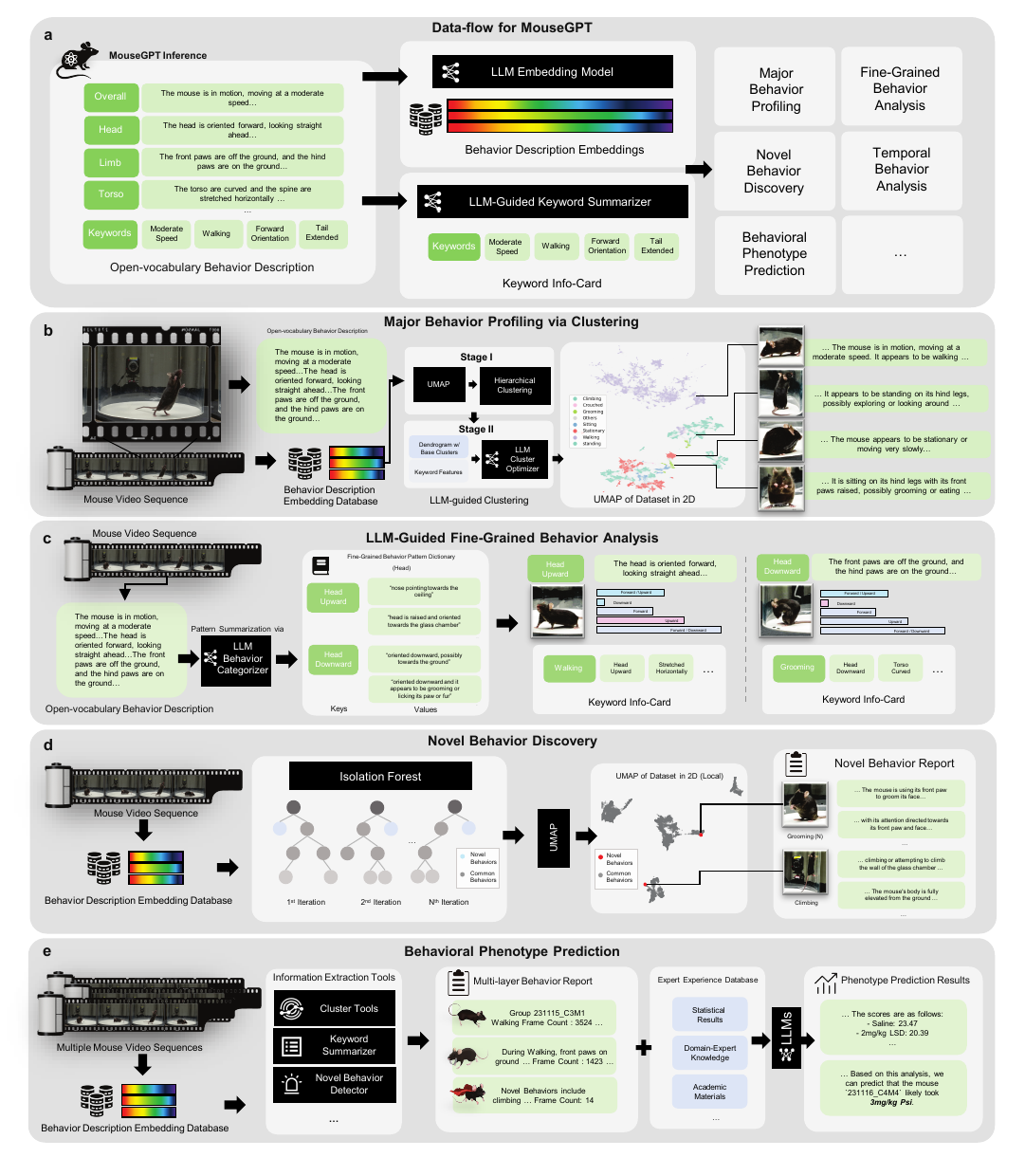}
\caption{\textbf{MouseGPT Behavior Analysis Framework.}
\textbf{a}, Data Flow: MouseGPT converts open-vocabulary behavior descriptions into two computable features: high-dimensional embeddings for semantic analysis and keyword-based summaries for interpretability, powering various downstream applications.
\textbf{b}, Major Behavior Profiling: A two-stage LLM-enhanced clustering groups mouse actions into distinct behavioral categories, with colors representing clusters, enabling robust profiling.
\textbf{c}, Fine-Grained Analysis: MouseGPT extracts and quantifies detailed behavior patterns via keyword analysis and frequency calculations, offering precise insights.
\textbf{d}, Novel Behavior Discovery: Using Isolation Forest in embedding space, MouseGPT identifies anomalous behaviors, visualized in 2D UMAP to highlight rare actions.
\textbf{e}, Phenotype Prediction: MouseGPT combines behavioral analysis with expert knowledge to predict drug treatments, aiding in phenotype understanding and experimental outcomes.}
\label{fig:fig3}
\end{figure}

The major behavior profiling capability of MouseGPT enables researchers to group similar behaviors based on their semantic descriptions through a two-stage LLM-guided clustering process (Fig.~\ref{fig:fig3}b). In the first stage, behaviors are clustered using Uniform Manifold Approximation and Projection (UMAP) and hierarchical clustering to create base clusters (Methods). In the second stage, the LLM refines these clusters using additional contextual features, optimizing the clustering outcome (Methods). This method allows behaviors to be grouped in an open-vocabulary manner, providing an intuitive overview of different behavior categories. For instance, applying this analysis to mice with depression-like conditions reveals distinct clusters such as ``grooming'', ``walking'', or ``rearing'', each representing characteristic behaviors of different experimental groups. In a direct comparison with the kinematics-driven syllable discovery approach Keypoint-MoSeq (Extended Fig.~\ref{fig:ef4}a--c, Methods), MouseGPT's semantic grouping yielded behavior categories that were directly interpretable in ethological terms. Notably, MouseGPT revealed substantial semantic heterogeneity within individual Keypoint-MoSeq syllables (Extended Fig.~\ref{fig:ef4}b,c) and demonstrated significantly higher cross-view stability, mitigating the view-dependence commonly observed in purely kinematics-based representations (Extended Fig.~\ref{fig:ef4}d--h).

MouseGPT’s framework offers fine-grained behavior analysis (Fig.~\ref{fig:fig3}c), focusing on body-part-level behaviors described in natural language, which provides additional detail by capturing specific actions and orientations of individual body parts. Using an LLM-guided categorization approach, MouseGPT constructs a pattern dictionary (Methods) that captures detailed behavioral characteristics at the body-part level, such as different head orientations (e.g., ``head forward'' or ``head downward'') or limb states (e.g., ``forelimb in the air'' or ``forelimb on the ground''). Each behavior is summarized with precise descriptions, which are then compiled into statistically analyzable keyword info-cards (Methods), offering a concise overview of the features and frequencies of these fine-grained behaviors. This capability enables researchers to explore detailed and nuanced aspects of mouse activity, making it suitable for in-depth studies of behavior dynamics and their temporal evolution.

MouseGPT also supports the discovery of novel behaviors by identifying outliers within the behavioral embedding space (Methods). Using an Isolation Forest algorithm (Methods), MouseGPT detects uncommon or previously unobserved behaviors across multiple iterations (Fig.~\ref{fig:fig3}d; Supplementary Fig.~5). These potential novel behaviors are then visualized using UMAP, allowing researchers to identify clusters of atypical activities. The identified behaviors are subsequently documented in novel behavior reports (Methods), offering detailed descriptions that help researchers investigate new behavioral phenotypes, particularly in experimental settings involving unknown variables, such as drug treatments or genetic modifications.

MouseGPT also incorporates a behavioral phenotype prediction module that uses insights from various behavior analysis techniques to predict underlying phenotypic traits (Fig.~\ref{fig:fig3}e, Methods). The system aggregates information from clustering results, fine-grained behavior analysis, and novel behavior detection, producing a multi-layer behavior report (Methods). This report is then combined with an expert experience database, which includes statistical findings, domain-expert knowledge, and relevant academic materials, to refine phenotype predictions (Methods). Using a large language model, MouseGPT synthesizes the gathered behavioral data and expert insights to predict potential phenotypes of the subjects. This approach provides researchers with a data-driven mechanism to link observed behaviors to corresponding biological conditions, aiding studies of neuropsychiatric disorders and other complex biological phenomena.

We further evaluated MouseGPT on two public datasets of mouse social behavior, CalMS21 and MABe 2022. Evaluated zero-shot on the public, single-camera CalMS21 benchmark (Extended Fig.~\ref{fig:ef5}a), MouseGPT-Lite achieved high expert-rated description quality (mean score 7.93/10; Extended Fig.~\ref{fig:ef5}b) and demonstrated superior data efficiency in held-out classification compared to standard keypoint features (F1 0.85 versus 0.79; Extended Fig.~\ref{fig:ef5}c, Methods). We adapted our pipeline to the MABe 2022 benchmark, using vision-language models to generate behavioral descriptions while retaining the downstream behavior-analysis pipeline (Extended Fig.~\ref{fig:ef5}d, Methods). The framework achieved average F1 scores of 0.549 with GPT-5.1 and 0.560 with GPT-6 Astra, compared with the published BAMS reference score of 0.196 (Extended Fig.~\ref{fig:ef5}e,f; Supplementary Table~2).

\subsection{MouseGPT Differentiates Behavioral Patterns in Mice Under Various Psychoactive Drugs}
\label{sec:drugs}

To demonstrate the capability of MouseGPT in revealing behavioral patterns and internal structure, we applied it to mice treated with a series of psychoactive drugs. We examined two classic psychedelics, Psilocybin (Psi) and Lysergic Acid Diethylamide (LSD), together with MK801, which is frequently used to model schizophrenia-related phenotypes in mice. All three compounds can induce hallucinatory-like responses in mice, yet the specific differences among the resulting behavioral states remain to be elucidated.

Recordings were initiated immediately after acute drug administration, and the collected data were subjected to 3D skeletal reconstruction and semantic description generation, followed by embedding and clustering to obtain behavioral classification results for all mice  (Extended Fig.~\ref{fig:ef6}a,b).

Based on these results, we mapped the spatial and temporal distributions of various behaviors in mice under different drug treatments (Fig.~\ref{fig:fig4}a,b), and quantified their occupancy (Fig.~\ref{fig:fig4}c,d). The results indicated that all three drug treatments altered the occupancy of multiple behaviors and their occurrence patterns in the arena. These differences collectively suggest that LSD approximates a sedation-like state in mice, Psi moderately reduces mouse activity, whereas MK801 induces hyperactivity.

\clearpage
\begin{figure}[p]
\setcounter{figure}{3}
\renewcommand{\thefigure}{\arabic{figure}}
\renewcommand{\theHfigure}{main.\arabic{figure}}
\renewcommand{\figurename}{Fig.}
\centering
\includegraphics[width=\textwidth,height=0.90\textheight,keepaspectratio]{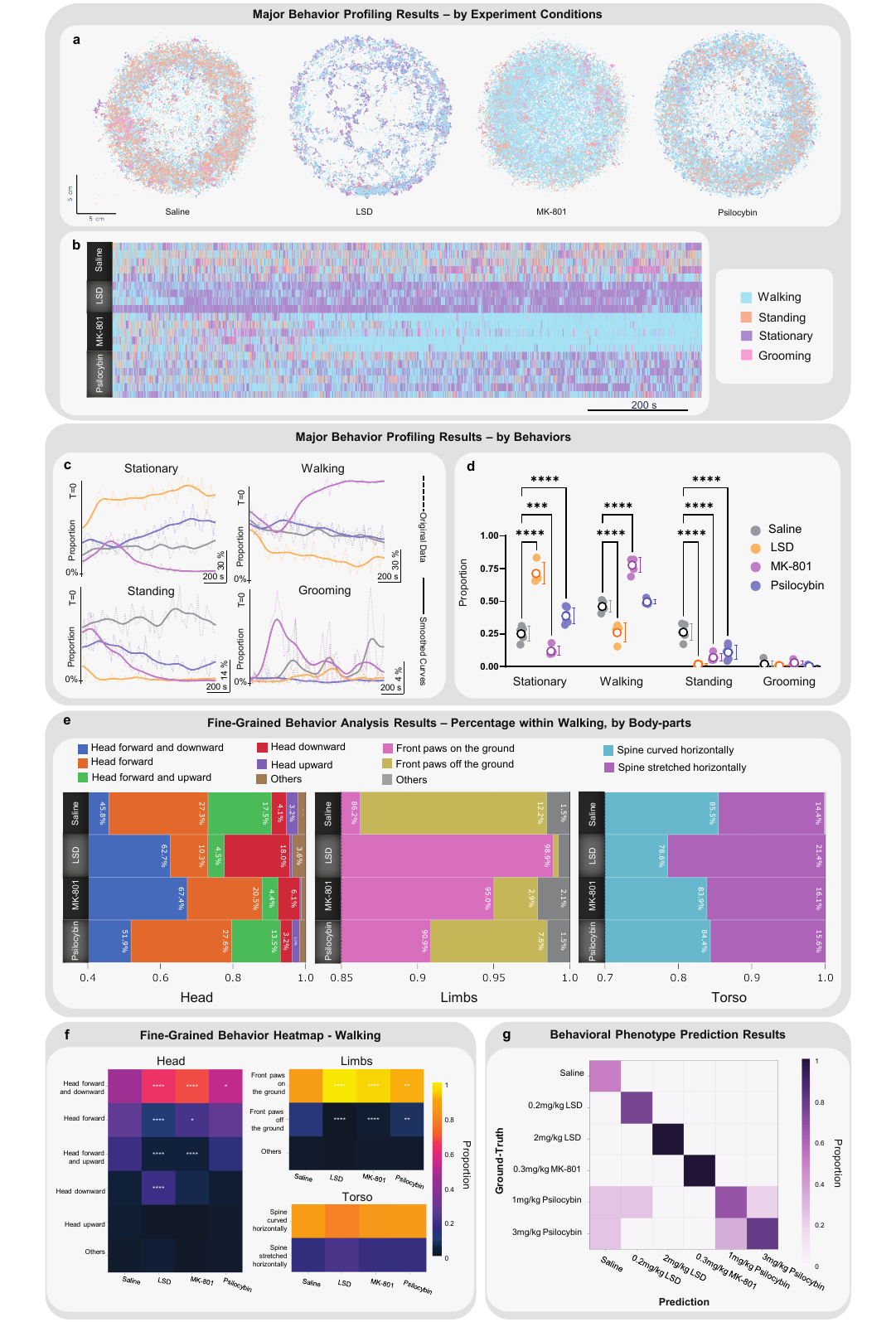}
\caption{\textbf{MouseGPT Analyzes Distinct Phenotypes of Spontaneous Behaviors Induced by psychoactive compounds.}}
\label{fig:fig4}
\end{figure}
\clearpage
\begingroup
\figurecaptionfont
\noindent\textbf{Fig.~\ref{fig:fig4} (continued)}\hskip.7em
\textbf{a}, Spatial Distribution of Major Behaviors: The spatial positions and behaviors of mice in a circular open field were mapped. Each circle represents all mice in a treatment group, with positions determined by spinal midpoints and behaviors color-coded as in b.
\textbf{b}, Major Behavior Ethogram: Time-resolved behavior changes during the first 20 minutes post-treatment are depicted for each mouse in all groups, sampled consistently with a. Rows represent individual mice, illustrating behavioral diversity.
\textbf{c}, Temporal Occurrence Patterns of Behaviors: Behavior proportions within 20-second time windows were visualized for different treatments. Dashed lines show group averages, and smoothed solid lines highlight trends, all starting from the same reference time T=0. Scales were adjusted to enhance the visualization of low-frequency behaviors.
\textbf{d}, Overall Behavioral Proportions: Mean behavior proportions during the entire recording session are presented with circles (mean) and error bars (95\% confidence interval). Statistical significance was tested using two-way ANOVA with Tukey's multiple comparisons (**\textit{P} \textless 0.01; ***\textit{P} \textless 0.001; ****\textit{P} \textless 0.0001).
\textbf{e}, Fine-Grained Postural Subtypes in Walking Behavior: Stacked bar charts illustrate the proportions of walking subtypes categorized by head, limb, and torso descriptors. Each bar represents the average subtype proportion across walking segments within each treatment group.
\textbf{f}, Heatmap of Fine-Grained Behaviors: Proportions of head, forelimb and torso postures during walking are shown for each treatment group. Asterisks indicate comparisons with the saline control. Statistical significance was assessed using two-way ANOVA with Dunnett's multiple comparisons (*\textit{P} \textless 0.05; **\textit{P} \textless 0.01; ****\textit{P} \textless 0.0001).
\textbf{g}, Drug Prediction Matrix: MouseGPT predicted drug treatments based on behavioral phenotypes, using data from six treatment groups (n=23). Rows indicate actual treatment groups, columns indicate predicted groups, and color intensities reflect the proportion of mice assigned to each predicted group.
\par
\endgroup
\medskip

We next investigated whether different drugs also affected the internal structure of behaviors, namely whether the postures within each behavior category, defined by the geometric characteristics and relative positional relationships of body parts, differed in type and proportion. We selected walking, a behavior with diverse postural subtypes, and explored the distribution of these subtypes along the dimensions of the head, limbs, and torso (Fig.~\ref{fig:fig4}e and  Extended Fig.~\ref{fig:ef7}a--c). None of the drugs altered the internal subtype categories of walking behavior, but they significantly affected the proportion and timing of specific posture types (Fig.~\ref{fig:fig4}f). This analysis shows that MouseGPT can resolve drug-dependent differences within a shared behavior category without altering the underlying fine-grained behavioral taxonomy.

Finally, we utilized the built-in behavioral phenotype classifier of MouseGPT to test whether MouseGPT could accurately predict relationships among the different drug conditions (Supplementary Fig.~2). By integrating the behavioral results described above, the classifier was able to discriminate between known and unknown drugs with a high degree of accuracy (Fig.~\ref{fig:fig4}g).

Through the analyses presented above, we elucidated the specific differences in the phenotypic manifestations of hallucinatory-like behavior induced by three distinct drugs, thereby constructing individual behavioral profiles. These findings provide useful assessment tools and indicators for further investigation into the neurobiological mechanisms underlying hallucination-related states, as well as for evaluating drug-specific effects.

\subsection{MouseGPT Provides a User Interface for Natural Language Interaction}
\label{sec:nlui}

We developed a natural-language user interface for mouse behavior analysis, enabling users to interact with an agent through open-ended natural-language queries (Fig.~\ref{fig:fig5}a). This interface incorporates the core analytical methods of MouseGPT, including clustering, fine-grained behavioral analysis, novel behavior discovery, behavior searching, and statistical summarization, allowing users to access the framework through a unified conversational workflow. Recent work has highlighted the potential of language-based agents in behavioral analysis~\cite{ye2023amadeusGPT}; here, we further connect such interaction to the semantic behavior representations produced by MouseGPT.

\begin{figure}[!htbp]
\setcounter{figure}{4}
\renewcommand{\thefigure}{\arabic{figure}}
\renewcommand{\theHfigure}{main.\arabic{figure}}
\renewcommand{\figurename}{Fig.}
\centering
\includegraphics[width=\textwidth,height=0.8\textheight,keepaspectratio]{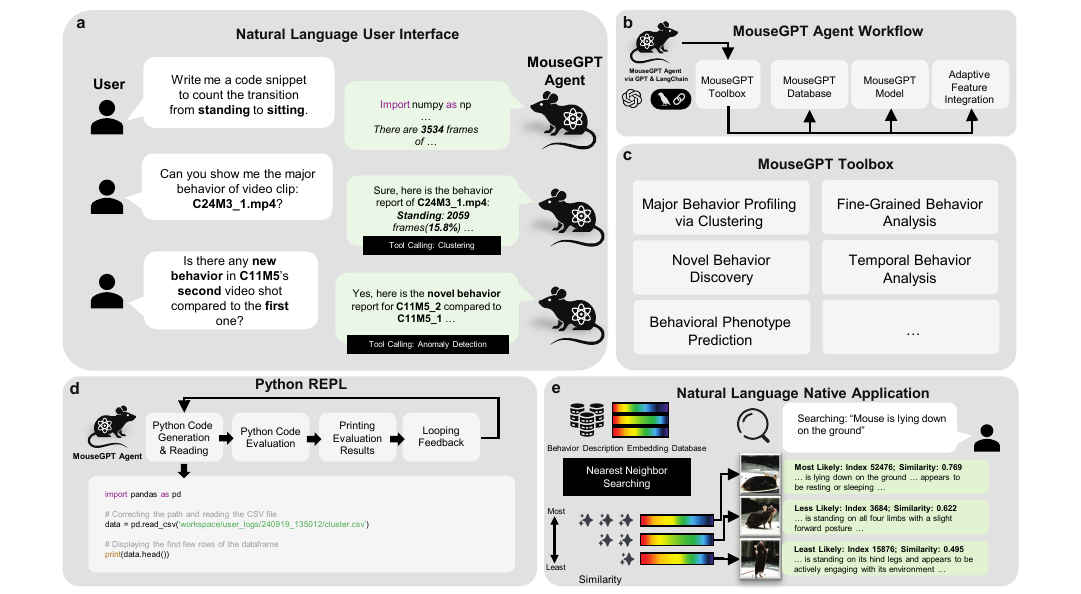}
\caption{\textbf{MouseGPT Natural Language User Interface.}
\textbf{a}, Interactive Interface: Users engage with the MouseGPT agent through natural language to perform clustering, behavior search, and automated code generation and execution, enabling intuitive interaction.
\textbf{b}, Agent Workflow: The agent integrates data access, MouseGPT model invocation, and analysis tasks via a unified workflow, streamlining complex behavior analyses.
\textbf{c}, Toolbox Overview: A comprehensive suite of tools supports major behavior profiling, fine-grained analysis, novel behavior detection, temporal analysis, and phenotype prediction.
\textbf{d}, Python REPL Integration: The agent dynamically generates and executes Python scripts to process data, exemplifying seamless automation.
\textbf{e}, Behavior Search: Embedding-based retrieval matches user-specified behaviors to the database, ranking results by similarity for precise behavior identification.}
\label{fig:fig5}
\end{figure}

At the heart of this framework is an agent powered by a large language model that processes open-ended natural-language input. This LLM is distinct from the MouseGPT model used to generate behavioral descriptions. The agent can analyze long video sequences holistically or examine specific frames in detail. Rather than operating only on raw visual data, the agent invokes a range of tools to access the MouseGPT database and analytical modules, improving both flexibility and scalability (Fig.~\ref{fig:fig5}b). Crucially, because MouseGPT represents behavior natively as natural-language descriptions together with associated semantic embeddings, these representations can be directly consumed by the LLM and reused across tools for retrieval, comparison, summarization, and multi-step reasoning. In this way, MouseGPT's semantic features serve not only as interpretable outputs for the user, but also as working representations within the agentic workflow.

To rigorously quantify the reliability of this agentic workflow, we conducted a standardized benchmark comprising 100 interaction trials spanning 20 diverse tasks across five independent datasets (Supplementary Table~3, Methods). The agent achieved an overall success rate of 87\% without requiring manual intervention, demonstrating highly robust multi-step reasoning, tool orchestration, and code execution capabilities (Extended Fig.~\ref{fig:ef9}).

The agent's toolset includes all of MouseGPT’s analytical methods, such as clustering, fine-grained behavior analysis, novel behavior discovery, and behavioral phenotype prediction, as well as tools integrated into the interactive interface (Fig.~\ref{fig:fig5}c). If the existing tools are insufficient for a specific task, the Python REPL tool (Fig.~\ref{fig:fig5}d) enables the agent to dynamically generate and execute code in a virtual environment, extending the framework to customized analyses while preserving the intermediate computational steps and outputs.

The search tool allows users to query behaviors using natural-language descriptions (for example, ``Mouse is lying down on the ground''). The query is mapped into the same high-dimensional semantic space used throughout the MouseGPT analysis framework, allowing the agent to retrieve the most relevant frames for detailed behavioral analysis (Fig.~\ref{fig:fig5}e). Throughout the interaction, outputs from individual tools are summarized back into natural language and incorporated into the dialogue context, enabling intermediate results to inform subsequent analytical steps (Extended Fig.~\ref{fig:ef8}).

\FloatBarrier

\section{Discussion}\label{sec12}

MouseGPT introduces open-vocabulary semantic analysis that makes behavioral observations explicit, interpretable and computationally reusable. Natural language serves as a shared representation connecting visual observations to descriptions of actions and body-part states, and those descriptions to computable behavioral features (Figs.~\ref{fig:fig2} and~\ref{fig:fig3}). This design supports flexible analysis across scientific questions, forward compatibility with newer VLMs, and interoperability with community pose-estimation models. An expert-verified benchmark with automated grading supports consistent evaluation of new VLMs as the framework evolves.

The scientific value of this representation extends from major behavior profiling to fine-grained postural analysis (Fig.~\ref{fig:fig3}). In the drug experiments, differences were evident in behavioral occupancy and the composition of postures within walking (Fig.~\ref{fig:fig4}), as well as in the temporal profiles of body-part postures (Extended Fig.~\ref{fig:ef7}). Semantic grouping complements kinematics-based behavioral decomposition~\cite{Luxem2022,B-SOiD,weinreb_keypoint-moseq_2024} (Supplementary Fig.~3), providing an explicit behavioral interpretation alongside measurements of movement (Extended Fig.~\ref{fig:ef4}). Compatibility with DeepLabCut and SLEAP~\cite{Lauer2022,Pereira2022} (Extended Fig.~\ref{fig:ef2}) and evaluation on single-camera recordings (Extended Fig.~\ref{fig:ef5}) support use within existing laboratory workflows. The natural-language interface further enables researchers to retrieve behavioral examples and coordinate comparison and summarization through analysis tools as their questions develop (Fig.~\ref{fig:fig5}).

Beyond interpretable behavioral representations, we seek quantitative explainability of how MouseGPT generates descriptions and derives downstream predictions. For description generation, we assess input dependence and output grounding. Predictions combine knowledge acquired through pretraining and fine-tuning with observation-specific evidence. Our kinematic-input ablation tests input dependence while holding the model and visual inputs fixed, with human evaluation separately assessing description quality where outputs differ (Extended Fig.~\ref{fig:ef2}). The expert-verified benchmark assesses output grounding through claim-level correctness and required-fact coverage; optional observations are checked for correctness when asserted but their omission is not penalized (Extended Fig.~\ref{fig:ef3}). Retained grading evidence connects claims to reference facts, making errors and omissions inspectable. Fixed inputs, references and scoring criteria support consistent automated comparisons as new VLMs become available.

Downstream measurements are traceable to descriptions and source frames. For phenotype predictions based on standardized behavioral profiles and Euclidean distances (Supplementary Fig.~2), future analyses could quantify each measurement's contribution to the squared distance and test whether omitting influential measurements changes the inferred phenotype. Generated reports could be assessed by the proportion of claims supported by underlying measurements and agreement between reported and computed values, making the basis of downstream predictions and explanations more explicit.

\subsection{Limitations}

The dataset, fine-tuned models and native kinematic pipeline were developed primarily for individual mice. Social-behavior descriptions were obtained with adapted prompts using MouseGPT-Lite on CalMS21 and general-purpose VLMs on MABe 2022 (Extended Fig.~\ref{fig:ef5}). These experiments demonstrate adaptability of the description-based analysis, while extension of the complete native workflow requires more generalizable kinematic extraction and reliable association of movements with individual animals over time. Broadly pretrained foundation models for animal pose estimation, tracking and re-identification offer a route toward reducing dependence on setup-specific training and maintaining consistent animal identities. 

The current MouseGPT models were fine-tuned on GPT-4o-generated descriptions and may inherit errors and limitations in that supervision. Expert evaluation supports the utility of these annotations (Fig.~\ref{fig:fig2}) but does not eliminate annotation errors. The higher description scores achieved by newer VLMs under the same benchmark protocol (Extended Fig.~\ref{fig:ef3}) demonstrate an opportunity to improve the supervision available to the framework. The MouseGPT Benchmark Index is a rubric-based score and is distinct from the earlier expert-rated correctness measure. Realizing these gains in locally deployable MouseGPT models would require generating or refining training descriptions with stronger annotation models, including newer open-source VLMs assessed with the benchmark, and repeating fine-tuning. Updating both the supervision and the open-source base models could improve future versions while retaining the benefits of local deployment.

\subsection{Future directions}

Video-capable VLMs, including general-purpose models such as Gemini~\cite{comanici2025gemini25} and Qwen3-VL~\cite{bai2025qwen3vl} and action-focused models such as LLaVAction~\cite{qi2026llavaction}, offer a complementary direction for behavioral description generation. Interpreting actions across image sequences could incorporate temporal and interaction information directly and potentially reduce dependence on separately supplied kinematic cues. Their integration would require adapting inputs and prompts and evaluating the resulting descriptions on mouse recordings. Together with more transferable pose-estimation and re-identification models, these developments could support more complex social settings while preserving the downstream language-based analysis.

Future benchmarking on MABe 2025~\cite{mabe2025challenge} would provide an opportunity to evaluate generalization across laboratories, recording conditions and behavioral settings. Combining such broader evaluations with consistent expert-verified assessment of description quality would support the continuing development of MouseGPT: an interoperable framework that makes behavioral observations computationally reusable through explicit, flexible semantic representations.

\newpage
\bibliography{sn-bibliography_remove}
\newpage

\FloatBarrier

\section{Methods}\label{sec11}
\subsection{Data Collection and Experiment Settings}\label{methods:data}
\paragraph{Dataset Overview}\label{methods:dataset}

We collected and curated a comprehensive open-vocabulary dataset to train and evaluate MouseGPT, our model for understanding mouse behavior. This dataset encompasses synchronized multi-view video sequences, 3D keypoint annotations for each frame (see Data Annotation and Curation), detailed textual descriptions (see Automated Open-vocabulary Behavior Annotation), as well as metadata related to medication and mouse modeling (see Animals and Husbandry).

Using a multi-view acquisition system, we finally gathered 75 multi-view videos of 55 different mice, totaling over 1,486 minutes and 42,822,048 frames across 5,352,756 time points. Each frame was automatically annotated with 3D keypoints and 2D bounding boxes to capture the precise position and movement of the mice. After training our models, 892,131 frames from experimental scenarios were automatically annotated with detailed open-vocabulary descriptions using our MouseGPT models, MouseGPT-Large, and MouseGPT-Lite. Access to these data is described in Data Availability.

We also developed two dedicated datasets for model training. The first dataset, used to train our 3D keypoint detection system, contained 2,161,416 manually labeled keypoints across 90,059 images. The second dataset was created for training the Vision-Language Model (VLM), where GPT-4o was utilized to generate detailed textual descriptions for 45 video sequences involving 37 mice, followed by a data curation process that produced 270,085 high-quality frame-level open-vocabulary annotations.

\begin{figure}[!htbp]
\setcounter{figure}{0}
\renewcommand{\thefigure}{\arabic{figure}}
\renewcommand{\theHfigure}{extended.\arabic{figure}}
\renewcommand{\figurename}{Extended Fig.}
\centering
\includegraphics[width=\textwidth,height=0.8\textheight,keepaspectratio]{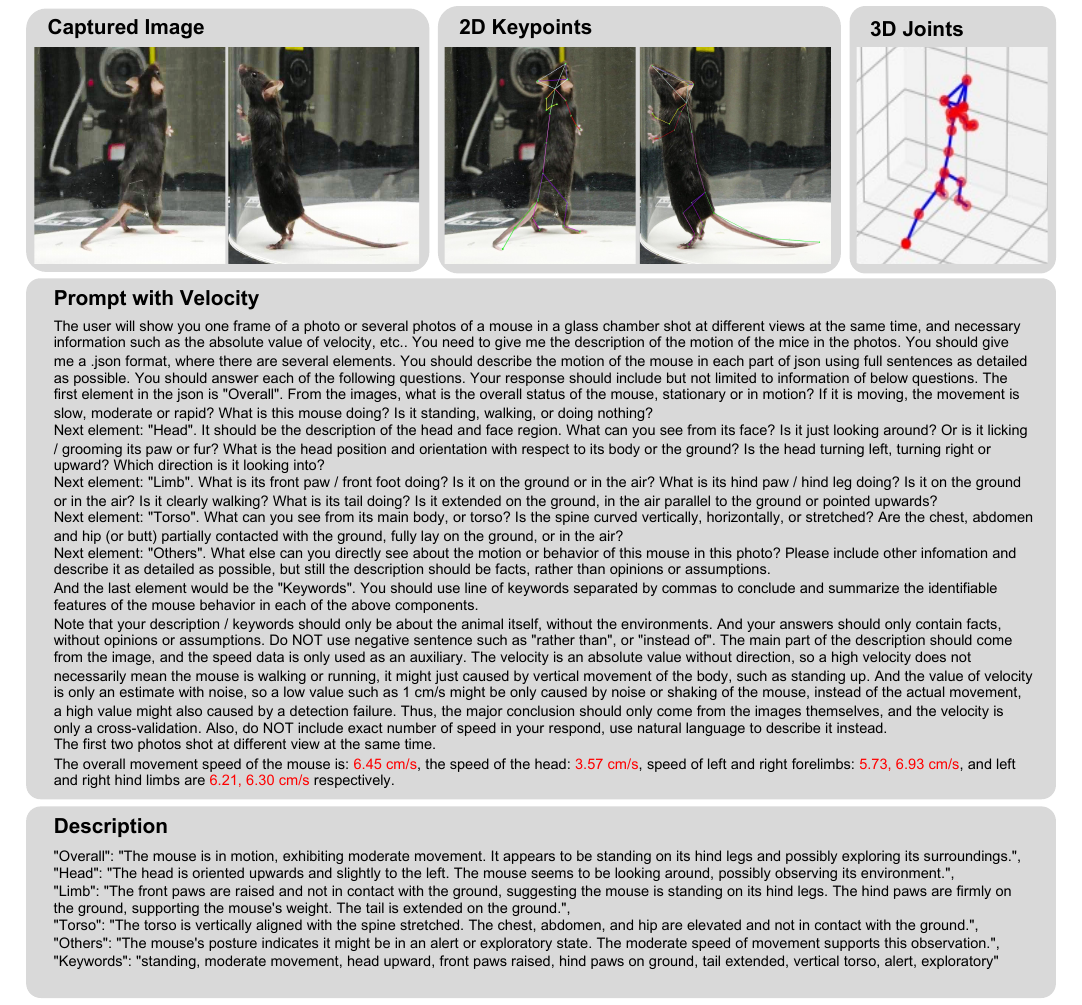}
\caption{\textbf{Data samples and structured behavioral descriptions in MouseGPT.}
An example of a single sample from the MouseGPT dataset, illustrating how multiview images, three-dimensional joint keypoints and velocity data are combined to produce structured, open-vocabulary behavioral descriptions. The figure demonstrates the dataset's ability to capture detailed annotations for overall behavior, head orientation, limb positioning, torso posture and additional traits, summarized through concise keywords.}
\label{fig:ef1}
\label{fig:exfig-data_sample}
\end{figure}

\paragraph{Animals and Husbandry}
A total of 55 8-12-week-old C57BL/6J male mice (Silaike Experiment Animal Co., Ltd., JAX:000664) were used for recording and analysis in this study, including 24 mice for depression behavior capture and 31 mice for hallucination. After introduced into the colony at 8 weeks, mice were housed in groups of 4 to 5 per cage to habituate for at least 1 week with a controlled temperature of 22-25\textcelsius, and 40-70\% humidity. Food and water were provided ad libitum. The circadian rhythm of the mice varied according to the different experiments. For mice in depression experiments, they were kept under a normal 12-hour light-dark cycle (lights on from 7:00 am to 7:00 pm). In contrast, mice in the hallucination experiments were placed under the reverse 12-hour cycle. 
All experiments procedures were approved by the Institutional Animal Care and Use Committee (IACUC) of ShanghaiTech University.

\paragraph{Behavioral experiment}

In depression behavior experiment, we randomly select n=16 mice as Chronic Restraint Stress (CRS) group and n=8 as control group. CRS mice were put into 50 mL conical tubes during the light phase for 2–3 h per day for 16 consecutive days. Holes were drilled on the tubes to enable mice to breathe freely. We recorded the mice on the day following the end of restraint, and then assessed their level of depression using the Tail Suspension Test (TST), in which mice were suspended by binding their tail with adhesive tape fastened to a rail 50 cm from the ground. A 6-minute video recording started immediately after the mice were taped. The immobility duration in the last 4 minutes of the experiment was measured by another member of our group who was blinded to mouse treatment. Finally, mice with immobility duration higher than the average time of the control group were classified as depression-like mice.
For experiments involving drug administration, in order to avoid the potential cumulative effects of hallucinogenic drugs, all mice in such experiment were recorded only once after drug administration to ensure the reliability of subsequent analysis.

\paragraph{Drug treatment}
We selected two classic psychoactive compounds, Lysergic acid diethylamide (LSD) and Psilocybin (both from Laboratory of Medicinal Chemistry and Biology, Changchun Institute of Applied Chemistry, Chinese Academy of Sciences), as well as MK801 (ApexBio, Cat\#A3100), an NMDA receptor antagonist, to induce hallucination-related behaviors in mice. 
Psilocybin and LSD in powder form were initially dissolved in 0.9\% saline to a concentration of 1 mg/mL, then stored in a -80\textcelsius~freezer until the day of use, when they were further diluted with fresh 0.9\% saline. MK801 was first dissolved in DMSO and then diluted with 0.9\% saline on the day of use. All drugs were delivered via intraperitoneal injection. During the two days prior to behavior recording, the mice, in their home cages, were transferred to the recording facility for 1-2 hours to preliminarily acclimate to the environment, typically during the same time window as the recording sessions (1:00 pm to 3:00 pm).

\paragraph{Video Recordings}\label{para:recording}
In our data collection settings (Fig.~\ref{fig:fig2}a-left), all 55 mice that appeared in our dataset were captured under the same controlled conditions and using identical capture devices. The capture setup consists of 8 synchronized Z-Cam E2 cinema cameras with Olympus 14-42mm 1:3.5-5.6 EZ lenses, capturing at 4K resolution, 60 FPS. The cameras are connected in a daisy-chain configuration, with the first camera serving as the master camera to send synchronization signals, and the remaining seven cameras functioning as slave cameras to ensure synchronization. Seven cameras are arranged centrally on a circular ring with a diameter of 1.5 meters at the same height as the mouse activity surface, directed towards the center of the activity area. Additionally, one camera is mounted at the top of the apparatus at a height of 1.2 meters, capturing footage from above with the center of the mouse activity area as the focal point.

On the day of recording, the mice were transferred to the recording facility 1 hour in advance and kept under dark conditions. Video recording began immediately after the mice were moved from their home cages to an open arena with a diameter of 25 cm, and continued for 20 minutes. The arena was then cleaned and prepared for the next mouse.

\paragraph{Camera Calibration}
In order to acquire accurate 3D information from 2D camera arrays, we pre-calibrate our camera using Agisoft Metashape 2.0.2. The camera calibration process includes camera intrinsic calibration and extrinsic calibration. During the camera intrinsic calibration process, we used a MacBook Pro (15-inch, 2017) to display a $22\times14$ checkerboard pattern, ensuring that the pattern was presented on a perfectly flat surface. With all cameras set to the same recording parameters as those used in the actual sessions, the checkerboard pattern was shown to each camera from various positions and angles, covering the entire capture area. The Agisoft Metashape 2.0.2 is used to calculate the camera intrinsic parameters, including the focal length: $f$, principal point coordinates: $c_x$, $c_y$, radial distortion coefficients: $k_1$, $k_2$, $k_3$, and the tangential distortion coefficients: $p_1$, $p_2$. These parameters are calculated from observation of chessboard, used for correcting lens distortions and ensuring accurate image capture. Once calibrated, these parameters are preserved the same during all of the dataset recording sessions.

The camera extrinsic calibration process was conducted differently to account for potential movements of the recording apparatus during filming, as well as shifts caused by gravity. To ensure accuracy, the extrinsic calibration was performed each day before the recording sessions, and the calibration results were only valid for the recordings made on that particular day. We printed an auxiliary pattern for extrinsic calibration on an A3-sized sheet of paper (with dimensions of $297 \times 420$-mm). The central area of the calibration sheet was left blank, measuring $250 \times 250$-mm, to preserve the mouse activity area. At the center of this blank area, a black dot with a diameter of 3mm was printed to serve as the origin of the coordinate system. Alignment patterns, each consisting of two AR Tags, were printed in the four corners of the calibration sheet. The distance between the centers of each pair of alignment patterns was 26.0 cm along the long side and 24.0 cm along the short side. This information was used in the subsequent calibration calculations. The calibration sheet was carefully adhered to the base of the recording apparatus to ensure the accuracy of the calibration pattern.

We designed calibration objects with complex pattern features that cover a $25.0 \times 25.0$-cm area. These objects included various cubes and rectangular prisms, which were placed within the filming area. Using the $48$mm lens of an iPhone 15 Pro Max (which had also undergone the intrinsic calibration process described earlier), we captured 20-50 photos uniformly sampled between each pair of adjacent cameras among the eight cameras. The purpose of this uniform sampling was to supplement the eight sparsely distributed cameras with sufficiently dense observational angles, allowing for a more accurate determination of the spatial positions of the primary cameras.

The sampled photos, along with images from the eight main cameras, were then imported into MetaShape for extrinsic camera calibration. During this process, we manually marked the five previously mentioned markers (the center point of the activity area and the center points of the four corner calibration patterns) within the viewpoints of the eight main cameras. These markers were assigned their actual distances in the reference module, establishing a 3D right-handed coordinate system with the activity area center as the origin.

Next, we manually selected 5\%-10\% of the iPhone-captured images and similarly annotated the five markers. Using MetaShape's extrinsic calibration tools, we aligned the photos with a Key Point Limit of 15,000 per Mpx, a Tie Point Limit of 10,000, and enabled features such as Guided Image Matching, Adaptive Camera Model Fitting, and Highest Accuracy. After aligning the photos, we performed an ``Optimize Camera Alignment'' process to further refine the 3D positions and rotations of the cameras.

With this, we completed the intrinsic and extrinsic camera calibration process. The resulting camera parameter files will be used for 3D spatial computations, supporting tasks such as pose estimation and speed calculation.

\subsection{Data Annotation and Curation}

\paragraph{Overview: Data Annotation and Curation} 
We developed a comprehensive data annotation and curation workflow for training and evaluating the MouseGPT model, using a combination of manual and automated methods to create a robust dataset. This workflow generated pose estimates and 2D/3D kinematic data, providing essential spatial and temporal information for VLM inference. Subsequently, we applied an automatic annotation strategy using the VLM to generate open-vocabulary behavioral descriptions (Extended Fig.~\ref{fig:ef1}).

Our approach began with manual annotation of 2D poses from eight different perspectives. Specifically, we annotated 2,161,416 keypoints across 90,059 images, providing a high-quality labeled dataset that served as the foundation to train a neural network for automated kinematic data extraction across the MouseGPT dataset. Automated pose estimation was conducted using well-established models, such as YOLOv5 and HRNet-based architectures. Through these methods, we obtained accurate 2D and 3D poses, and thus calculated velocities, which served as input to Vision-Language Models (VLMs), providing essential spatial and temporal information for understanding complex mouse behaviors across different experimental conditions.

With the essential spatial and temporal data in place, we used an automatic annotation strategy for generating open-vocabulary behavioral descriptions, enabling a broader contextual understanding of mouse activity that can be directly queried using natural language. Through this automated yet flexible pipeline, we efficiently extracted and labeled kinematic and descriptive data, creating a comprehensive resource for downstream behavioral analysis and supporting novel insights into ethological research.

\paragraph{Data Pre-processing and Augmentation}

In this study, meticulous data preprocessing and augmentation procedures were employed to ensure the robustness and accuracy of our annotated datasets. We utilized a MATLAB-based software, Label3D, to perform precise 3D annotations from multiple 2D views captured simultaneously from eight distinct angles. We defined our own keypoint configuration, comprising 24 keypoints (Supplementary Table~1) aligned with essential skeletal landmarks on the mouse—such as the snout, various spine segments, limbs, and tail—based on insights from established works like DANNCE and SLEAP.

The anatomical keypoints were meticulously defined to capture motions of the mouse in three-dimensional space, with 24 distinct skeletal landmarks representing critical joints and body segments integral to movement analysis. These included keypoints like EarL and EarR for orientation, Snout for head position, and SpineF, SpineG, and SpineH for spinal movements. Notably, SpineG and SpineH are located as two equidistant points between SpineF and Hip, further enhancing the granularity of spinal movement analysis. Additionally, various points along the limbs and tail were annotated for detailed movement studies.

The annotation process commenced by marking each keypoint in at least two camera views, followed by employing the ``triangulate'' function to project these 2D points into 3D space. This initial automated triangulation provided a baseline, which was refined through manual adjustments of the 2D annotations and subsequent re-triangulation, thus enhancing the accuracy. This iterative process ensured the consistent placement of keypoints across all views.

For keypoints obscured by complex mouse movements or overlapping anatomical features, an additional validation step was implemented to enhance precision. Each annotation was rigorously reviewed, and those identified as being of poor quality are re-annotated. This meticulous validation process helps mitigate errors stemming from both the automated triangulation inaccuracies and subjective judgments during manual annotation, thus bolstering the reliability and consistency of the dataset for subsequent analytical tasks.

To facilitate the effective localization of mice within images, our initial step involved the generation of bounding boxes. We computed the minimum bounding box enclosing all detected keypoints on the mice. To accommodate any variations in the mice's positioning and ensure the entire object was captured, we added a 35-pixel padding to each side of the bounding box. This precaution prevents the bounding boxes from being overly restrictive, thereby enhancing the robustness of the subsequent detection and analysis tasks.

In preparation for the use of the YOLOv5 model, it was imperative to optimize anchor points, which serve as preliminary candidate bounding boxes for object prediction. Given that objects in different datasets exhibit diverse sizes and aspect ratios, unsuitable anchors could impair the model’s ability to precisely localize objects. To address this, we conducted an analysis of the bounding boxes of the target object using K-Means clustering. This analysis aimed to optimize the anchor sizes and aspect ratios, thereby enabling dynamic adjustments during training to better align with the object distribution in our dataset. This alignment is crucial for enhancing the detection accuracy and the generalization capabilities of the model.

Data augmentation is essential in enhancing the model's adaptability and performance during the preprocessing phase. To optimize the input for the mmyolo~\cite{mmyolo2022} and mmpose~\cite{mmpose2020} networks, we adjusted the size of the images, which not only conforms to the network requirements but also facilitates faster processing. Focusing on strengthening the model's ability to generalize across diverse scenarios and increase its resilience to overfitting, we implemented a variety of data augmentation strategies. Specifically, we applied random cropping to simulate different object positions and scales, horizontal flipping to introduce variability in object orientation, random rotation to account for various object angles, and color jittering to ensure robustness against variations in lighting and color. These augmentation techniques diversify the training dataset, promoting the development of more universal feature recognition within the model and minimizing its tendency to overfit specific patterns. This strategic enhancement of data variability ensures that our model maintains high performance under a wide range of environmental conditions and animal movements.

\paragraph{2D Kinematics Estimation Workflow and Network Design}

In order to automatically estimate 2D kinematic information from video sequences, we employed a two-stage detection and pose estimation framework, utilizing the mmyolo and mmpose training frameworks provided by mmlab to ensure better processing speed and robustness.

In the first stage, we use the YOLOv5-s~\cite{glenn_jocher_2022_7347926} network as the object detector to detect and extract the bounding boxes of mice from the input images. YOLOv5 is a widely-used real-time object detection network that offers a good balance between speed and accuracy. The network consists of three main components: the backbone, neck, and head. The backbone extracts feature maps from the input image through a series of convolutional layers, capturing essential visual information. The neck component enhances these feature maps by combining multi-scale information, improving the detection of objects of varying sizes. Finally, the head component makes the final predictions by generating bounding boxes and classifying the detected objects. In our implementation of YOLOv5-s network, we use CSPDarknet as backbone to extract multi-dimensional features from the image, facilitating the detection of objects of various sizes. The deepen factor of our backbone is 0.33 and the widen factor is 0.5, controlling the number of layers in the network and the number of channels in the output feature map, respectively. The neck module employs a PAFPN (Path Aggregation Network with Feature Pyramid Network) with a deepen factor of 0.33 and a widen factor of 0.5, and channel numbers of [256, 512, 1024]. This component processes and fuses feature maps from different layers of the backbone, outputting a series of feature maps at various scales. The head module is a 3-layer convolutional network that converts these feature maps into outputs that include object presence, bounding box, and confidence levels within anchors.

After obtaining the bounding boxes of the mice, the process moves to the pose estimation stage. The pose\_hrnet\_w32 architecture typically consists of a backbone and a head module. We use HRNet~\cite{wang2020deephighresolutionrepresentationlearning} (High-Resolution Network) as the backbone, composed of 4 stages, processing feature maps of different resolutions through a multi-branch structure. The first stage has a branch with 4 Bottleneck blocks and 64 channels. In the second stage, two branches are introduced, with 48 and 96 channels, utilizing Basic residual blocks. The third stage further expands the network’s feature extraction capabilities with three branches having 48, 96, and 192 channels, respectively. Finally, in the fourth stage, the network has four branches, with channels increasing from 48 to 384. This multi-branch, multi-channel design allows HRNet to maintain high resolution while fully integrating multi-scale features, providing strong feature support for pose estimation. The head of the model employs a Heatmap Head. This module's main function is to convert the feature maps extracted by the backbone into heatmaps of keypoints, where each location on the heatmap corresponds to the probability of a keypoint. The Heatmap Head receives feature maps from the lowest resolution branch of HRNet, with 48 input channels, and uses a series of convolution operations to convert these feature maps into heatmap outputs of keypoints. Ultimately, the head outputs heatmaps with 24 channels, each corresponding to the probability distribution of a keypoint location. This design allows the Heatmap Head to effectively convert high-level feature maps into precise keypoint predictions, providing reliable positioning capabilities for pose estimation.

Given the substantial volume of data, we implemented parallelizable data loading, pre-processing, and post-processing features based on the original mmlab framework, utilizing high-performance CPUs and GPUs. This parallelization significantly accelerated training and reduced the computational cost associated with processing large video datasets.

\paragraph{Training Details of 2D Kinematics Estimation Network}
For the 2D mouse detector, mmyolo, we used the SGD optimizer with a learning rate (lr) of 0.001, momentum of 0.937, and weight decay of 0.0005. To further optimize the training process, we implemented a linear dynamic learning rate strategy, with the lr factor set to 0.01, which means that the learning rate will gradually decrease linearly to 1\% of the original learning rate as the number of training epochs increases. These settings enable the optimizer to gradually adjust the learning rate during training, improving the model's convergence speed and final performance.

For the 2D mouse pose estimator, mmpose, we used the Adam optimizer with an initial learning rate (lr) of 5e-4. The learning rate schedule consisted of two parts: first, a linear learning rate adjustment (LinearLR) was used for warm-up, with a start factor of 0.001, beginning from the 0th epoch and continuing until the 200th epoch. After the warm-up, a multi-step learning rate adjustment strategy (MultiStepLR) with a gamma parameter of 0.1 was employed. This combination strategy allows the model to gradually increase the learning rate in the early stages and then decay the learning rate during critical training stages, enhancing the model's stability and accuracy.

To ensure the generalization performance of the model, we reserved the first five minutes of data from each video in the dataset as the test set. The remaining data was divided into training and validation sets in an 8:1 ratio. The training set was utilized for optimizing model parameters, the validation set for selecting the optimal model, and the test set for evaluating the final model's generalization capability. This partitioning strategy ensures data independence across different stages, thereby reducing the potential for model overfitting.

We conducted the training using an NVIDIA RTX 4090 24GB. The mmyolo training lasted for 200 epochs, taking approximately 11 hours, while the mmpose training also lasted for 200 epochs and took around 11 hours.

\paragraph{Error Metrics}

During training, the mmyolo framework separately computed object loss and bounding box loss using Cross-Entropy Loss. Given that our dataset contains only a single class (mice), class loss was not calculated. We assigned weights of [1.0, 0.05] to the object loss and bounding box loss, respectively, to achieve more effective joint optimization. In the mmpose framework, we computed MSE Loss of keypoints, which supervises the proximity of keypoints derived from the heatmap to the ground truth (GT).

In evaluating the two-stage pose estimation, we primarily utilized Average Precision (AP) and mean Average Precision (mAP) as performance metrics, alongside PCKAccuracy to assess performance on the test set. Our approach achieved pixel-level accuracy on 4K images.

Model performance was predominantly assessed using the metrics of AP and mAP. AP measures the accuracy of predictions for individual keypoints, whereas mAP provides an overall evaluation of accuracy across all keypoints. Additionally, we monitored the evolution of the loss function to evaluate training progress and convergence.

\paragraph{3D Pose Triangulation}

To accurately capture the three-dimensional motion trajectories of mice, we employed the Easymocap~\cite{easymocap} library, integrating camera parameters with the two-dimensional coordinates of keypoints detected from n distinct camera perspectives (in our study, 
n=8). This setup employed a sophisticated triangulation methodology essential for the precise reconstruction of the spatial positions of these keypoints. The core of our approach is anchored in the optimization of the triangulation equations derived from combining each camera's viewpoint:

\begin{equation} 
\arg \min_{(x, y, z)} \sum_{i=1}^{n} \left\| \left( u_i, v_i \right)^{\mathsf{T}} - \frac{P_{i,1:2}X}{P_{i,3}X} \right\|_2
\end{equation}

where $X=(x,y,z,1)^{\mathsf{T}}$, $P_i$ denotes the projection matrix for camera $i$, $P_{i,1:2}$ denotes its first two rows, and $P_{i,3}$ denotes its third row, which is used to normalize the homogeneous coordinates after projection.
$(u_i,v_i)$ represent the image coordinates of the keypoints. Our methodology synthesizes data across all camera angles, leveraging the collective information to compute the most probable three-dimensional coordinates $(x,y,z)$. This multi-view triangulation approach minimizes spatial uncertainties and refines the accuracy of the positional data.

Utilizing data from multiple perspectives substantially enhances the robustness and reliability of the tracking system. This approach not only augments the spatial resolution of the reconstructed keypoints but also offers a more detailed and nuanced analysis of mouse behaviors, which is pivotal for sophisticated behavioral studies. The comprehensive coverage provided by multiple camera angles significantly reduces occlusions and ambiguities, delivering clearer and more actionable insights into the intricacies of animal movement dynamics.

\paragraph{3D Velocity Calculation}

The 3D keypoint data are used as input for the 3D velocity calculation process, including the keypoint positions of the mouse at each frame, stored in .json format. 

For the $j$-th keypoint position $k^j_i\in \mathbb{R}^3$ at frame $i$, captured under frame rate $fps$, the $j$-th keypoint's velocity at frame $i$, $\mathbf{v}^j_i\in \mathbb{R}^3$, is calculated as follows, representing the mean velocity of the keypoint from the previous frame to the current frame:
\begin{equation}
    \mathbf{v}^j_i = (k^j_i - k^j_{i-1})\,fps
\end{equation}

In order to demonstrate the trend of the movements in a more obvious way, we convert the velocities in reference to the overall mouse movement using direction vector $d^{base}\in \mathbb{R}^3$ defined by keypoints strategically chosen:
\begin{equation}
    d^{base}_i = k^5_i - k^6_i
\end{equation}
where $k^5_i$ (SpineG) and $k^6_i$ (SpineH) denote the 3D coordinates of the 5-th and 6-th keypoints at frame $i$.

The converted velocity $v^j_i \in \mathbb{R}$, is a scalar representing the magnitude of the keypoint velocity $\mathbf{v}^j_i$, with the sign indicating the relative tendency of the movement. It is computed as:
\begin{equation}
    v^j_i = \text{sign}\left(\arccos\left(\frac{\mathbf{v}^j_i \cdot d^{base}_i}{\|\mathbf{v}^j_i\| \|d^{base}_i\|}\right) - \frac{\pi}{2}\right)\cdot \|\mathbf{v}^j_i\|
\end{equation}

\paragraph{Automated Open-vocabulary Behavior Annotation}

We collect a behavior dataset with open-vocabulary annotations using GPT-4o (gpt-4o-2024-05-13). Specifically, we extract images at 5 frames per second from all of our video, utilizing images from two mutually perpendicular viewpoints. We concatenate the two images horizontally as input to GPT-4o. We then generate the prompt with speed information to acquire detailed descriptions of mouse behavior. 

An example prompt is as follows: 

\texttt{The user will show you one frame of a photo or several photos of a mouse in a glass chamber shot at different views at the same time, and necessary information such as the absolute value of velocity, etc.. You need to give me the description of the motion of the mice in the photos. You should give me a .json format, where there are several elements. You should describe the motion of the mouse in each part of json using full sentences as detailed as possible. You should answer each of the following questions. Your response should include but not limited to information of below questions. The first element in the json is "Overall". From the images, what is the overall status of the mouse, stationary or in motion? If it is moving, the movement is slow, moderate or rapid? What is this mouse doing? Is it standing, walking, or doing nothing?Next element: "Head". It should be the description of the head and face region. What can you see from its face? Is it just looking around? Or is it licking / grooming its paw or fur? What is the head position and orientation with respect to its body or the ground? Is the head turning left, turning right or upward? Which direction is it looking into? Next element: "Limb". What is its front paw / front foot doing? Is it on the ground or in the air? What is its hind paw / hind leg doing? Is it on the ground or in the air? Is it clearly walking? What is its tail doing? Is it extended on the ground, in the air parallel to the ground or pointed upwards?Next element: "Torso". What can you see from its main body, or torso? Is the spine curved vertically, horizontally, or stretched? Are the chest, abdomen and hip (or butt) partially contacted with the ground, fully lay on the ground, or in the air? Next element: "Others". What else can you directly see about the motion or behavior of this mouse in this photo? Please include other infomation and describe it as detailed as possible, but still the description should be facts, rather than opinions or assumptions. And the last element would be the "Keywords". You should use line of keywords separated by commas to conclude and summarize the identifiable features of the mouse behavior in each of the above components.Note that your description / keywords should only be about the animal itself, without the environments. And your answers should only contain facts, without opinions or assumptions. Do NOT use negative sentence such as "rather than", or "instead of". The main part of the description should come from the image, and the speed data is only used as an auxiliary. The velocity is an absolute value without direction, so a high velocity does not necessarily mean the mouse is walking or running, it might just caused by vertical movement of the body, such as standing up. And the value of velocity is only an estimate with noise, so a low value such as 1 cm/s might be only caused by noise or shaking of the mouse, instead of the actual movement, a high value might also caused by a detection failure. Thus, the major conclusion should only come from the images themselves, and the velocity is only a cross-validation. Also, do NOT include exact number of speed in your respond, use natural language to describe it instead. The first two photos shot at different view at the same time. The overall movement speed of the mouse is: 2.49 cm/s, the speed of the head: 5.18 cm/s, speed of left and right forelimbs: 17.18, 19.87 cm/s, and left and right hind limbs are 3.38, 2.63 cm/s respectively.}

We access the GPT-4o API through Microsoft Azure to generate our dataset. This dataset encompasses 15 hours of video, from which 270085 frames have been selected. For each selected image, we generated 3D keypoints, velocity data, and a detailed textual description using GPT-4o, alongside annotations regarding the mice's modeling and drug use. The same prompts and descriptions will later be utilized to train our Vision-Language Model. 

An example answer is as follows:

\texttt{\{  "Overall": "The mouse is in motion, exhibiting a moderate movement speed. It appears to be standing on its hind legs and interacting with the glass chamber.",  "Head": "The mouse's head is oriented upward, looking towards the top of the glass chamber. It is not engaged in grooming or licking.",  "Limb": "The front paws are in the air, reaching towards the glass. The hind paws are on the ground, supporting the mouse's body. The tail is extended on the ground.",  "Torso": "The torso is vertically stretched, with the spine aligned vertically. The chest, abdomen, and hip are elevated and not in contact with the ground.",  "Others": "The mouse is actively engaging with its environment, possibly exploring or attempting to climb the glass chamber.", "Keywords": "motion, standing, head upward, front paws in air, hind paws on ground, tail extended, torso vertical, exploring"\}}

\paragraph{Dataset Curation}

To ensure that the outputs generated by GPT-4o conform to a strict JSON format, which is essential for subsequent readability and processing, we have developed a robust JSON validation method. Initially, we capture the response from OpenAI as a raw string, allowing us to apply regular expressions effectively. This approach enables us to search for and extract the content enclosed within the JSON object, thereby filtering out any responses that do not meet the required format. By focusing on this precise extraction process, we can systematically eliminate improperly formatted data before it reaches the next stage of our workflow.

Once we have validated the JSON outputs, we further enhance the quality of our dataset by employing a faster LLM model, specifically GPT-4o mini, to score all generated JSON files. In this phase, we provide both the generated JSON results and the corresponding input images to the model, prompting it to evaluate critical aspects such as accuracy, coherence, and relevance and scoring them from 0 to 10. The evaluation process is guided by specific criteria that help ensure the data meets our standards for training purposes. Consequently, any entries that receive the lowest scores are removed from the training dataset, allowing us to refine our data and improve the overall quality of our training inputs. This dual-layer validation process not only strengthens the integrity of our dataset but also enhances the performance of our models in subsequent training phases.

\subsection{Model Architecture and Training Details}

\paragraph{Training Our Model with the SWIFT Framework}
Our model is trained using the Scalable lightWeight Infrastructure for Fine-Tuning (SWIFT) framework, a comprehensive solution designed for training, inference, evaluation, and deployment of Large Language Models and Vision-Language Models. SWIFT allows us to train  models of different architectures in the same training manner. In addition, SWIFT supports various third-party optimization libraries for training and deployment. In our practice, we use DeepSpeed to accelerate the model for multi-GPU training and LMDeploy to accelerate the model for deployment on multiple GPUs.

We registered the aforementioned dataset as the training dataset within the SWIFT framework and utilized it to train two distinct models: InternVL2-Llama3-76B and MiniCPM-V-v2.6-chat.  

\paragraph{Model Architecture}

We present two distinct models in our work, MouseGPT-Large, and MouseGPT-Lite. The first and most powerful Large Model, leveraging InternVL2-Llama3-76B~\cite{chen2023internvl, chen2024far}, is a Vision-Language Model (VLM) comprising 76 billion parameters. The model demonstrates superior performance in multi-modal tasks, offering advanced capabilities in handling diverse data and delivering accurate results.

The model architecture consists of three main components. First, the vision module employs a query-key-value (qkv) attention structure~\cite{vaswani2017attention} to process visual inputs. This module transforms images into a high-dimensional feature space, capturing spatial and contextual information. Second, the language module utilizes the Llama~\cite{dubey2024llama} for Causal Language Modeling (LlamaCausalLM) architecture, designed for efficient text generation and understanding. Finally, the integration module bridges the vision and language outputs, combining the two modalities to facilitate coherent inference across both visual and textual inputs.

\begin{enumerate}
\item Vision Module. 
    The vision processing module is designed for high-capacity visual tasks, starting with a convolutional patch embedding layer. This layer processes input images (typically 448×448×3) by dividing them into patches using a 14×14 kernel and a stride of 14. Each patch is then projected into a 3200-dimensional feature space, resulting in a grid of 32×32 patches, each represented by a 3200-dimensional feature vector. The encoder consists of 45 layers, each containing a multi-head self-attention mechanism, which projects the input features (3200 dimensions) into a higher 9600-dimensional space for queries, keys, and values (QKV). After attention calculations, the outputs are projected back to 3200 dimensions. Following the attention mechanism, a multi-layer perceptron (MLP) transforms the 3200-dimensional features by first expanding them to 12800 dimensions and then reducing them back to 3200 dimensions. This is done using two linear layers, with GELU activations applied between them. The outputs of both the attention and MLP components are followed by normalize layer and dropout layer.
\item Language Module.
    The language processing component handles natural language tasks by embedding input tokens into an 8192-dimensional space. The language model consists of 80 layers, each equipped with a self-attention mechanism that projects the 8192-dimensional input features into query, key, and value spaces. The key and value projections are reduced to 1024 dimensions, while the query remains at 8192 dimensions. The attention output is projected back to 8192 dimensions. The MLP following the attention mechanism expands the 8192-dimensional features to 28672 dimensions before reducing them back to 8192 dimensions using three linear layers with SiLU activations. Layer normalization is applied after both the attention and MLP operations.
\item Integration and Multi-Modal Processing.
    To combine visual and language information, pixel shuffle spatially downsamples the visual features by a factor of 0.5, increasing their channel dimension from 3200 to 12800. After layer normalization, an MLP projects these features to 8192 dimensions through a linear layer, a GELU activation and a second linear layer that preserves the 8192-dimensional output. The projected visual embeddings are inserted into the language model's input sequence.

\end{enumerate}

The second lightweight model, MouseGPT-Lite, utilizes MiniCPM-V-v2.6-chat ~\cite{yao2024minicpm} and is a Vision-Language Model (VLM) with 8 billion parameters, approximately one-tenth the size of our MouseGPT-Large. This model effectively balances powerful analysis capabilities with enhanced inference speed, making it ideal for use in resource-constrained environments.

The architecture consists of three key components. The language module is based on the Qwen2ForCausalLM model~\cite{qwen2}, which provides robust language processing capabilities optimized for efficiency. The vision module employs the SigLIP model~\cite{zhai2023sigmoid}, designed for high-performance visual feature extraction. These two modules are integrated using a Resampler, which combines the outputs of both the vision and language components.

Specifically, the network structure can be divided into the following components:
\begin{enumerate}
    \item Text generation Module.
The text generation module is based on a decoder-only transformer with 28 layers. Each layer employs a multi-headed attention mechanism, which projects the input features (3584 dimensions) into query, key and value vectors. The attention outputs are processed by a gated multi-layer perceptron (MLP): separate gate and up projections expand the 3584-dimensional features to 18944 dimensions, and the SiLU-activated gate is multiplied elementwise with the up projection. A down projection reduces the result to 3584 dimensions. RMS normalization is applied before the attention and MLP sublayers. 

    \item Vision Processing Module.
The vision processing module is based on a vision transformer that includes 27 layers for processing visual input. Each layer contains a multi-head attention mechanism, projecting the 1152-dimensional visual features into higher-dimensional spaces for queries, keys, and values. The attention outputs are processed through an MLP, which expands the features from 1152 dimensions to 4304 dimensions before reducing them back to 1152 dimensions. This process is regulated by LayerNorm and activated using a GELU-Tanh function, ensuring non-linear transformations of the visual features. The vision module employs a patch-based embedding system, where the input image is divided into patches using a 14×14 kernel, embedding each patch into an 1152-dimensional feature space.
    \item Integration and Resampling.
The integration module plays a critical role in bridging the gap between the outputs of the text and vision processing modules. A resampler module is employed to align the feature dimensions from the two modalities. It utilizes a multi-head attention mechanism to resample the visual features (1152 dimensions) and align them with the text-based features (3584 dimensions). The resampled features are then fused, and a normalization layer ensures that the output features from both the text and vision modules are consistently integrated, facilitating seamless multi-modal processing.
\end{enumerate}

\paragraph{Training Details}

We train our Vision-Language Models (VLMs) using a multi-node distributed setup across 4 nodes, each equipped with 8 NVIDIA A100 GPUs. To optimize training, we use DeepSpeed~\cite{aminabadi2022deepspeed} for efficient distributed operations and employ mixed-precision (bfloat16) to reduce memory usage and improve computational speed. Communication and gradient updates are coordinated using PyTorch’s torchrun utility.
 
The AdamW optimizer is used with a learning rate of 1e-5, following a cosine decay schedule with a warmup ratio of 0.05. Flash attention is applied to enhance the computational efficiency of attention layers, particularly for processing long sequences. To avoid padding and ensure maximum efficiency, we set the batch size to 1 for all training jobs.

Our models are trained on a custom dataset (See Dataset Overview), with 1\% of the data reserved for testing. To align with standard Vision-Language Model (VLM) training protocols, we horizontally combine two images per sample and resize them to a resolution of 896×448. The InternVL2 model undergoes 18,000 iterations, while the MiniCPM model is trained for 45,900 iterations. 

\subsection{MouseGPT Behavior Analysis Framework}

We designed the Advanced Behavior Analysis Framework to best work with our MouseGPT models, utilizing open-vocabulary text descriptions to facilitate comprehensive analysis of mouse behavior. This framework begins with data pre-processing and text embedding generation to convert open-vocabulary descriptions into computable features. It then integrates key components such as major behavior profiling, fine-grained behavior analysis, novel behavior discovery, behavioral phenotype prediction, and search capabilities, providing researchers with powerful tools to derive high-level insights and conduct detailed analyses effectively.

\paragraph{Data Pre-processing}
MouseGPT's behavioral analysis framework is built on the detailed descriptions for each frame using a Vision-Language Model (VLM). In practice, we first sample frames from the single-camera videos or multi-view videos and calculate the speed for each frame. Next, we apply a trained VLM to provide detailed behavioral descriptions for each sampled frame, enabling a comprehensive analysis of the mouse’s behavior.

To reduce computational overhead while preserving important behavioral information, we sample video frames at a rate of 5 frames per second (fps). Most of our video recordings are approximately 20 minutes long, and after sampling, this results in about 6,000 frames per video. This sampling rate is chosen to capture sufficient behavioral changes in the mouse’s movements without overloading the model with redundant frames. The velocity of the mouse (See Data Annotation and Curation) is derived from changes in the mouse's position across consecutive frames, which is critical for understanding motion-based behaviors such as locomotion, exploration, and rearing.

For precise behavior tracking, we calculate the bounding box around the mouse using a trained 2D mouse detection neural network (See Data Annotation and Curation). The bounding box helps localize the mouse in each frame, facilitating the subsequent analysis. For multi-view settings, we can have more robust analysis result by selecting two orthogonal side views from the video recordings, which provide complementary perspectives on the mouse's movements. These information are crucial for capturing both lateral and vertical behaviors, enabling a comprehensive understanding of the mouse’s actions.

Given the sampled frames, we leverage trained MouseGPT model to automatically annotate the behavior descriptions for each frame. For each request, trained MouseGPT model provides both the overall behavior and fine-grained descriptions of specific body parts such as head, limbs, torso, and tail. This approach allows for detailed and contextually rich annotations, capturing the nuances of the mouse's behavior. The detailed prompt and the specific output format for these annotations can be found in Data Annotation and Curation.

\paragraph{Transform Texts into Computable Features}
To enable quantitative analysis in our framework, we designed a dual-approach to convert the descriptive texts generated by our MouseGPT Model into computable features, using text embeddings and Keyword Info-Cards.

First, we convert the textual descriptions into high-dimensional embedding vectors to calculate distances and similarities for applications such as search and clustering. By employing this text embedding approach, we converted the full text of the behavioral descriptions into continuous high-dimensional feature vectors. In our implementation, we use the text-embedding-3-large model~\cite{openai2024embeddingmodels}, which is a text embedding model designed for capturing semantic meaning, to generate a 3072-dimensional embedding vector for each behavioral description. This semantic-aware embedding vector allows similar behaviors to have smaller distance metrics between their corresponding vectors, enabling more accurate comparisons in high-dimensional space.

We also developed a parallel keyword-based matching algorithm that leverages Large Language Models (LLMs). In addition to the fuzzy matching and unsupervised clustering capabilities enabled by high-dimensional text embeddings, the keyword info-card provides an exact matching functionality. Using an LLM, keywords are extracted from the behavioral descriptions and represented as a ``bag-of-words'' — an approach that captures key features of the text without considering word order. This bag-of-words representation forms a computable feature that allows for precise and efficient identification of behaviors, complementing the more flexible semantic embeddings with direct, interpretable metrics.

\paragraph{Major Behavior Profiling - Clustering Implementation}

We designed an LLM-based clustering algorithm for major behavior profiling during behavior analysis, employing a two-stage approach to ensure minimal semantic overlap between clusters. In the first stage, clustering is performed based on vector distances in the embedding space, while in the second stage, a Large Language Model (LLM) refines these clusters by leveraging open-vocabulary descriptions. The final output assigns each behavior cluster a primary descriptive label.

In the initial clustering stage, distances between data points are calculated using cosine similarity, which measures the angular difference between vectors in the embedding space. These distances are then used as input to the Agglomerative Hierarchical Clustering algorithm from SciPy (using \textbf{scipy.cluster.hierarchy.linkage}), constructing a dendrogram that provides a hierarchical structure of data points. The dendrogram helps determine how to iteratively merge clusters, using the shortest inter-cluster distance as a measure.

We define two critical hyperparameters: the target number of clusters $\theta_c$ and the minimum number of elements per cluster 	$\theta_e$. We employ the \textbf{fcluster} algorithm from SciPy with \texttt{criterion='maxclust'} to cut the dendrogram at a trial cluster count. Clusters with fewer than $\theta_e$ elements are assigned the label $-1$. The trial cluster count is increased until at least $\theta_c$ clusters satisfy the minimum-size criterion.

In the refinement stage, we use an LLM to enhance the quality and semantic coherence of the clusters. First, the most frequently occurring keywords within each cluster are extracted, and the top five are retained as a preliminary label for the cluster. The LLM then evaluates whether similar clusters should be merged, based on their labels, to improve interpretability and ensure consistency in behavioral characterization. The LLM makes the final decision on merging clusters, allowing a flexible and data-driven adjustment of cluster boundaries.

This two-stage clustering method facilitates nuanced major behavior profiling by creating behaviorally meaningful groups, while leveraging natural language descriptions to improve cluster coherence and accuracy. It allows researchers to explore both overarching patterns and subtle distinctions in mouse behavior, contributing to a more complete understanding of behavioral phenotypes (Extended Fig.~\ref{fig:ef6}a,b).

\begin{algorithm}
\caption{Two-Stage Behavioral Clustering with LLM Refinement}
\begin{algorithmic}[1]
\STATE \textbf{Input:} 
    \begin{itemize}
        \item Description embedded vector list: $V = \{v_1, v_2, \dots, v_n\}$
        \item Description keyword list: $T = \{t_1, t_2, \dots, t_n\}$
    \end{itemize}

\STATE \textbf{Hyperparameters:}
    \begin{itemize}
        \item Desired number of clusters: $\theta_c$
        \item Minimum number of elements per cluster: $\theta_e$
    \end{itemize}

\STATE \textbf{Initialization:}
    \begin{itemize}
        \item Cluster ID for each frame: $C$
        \item Keyword list for each cluster: $K$
        \item Construct linkage matrix: $lm = \texttt{construct\_linkage\_matrix}(V)$
        \item $current\_cluster\_num \gets 0$
        \item Trial cluster count: $q \gets \theta_c$
    \end{itemize}

\WHILE{$current\_cluster\_num < \theta_c$}
    \STATE $C \gets \texttt{fcluster}(lm, q, \texttt{criterion='maxclust'})$
    \STATE $cluster\_elem\_num \gets \texttt{unique}(C)$
    \FOR{each cluster $i$ in $cluster\_elem\_num$}
        \IF{$\texttt{len}(C[C == i]) < \theta_e$}
            \STATE $C[C == i] \gets -1$
        \ENDIF
    \ENDFOR
    \STATE $current\_cluster\_num \gets \texttt{len}(\texttt{unique}(C[C \ne -1]))$
    \IF{$current\_cluster\_num < \theta_c$}
        \STATE Increase $q$ for the next cut.
    \ENDIF
\ENDWHILE

\STATE $K \gets \texttt{union\_keywords}(T, C)$
\STATE $C \gets \texttt{LLM\_merge}(K, C)$
    
\STATE \textbf{Output:} Final cluster assignments $C$

\end{algorithmic}
\end{algorithm}

\begin{figure}[!htbp]
\setcounter{figure}{5}
\renewcommand{\thefigure}{\arabic{figure}}
\renewcommand{\theHfigure}{extended.\arabic{figure}}
\renewcommand{\figurename}{Extended Fig.}
\centering
\includegraphics[width=\textwidth,height=0.8\textheight,keepaspectratio]{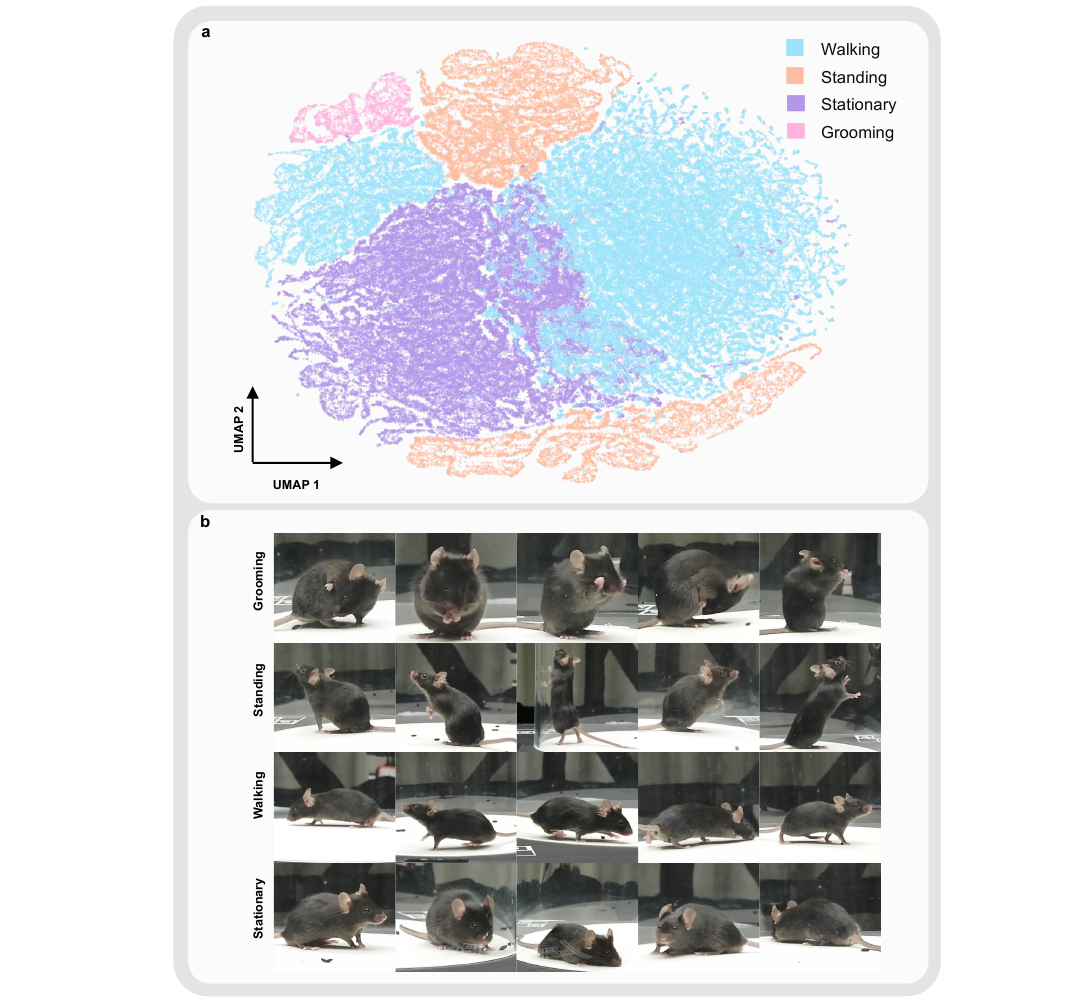}
\caption{\textbf{Semantic embedding distribution and examples of major behaviors.}
\textbf{a}, Distribution of major behaviors in MouseGPT description embeddings: UMAP visualization of behavioral description embeddings from the hallucinogen experiment dataset, colored by the major behavior categories summarized by MouseGPT.
\textbf{b}, Representative images of major behavior categories: Representative examples of each behavior category.}
\label{fig:ef6}
\label{fig:exfig-umap_sample}
\end{figure}

\paragraph{Fine-Grained Behavior Analysis - LLM-Guided keyword matching}

We developed an LLM-guided keyword-matching method to effectively quantify and summarize fine-grained behavioral descriptions of mice. This approach allows for detailed tracking of body part movements, capturing subtle variations in behavior that are essential for comprehensive behavior analysis.

For this purpose, we maintain a Python dictionary for each key body part: ``Torso'', ``Head'', and ``Limb''. Each dictionary consists of predefined behaviors (as keys) and lists of matching keywords or phrases (as values). For instance, in the ``Torso'' dictionary, keys such as ``upright'', ``horizontally stretched'', and ``curved'' categorize different postures, with corresponding values detailing specific descriptions of these behaviors.

To perform fine-grained behavior analysis, we process the frame-by-frame textual descriptions and attempt to match each description with the dictionary entries using regular expression matching. If a match is found, we annotate the behavior for that frame using the corresponding key from the dictionary. For any behavior descriptions that do not match existing entries, we pass the description to the LLM, which either assigns it to an existing key (if semantically similar) or generates a new key, thus updating the dictionary dynamically. This enables continuous learning and expansion of the behavior lexicon as more nuanced behaviors emerge.

By counting the occurrences of these keys across video frames, we can quantitatively track specific behaviors over time, providing a comprehensive and dynamic analysis of fine-grained behaviors throughout the video sequence (Extended Fig.~\ref{fig:ef7}a--c).

\begin{algorithm}
\caption{Fine-grained Behavioral Classification for Body Parts}
\begin{algorithmic}[1]
\STATE \textbf{Input:} 
    \begin{itemize}
        \item Description list $T = \{t_1, t_2, \dots, t_n\}$
    \end{itemize}
\STATE \textbf{Initialization:} 
    \begin{itemize}
        \item Python Dictionary for each body part: $D_{\text{Head}}$, $D_{\text{Torso}}$, $D_{\text{Limb}}$
        \item Fine-grained behavioral description for each frame: $F = \{F_1, F_2, \dots, F_n\}$
    \end{itemize}

\FOR{$D_j \in \{D_{\text{Head}}, D_{\text{Torso}}, D_{\text{Limb}}\}$}
    \FOR{$t_i \in T$}
        \STATE $is\_matched \gets \textbf{False}$
        \FOR{$key_k \in D_j.keys$}
            \FOR{$value_v \in D_j[key_k]$}
                \IF{\texttt{regular\_expression\_matching}($value_v$, $t_i$) \textbf{is True}}
                    \STATE Append $key_k$ to $F_i$
                    \STATE $is\_matched \gets \textbf{True}$
                \ENDIF
            \ENDFOR
        \ENDFOR
        \WHILE{$is\_matched = \textbf{False}$}
            \STATE $D_j \gets \texttt{ask\_LLM\_to\_update\_dictionary}(t_i, D_j)$
            \STATE Reapply keyword matching to $t_i$ using $D_j$; append matched keys to $F_i$ and set $is\_matched$ accordingly.
        \ENDWHILE
    \ENDFOR
\ENDFOR
\STATE \textbf{Output:} Fine-grained behavioral description list $F$
\end{algorithmic}
\end{algorithm}

\paragraph{Novel Behavior Discovery - Anomaly Detection}

We employ an LLM-based Novel Behavior Discovery method, using anomaly detection algorithms to identify rare or previously unobserved behaviors within a sequence of videos. These rare events, such as a mouse jumping in just 1 or 2 frames in a 10,000-frame-length video sequence, can reveal subtle behavioral patterns that human observers are likely to miss and other machine learning methods may overlook due to noise. For this, we applied the Isolation Forest algorithm to the embedded vectors representing behavior descriptions of each frame.

The Isolation Forest algorithm works by randomly sub-sampling vectors and constructing an ensemble of isolation trees. Each tree is trained on a subset of vectors, and the algorithm isolates rare behaviors more effectively through these random splits. As these behaviors are distinct, they are more likely to appear as isolated nodes within the trees. Rare behaviors are thus identified by detecting isolated samples, where behaviors with fewer splits indicate a lower frequency.

We tuned two critical hyperparameters: the number of estimators and the contamination rate, which together control the sensitivity of the model to rare events. In our implementation, we set these parameters to 100 and 0.0001, respectively, balancing the detection sensitivity and the noise level to achieve optimal results. By using this approach, we ensure that rare but significant behavioral anomalies are captured, facilitating novel behavior discovery and advancing our understanding of nuanced behavioral phenomena in mouse models.

\paragraph{Behavioral Phenotype Prediction}

Our Behavioral Phenotype Prediction framework utilizes a large language model (LLM) to analyze behavioral reports of mice, enabling the determination of the administered drugs with high precision (Supplementary Fig.~2).

First, we constructed a ``knowledge base'' for the LLM, consisting of two components that provides essential background knowledge for effective behavioral phenotype analysis. The first component is the Expert Experience Database, containing academic materials and reports written by experts based on existing knowledge and statistical data of characteristic behavioral changes in mice following specific drug injections or particular experimental procedures, ensuring the reliability of this information. 

The second component is the Multi-layer Behavior Report documenting behaviors observed in a sample mouse for each experimental condition, such as drug injections, collected during controlled experimental sessions. This table includes Major Behavior Profiling results like walking and rearing, as well as fine-grained behavior details such as instances of head-up movements during walking. Also, a report documenting novel behavior happened during the experiment is presented. The Behavior reports, together with the Expert Experience Database, are provided to the LLM as system prompts.

Next, we supply the LLM with another Multi-layer Behavior Report of a test subject under specific experimental conditions, and prompt it to infer the drug administered based on the knowledge base, using its ability to interpret behavioral patterns and match them with existing data in the knowledge base. During this analysis, the LLM extracts relevant data from the database and generates python code (see Natural Language User Interface) to identify the mouse with the closest behavior distribution in the knowledge base of each behavioral phenotype. When assessing behavioral similarities, the LLM calculates scores to determine the closest matching behavioral category, dynamically adjusting the weighting of various behaviors based on insights from the expert report.

\paragraph{Search}

We designed a natural language search method that allows users to locate specific behaviors in video sequences, leveraging the subtle nuances that natural language descriptions provide compared to single-word labels. This search approach utilizes fuzzy matching based on embedding distances, ensuring that even slight variations in behavior can be identified.  Users can describe behaviors of interest by inputting either a single word, like ``rear up'', or a detailed phrase such as ``mouse standing on hind legs''.

We embed the user's input into a high-dimensional vector using the same embedding model, text-embedding-3-large, and calculate cosine similarity with each frame's embedded vector. Frames corresponding to the top N vectors with the highest similarity scores are returned, ensuring users receive the most relevant results. This approach leverages natural language descriptions to intuitively match and locate the desired frames, enhancing the user experience.

\subsection{Behavior analysis and statistics}
All statistical analyses of hallucinatory or schizophrenia-related behaviors were completed on GraphPad Prism 8.0.  We selected mice treated with Saline, 2 mg/kg Lysergic acid diethylamide (LSD), 0.3 mg/kg MK-801, and 3 mg/kg Psilocybin (n=5,4,5,6 respectively) from the dataset to showcase differences in hallucinogen-induced behavioral phenotypes. The hallucination dataset also includes three groups that did not participate in the statistical analysis: 1mg/kg Psilocybin (n=4), 10mg/kg Psilocybin (n=2), and 0.2mg/kg LSD (n=5). For each mouse, MouseGPT processed 6000 images at a time resolution of 5 fps to identify behavioral categories. After obtaining the categories of primary behaviors and posture-based behavior subtype, we calculated the proportion of various main behaviors within 20 minutes for each mouse, as well as the proportion of fine behaviors within walking behavior. Two-way ANOVA was used to test the significance of differences. For the temporal and spatial distribution corresponding to each type of behavior, we used custom Python scripts for plotting. To allow MGPT to learn the behavioral phenotypes of various drugs and make predictions for given mice, we randomly selected one mouse from each category and input its behavioral results into the model, then had the remaining mice predict the category they might belong to.
\begin{figure}[!htbp]
\setcounter{figure}{6}
\renewcommand{\thefigure}{\arabic{figure}}
\renewcommand{\theHfigure}{extended.\arabic{figure}}
\renewcommand{\figurename}{Extended Fig.}
\centering
\includegraphics[width=\textwidth,height=0.8\textheight,keepaspectratio]{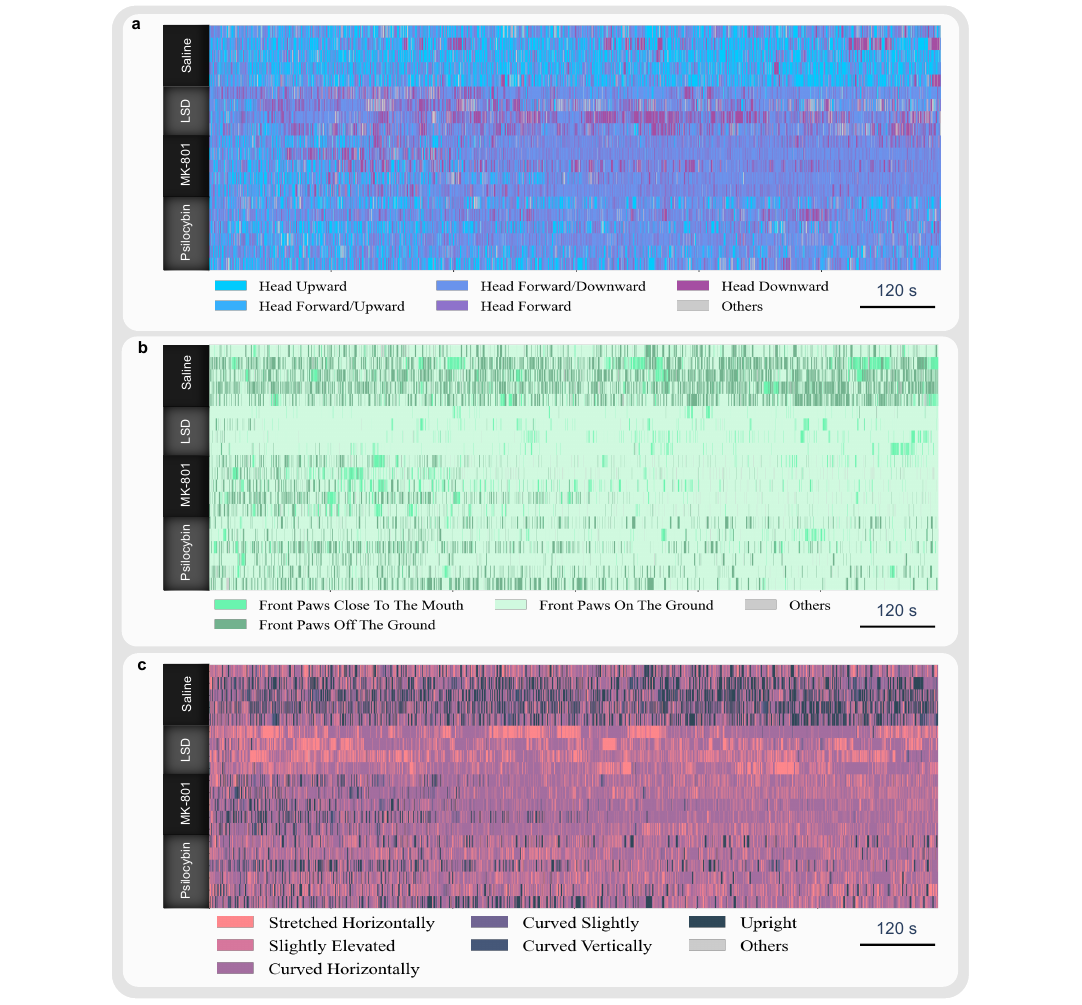}
\caption{\textbf{Temporal profiles of fine-grained postures across treatment groups.}
\textbf{a}, Temporal profiles of head postures across treatment groups: Ethograms showing head posture features at each time point throughout the 20-minute recording for all mice in the Saline, LSD, MK-801 and Psilocybin groups.
\textbf{b}, Temporal profiles of forepaw states across treatment groups: Ethograms showing limb posture features at each time point throughout the 20-minute recording for all mice in the four groups.
\textbf{c}, Temporal profiles of torso postures across treatment groups: Ethograms showing torso posture features at each time point throughout the 20-minute recording for all mice in the four groups.}
\label{fig:ef7}
\label{fig:exfig-ethogram_sample}
\end{figure}

\paragraph{Evaluation of Behavioral Phenotype Prediction}

We evaluated the Behavioral Phenotype Prediction results by comparing the predictions made by the model with the actual behavioral phenotypes of the mice. Since the output format from the LLM is not predefined, the model typically generates text representing its prediction, which we then interpret and standardize for consistency. We adjusted the format of the LLM’s output to align with the correct prediction format, ensuring uniformity. In cases where the model encountered runtime errors (e.g., failure to generate code or incomplete output), we re-executed the process. To assess the model’s performance, we computed the accuracy and F1-score based on the predictions from the selected test set, providing a quantitative evaluation of the model’s ability to accurately predict the behavioral phenotypes.

\subsection{Natural Language User Interface}

Our natural language user interaction is facilitated through an Agent that integrates multiple Vision-Language Models and a customizable toolset. Users can seamlessly perform tasks like mouse behavior analysis, including search, clustering, and more, all through natural language commands. Additionally, the Agent can automatically generate and execute code to help users complete statistical analyses (Extended Fig.~\ref{fig:ef8}).

\begin{figure}[!htbp]
\setcounter{figure}{7}
\renewcommand{\thefigure}{\arabic{figure}}
\renewcommand{\theHfigure}{extended.\arabic{figure}}
\renewcommand{\figurename}{Extended Fig.}
\centering
\includegraphics[width=\textwidth,height=0.8\textheight,keepaspectratio]{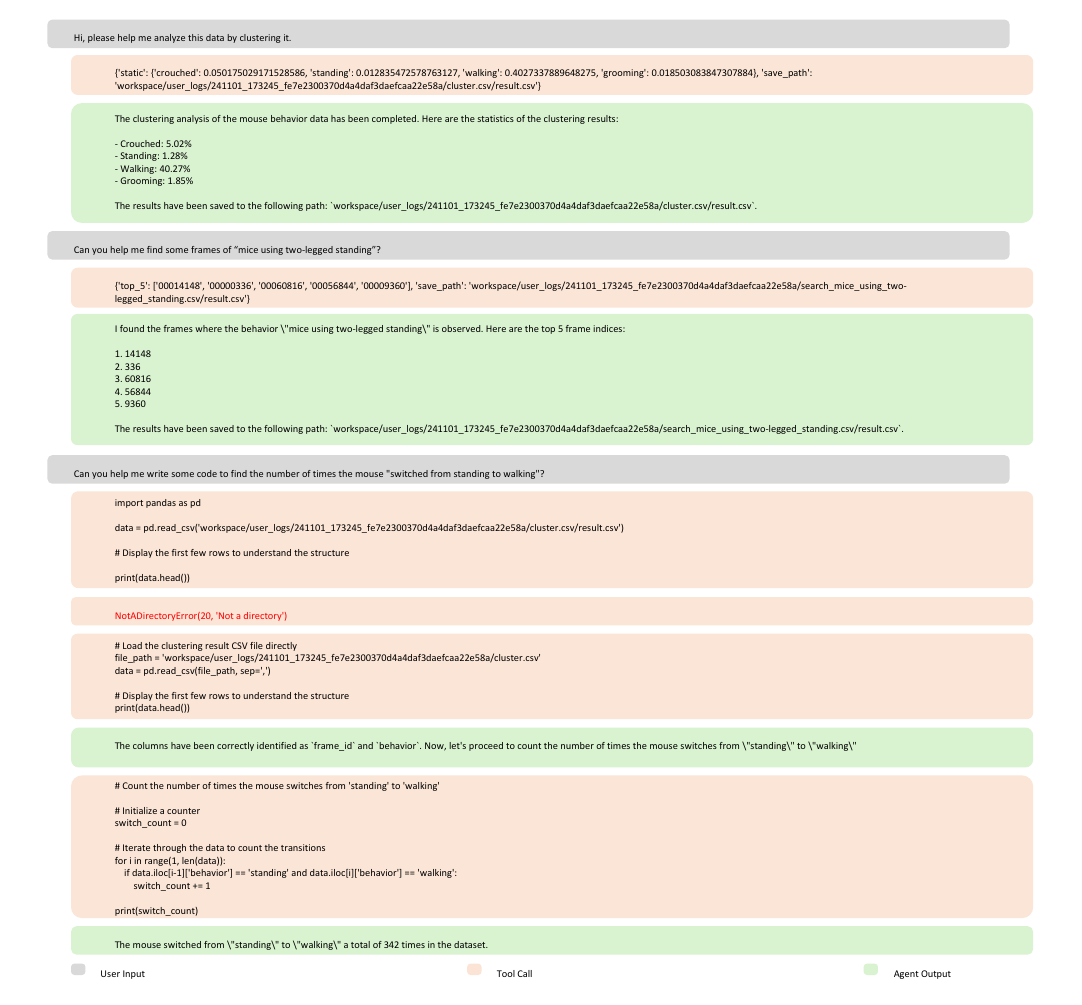}
\caption{\textbf{Example interactions with the MouseGPT agent.}
An example of a natural-language interaction demonstrating the capabilities of the MouseGPT agent. The dialogue shows user requests for behavior-specific frame retrieval, clustering analysis and code generation for analyzing behavioral transitions. The agent effectively combines tool invocation, data processing and execution to provide results that support subsequent operations, illustrating its versatility and ease of use for mouse behavior analysis tasks.}
\label{fig:ef8}
\label{fig:exfig-dialogue_sample}
\end{figure} 

We build our Agent on the LangChain~\cite{LangChain}, a framework designed for building applications that integrate large language models as agents with external tools. 
The Agent processes complex user commands by breaking them down into executable steps, invoking the appropriate tools or models in sequence. This approach allows users to interact with the system intuitively, while the Agent efficiently manages the underlying complexity.

The system prompt used to create the Agent is shown below:
\begin{lstlisting}[language=Python, caption=Python Example]
instructions = """You are a research assistant specializing in mouse behavior analysis. Your next task is to assist me in analyzing a mouse video using the available tools.

You have access to a Python REPL for executing code. When writing code, begin by reading the necessary files from the database, followed by reviewing the table headers to confirm the structure and content. This will help you avoid issues like key errors.

If you encounter any errors, debug the code and retry until successful.
"""
\end{lstlisting}

\paragraph{Workflow Design}

\begin{enumerate}
    \item \textbf{User Prompt}: At the beginning of each round, the user sends a prompt to the agent through the dialog box. The agent could receive both the prompt and the frame ID that the user is analyzing, which ensures that the agent has context regarding the specific frame.
    \item \textbf{Tool Invocation Decision}: Upon receiving the prompt, the agent analyzes the input to determine whether a tool invocation is required to answer the user's request. If necessary, the agent formulates a tool request, which includes specific parameters such as the frame ID, relevant data, and task requirements, and waits for the tool to return results.
    \item \textbf{Tool Execution and Data Processing}: Once the tool receives the request, it accesses the local data (e.g., video frames or experimental results) and performs the corresponding calculations or analyses. The detailed results generated by the tool are saved as a file for later access. Meanwhile, a summarized version of the results is returned to the agent in text form.
    \item \textbf{Response to User and Logging}: The agent then returns a detailed answer to the user, incorporating the summarized results from the tool. Simultaneously, the detailed results, including tool invocations and intermediate calculations, are logged and presented in the interface’s logging section. This ensures transparency in the workflow, allowing users to review the exact process and results of the tool invocations. 
\end{enumerate}

\paragraph{User Interface} 
We implemented the user interface using Gradio~\cite{abid2019gradio}, creating an accessible and interactive environment for users to engage with the Vision-Language Model (VLM). After importing the pre-processed data, users can utilize our natural language interface to conduct further analysis. The interface is organized into the following main sections: 
\begin{enumerate} 
\item \textbf{Dialog Box.} This section allows the user to interact with the Agent by typing queries. Users can ask for various tasks, such as analyzing a specific data frame or requesting classification for the video. 
\item \textbf{Video Frame Preview and Slider.} Users can navigate video content using the frame preview and slider feature. By dragging the slider, users can select a specific frame for inspection. During the conversation, the Agent automatically detects the currently selected frame. 
\item \textbf{Record of Tool Invocation.} This section provides a clear overview of how the Agent employs various tools, displaying the input and output of each tool as well as intermediate results. Users can also export results, such as clusters and generated code, using the data packaging functions integrated into the interface. 
\end{enumerate}

\paragraph{Tools Integration}

Tools integration allows interaction between the LLM-based Agent and various mouse analysis tools.
We utilize LangChain's \verb|langchain_core.tools| to incorporate the major behavior profiling, novel behavior discovery, search, and classification functions described in the MouseGPT Behavior Analysis Framework subsection into tools that can be invoked by the Agent.

Each tool is meticulously designed with clearly defined input and output formats, as well as variable types, ensuring seamless communication between the Agent and the tools. Additionally, we provide a detailed description for each tool, allowing the Agent to fully understand the tool’s functionality and usage scenarios. To streamline the process, we summarize the results after they are computed by the tools and return these summaries to the Agent. For example, in the classification and clustering functions, we summarize the percentage of occurrences for each behavioral classification. For the search and novel behavior discovery functions, we return the specific frame IDs along with the corresponding behaviors to the Agent.

Here is an example of a description of the tool, used to explain the purpose and input/output format of the tool to the agent:

\begin{lstlisting}[language=Python, caption=Python Example]
@tool
def tool_search(behavior_description: str) -> dict:
    """
    A search tool for behaviour. Based on the behaviour description entered by the user, search for the frame index corresponding to the closest behaviour description among all the data. return the five closest frame indexes, and sort and save the closeness of all the frames.

    Input param:
        - behavior_description: Description of target behaviour

    Output: 
        - a dict
        {
            "top_5": a list with 5 frame index
            "save_path": str, the paths of the searching result
        }
    """
\end{lstlisting}
To meet the users' more specific requirements, such as ``counting the frequency of transitions between certain behaviors'', we integrated the Python toolkit \verb|langchain\_experimental.utilities.PythonREPL|. The Python REPL (Read-Eval-Print Loop) is an interactive programming environment that allows the Agent to input Python code and execute it immediately. This tool enables the Agent to write and execute custom Python code in a virtual environment, providing access to local data and facilitating more complex calculations. The REPL operates by following these steps: it reads a line of Python code from the Agent input, evaluates the code and computes the result, prints the result to the Agent, and repeats the process until the Agent exits, making it ideal for rapid experimentation and debugging. Additionally, the generated code is saved as a file for future modification and reproduction, offering users greater flexibility in their analysis.

The agent can automatically read the local data structure and correct errors in the code. Here we provide an example:  

\texttt{Query: Calculate the frequency of the mouse changing from Standing to Walking.}

\begin{lstlisting}[language=Python, caption=Python Example]
# 1st try:

import pandas as pd

data = pd.read_csv('workspace/user_logs/example_session/cluster.csv/result.csv')
data.head()`
\end{lstlisting}

\texttt{Python output:}

\begin{verbatim}
    NotADirectoryError(20, 'Not a directory')}
\end{verbatim}

\texttt{Agent Responds:}

\texttt{
It seems there was an error because the path provided is not a directory. Let's correct the path and try again. 
I'll read the CSV file and inspect its contents to understand its structure.
}

\begin{lstlisting}[language=Python, caption=Python Example]
# 2nd try:

import pandas as pd

# Correcting the path and reading the CSV file
data = pd.read_csv('workspace/user_logs/example_session/cluster.csv')

# Displaying the first few rows of the dataframe to understand its structure
print(data.head())
\end{lstlisting}

\texttt{Python output}
\begin{verbatim}
   frame_id  behavior
0         0  crouched
1        12  standing
2        24  walking
3        36  walking
4        48  walking
\end{verbatim}

\texttt{Agent Responds:}

\begin{verbatim}
The data has been successfully read.
The structure of the dataframe is as follows:

| frame_id | behavior |
|----------|----------|
| 0        | crouched |
| 12       | standing |
| 24       | walking  |
| 36       | walking  |
| 48       | walking  |

Now, let's calculate the frequency of the mouse changing
from "standing" to "walking".
\end{verbatim}

\begin{lstlisting}[language=Python, caption=Python Example]
# 3rd try:

# Calculate the frequency of transitions from 'standing' to 'walking'

# Shift the behavior column to compare current and next behavior
data['next_behavior'] = data['behavior'].shift(-1)

# Filter the transitions from 'standing' to 'walking'
transitions = data[(data['behavior'] == 'standing') & (data['next_behavior'] == 'walking')]

# Calculate the frequency
frequency = len(transitions)

print(f'The frequency of transitions from standing to walking is: {frequency}')
\end{lstlisting}

\texttt{Python output}
\begin{verbatim}
    The frequency of transitions from standing to walking is: 342
\end{verbatim}

\texttt{Agent Responds:}
\begin{verbatim}
    The frequency of transitions from "standing" to "walking"
    is: **342**.
If you need any further analysis or have additional questions,
feel free to ask!
\end{verbatim}

\subsection{Comparative Studies}
\paragraph{Comparison with other language models}
We compared MouseGPT-Large, and MouseGPT-Lite, with InternVL2, MiniCPM, and GPT-4o, using a carefully selected test set. 

First, we performed inference on all test sets using each model. Then, we involved 15 senior researchers in animal behavior to evaluate the results based on human expertise. The evaluation was conducted through two methods, with the reviewers being blind to the identity of the models being reviewed. The first method involved a \textbf{correctness evaluation} (Supplementary Fig.~1a), where researchers were shown an image, a prompt containing speed information, and the model's output. They were asked to assess whether the output was ‘correct’ or ‘incorrect’ based on three criteria: (1) whether there were factual errors, (2) whether the formatting adhered to the specified requirements, and (3) whether there was unnecessary repetition.

The second method involved a \textbf{side-by-side model comparison} (Supplementary Fig.~1b), where the reviewers were provided with the inputs to the models and the outputs from two different models, selecting the better result based on (1) factual correctness and formatting, (2) whether the terminology used matched industry standards, and (3) the level of detail in describing the behavior.

To ensure the consistency of the evaluation, we provided some repeated questions to different reviewers. The results indicated that the reviewers’ responses were consistent, demonstrating a high level of agreement in their evaluations.

\paragraph{Direct comparison with Keypoint-MoSeq}

To clarify what MouseGPT contributes beyond existing unsupervised behavior-parsing approaches, we performed a head-to-head comparison with Keypoint-MoSeq on the same freely moving mouse dataset. This benchmark was not intended to argue that MouseGPT is universally ``better'' in every respect. Rather, the goal was to identify what additional information becomes available when behavior is represented in a language-aligned semantic space instead of only through kinematic motifs.

We recorded three na\"ive mice for 20 minutes each (total 60 minutes) in an open circular arena. Keypoint-MoSeq was applied first to 3D keypoint trajectories under default settings, producing 20 unlabeled syllables. MouseGPT was then applied directly to the raw videos (using two selected views) to generate frame-level semantic labels. This design enabled a direct comparison of the two pipelines at matched timestamps while preserving each method's native input representation.

\textbf{Semantic coherence relative to human-interpretable behavior.}
Keypoint-MoSeq effectively clustered movement patterns according to kinematic similarity, but the resulting syllables were not necessarily behaviorally homogeneous in semantic terms. The distribution of the 20 syllables was highly uneven, with the largest syllable alone accounting for $31.63\%$ of frames (Extended Fig.~\ref{fig:ef4}a), and several high-occupancy syllables exhibited mixed semantic content. To quantify this, we aligned each Keypoint-MoSeq syllable with MouseGPT frame-level semantic labels and examined the within-syllable label composition. Representative cases showed that syllables treated as a single behavioral unit by Keypoint-MoSeq frequently contained multiple semantic categories: for example, syllables dominated by standing still contained substantial fractions of walking, and walking-dominant syllables often also contained stationary or standing frames (Extended Fig.~\ref{fig:ef4}b). Qualitative inspection of randomly sampled frames from the largest syllable further confirmed this semantic mixing, whereas regrouping those same frames by MouseGPT labels produced behaviorally coherent subsets such as walking, standing, grooming, and stationary behavior (Extended Fig.~\ref{fig:ef4}c). These results indicate that MouseGPT adds semantic coherence that is not guaranteed by kinematics-driven clustering alone.

\textbf{Viewpoint stability.}
We next asked whether labels remain stable when the camera viewpoint changes, which is a practical requirement for cross-laboratory use. Using synchronized recordings of the same behavior, we applied Keypoint-MoSeq independently to four kinematic inputs derived from identical time periods: fused 3D keypoints, top-view 2D keypoints, and two single side-view 2D keypoint sets, all under identical algorithmic settings. The resulting syllable distributions changed substantially across viewpoints, and cross-view similarity matrices showed only sparse, inconsistent correspondences without a stable one-to-one mapping (Extended Fig.~\ref{fig:ef4}d,e). In contrast, MouseGPT produced labels within a shared semantic category space independent of viewpoint. Across different viewpoint pairs, the overall behavior distributions remained consistent, the cross-view similarity matrices showed strong diagonal structure, and the corresponding ethograms preserved behavior onset, duration, and transitions over time (Extended Fig.~\ref{fig:ef4}f--h). Thus, whereas kinematics-driven syllables were strongly viewpoint dependent, MouseGPT labels were comparatively stable across views.

\textbf{Human interpretability and functional output.}
The two pipelines also differed in the nature of their outputs. Keypoint-MoSeq outputs single syllable assignments drawn from a closed, automatically learned category set, and behavioral meaning must be assigned post hoc through manual inspection. MouseGPT, by contrast, produces natural-language descriptions that explicitly encode posture, action, and interaction cues and can subsequently be reorganized into different task-relevant categories. In this sense, Keypoint-MoSeq optimizes \emph{kinematic consistency}, whereas MouseGPT optimizes \emph{semantic consistency}. We summarize this distinction at the level of input requirements, intermediate representation, and final analytical output (Supplementary Fig.~3).

Taken together, these analyses show that MouseGPT contributes a language-native semantic layer that complements rather than replaces kinematics-driven unsupervised methods. Relative to syllable-based outputs, the added value of MouseGPT lies in three properties: semantically coherent labeling, stability across viewpoints within a shared category space, and direct human interpretability that facilitates downstream ethological analysis and flexible reorganization by experimental needs.

\begin{figure}[!htbp]
\setcounter{figure}{3}
\renewcommand{\thefigure}{\arabic{figure}}
\renewcommand{\theHfigure}{extended.\arabic{figure}}
\renewcommand{\figurename}{Extended Fig.}
\centering
\includegraphics[width=\textwidth,height=0.8\textheight,keepaspectratio]{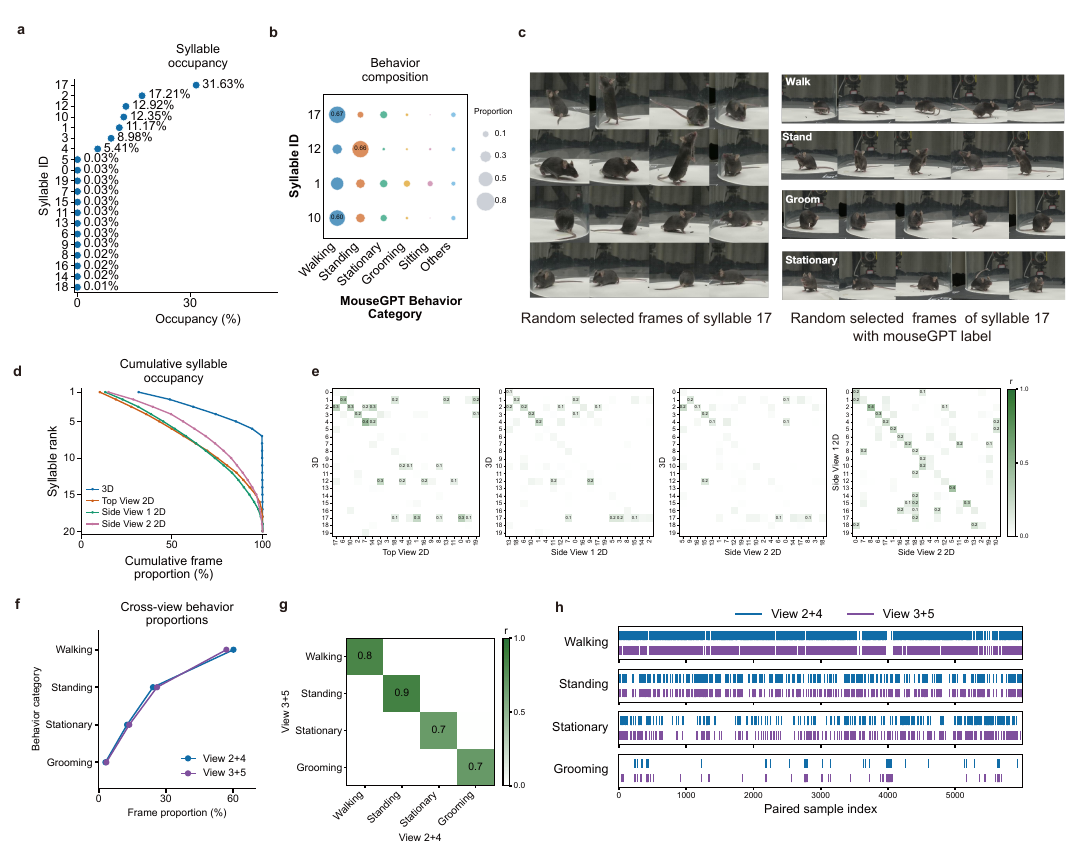}
\caption{\textbf{Evaluation of cross-view stability in MouseGPT behavior classification.}
\textbf{a}, Frame proportions of Keypoint-MoSeq behavioral syllables. Frame proportions of 20 Keypoint-MoSeq syllables derived from three-dimensional keypoint trajectories, ranked by occupancy. A few syllables account for most frames.
\textbf{b}, Semantic behavior composition within individual syllables. MouseGPT semantic labels within major syllables; bubble area denotes the proportion of each behavior category. 
\textbf{c}, Examples from the highest-occupancy syllable: randomly sampled frames showing different behaviors (left) and subsets distinguished by MouseGPT semantic labels (right).
\textbf{d}, Cumulative syllable occupancy for Keypoint-MoSeq applied with identical settings to the same video using fused three-dimensional, top-view two-dimensional or side-view 1/2 two-dimensional keypoints. Syllables are ranked within each condition; distributions differ across viewpoints.
\textbf{e}, Behavioral syllable correlations across viewpoints. Pearson correlations of syllable time series for 3D--top, 3D--side 1, 3D--side 2 and side 1--side 2 comparisons. Following the vertical-axis order, each syllable is matched to the highest-correlated unmatched syllable in the other condition, reordering the horizontal axis. Sparse high correlations indicate limited cross-view correspondence.
\textbf{f}, MouseGPT behavior proportions for View 2+4 and View 3+5 show similar category distributions.
\textbf{g}, Semantic behavior correlations across viewpoint pairs. Pearson correlations of semantic label sequences between the viewpoint pairs. Explicit semantic categories align the axes directly; the diagonal indicates agreement for corresponding behaviors.
\textbf{h}, Semantic behavior time series across viewpoint pairs: Ethograms show preserved behavior onset, duration and transitions across viewpoint pairs.}
\label{fig:ef4}
\label{fig:reb_fig_sup10}
\label{fig:reb_fig_sup11_1}
\label{fig:reb_fig_sup11_2}
\end{figure}

\subsection{Evaluation and Benchmark Protocols}

\paragraph{Expert-verified benchmark of behavioral descriptions}
\label{methods:description_benchmark}

\textbf{Benchmark dataset and reference annotations.}
We constructed a benchmark to evaluate the factual content and task adherence of vision--language descriptions of mouse behavior (Extended Fig.~\ref{fig:ef3}). The benchmark comprised 435 manually selected frames from 25 recordings, spanning routine behaviors, fine postural details and less-frequent, challenging postures. Each evaluation input contained two synchronized $512\times512$-pixel JPEG camera views, supplied at high image detail, and a frame-specific prompt with kinematic information. All inputs used the same system instructions requesting six output fields. Supplied velocities were treated as auxiliary evidence; scalar speed alone was not sufficient evidence of locomotion. The camera images, their order and the velocity-bearing prompt were held fixed across model settings for each frame. This deliberately selected set was intended to assess descriptive performance across varied examples, rather than estimate the prevalence of behaviors in natural recordings.

Reference annotations were developed through AI-assisted drafting, reconciliation and checking with GPT-6 Astra~\cite{openai2026gpt6}. Initial drafts used the images and prompt information without archived descriptions or ratings. A subsequent reconciliation pass incorporated available archived expert evidence and descriptions, and a separate checking pass assessed the proposed reference against the inputs without archived candidate descriptions or ratings. Experts in mouse behavior then reviewed, corrected and verified the references. The references comprised six semantic fields (Overall, Head, Limb, Torso, Others and Keywords) and a set of individual factual statements. Each fact was assigned an identifier, semantic field, required or optional status, and supporting evidence; uncertainty and accepted alternative descriptions were recorded where applicable. Required facts defined the information expected in an adequate description. Optional facts provided additional valid observations without making their omission an error. The verified reference version and evaluation input were fixed for scoring and linked to the corresponding model output and assessment.

\textbf{Model evaluation and automated grading.}
We evaluated eight VLMs across 14 settings: GPT-4o~\cite{openai2024gpt4o} and GPT-4.1~\cite{openai2025gpt41}; GPT-5~\cite{openai2025gpt5} with low and high requested reasoning effort; and GPT-5.6 Sol, GPT-5.6 Luna~\cite{openai2026gpt56}, GPT-6 Sol, GPT-6 Luna and GPT-6 Astra~\cite{openai2026gpt6}, each with low and max requested reasoning effort. Output-token limits were 8,192 for GPT-4o, GPT-4.1 and all low-effort settings, and 32,768 for GPT-5 high and the five max-effort settings. No reasoning-effort parameter was specified for GPT-4o or GPT-4.1. The comparison included a fresh GPT-4o evaluation on the same benchmark inputs. Each setting yielded one saved description for every frame, giving 6,090 descriptions in total. Provider-returned model identifiers and generation settings were retained with the outputs; more-specific dated model snapshots were not exposed in these records. Temperature was not explicitly set. Because both reasoning effort and output-token limits differed between some settings, these comparisons do not isolate the effect of reasoning effort alone.

All descriptions were assessed using one fixed GPT-6.1 Sol~\cite{openai2026gpt61sol} judge configuration (medium reasoning effort; 32,768-token output limit) and a common rubric. The judge received the same two images and input prompt, the expert-verified reference annotations and the candidate description; candidate model identifiers were not supplied to the judge. The judge decomposed each description into distinct claims and recorded each claim's text, semantic field, correctness grade, evidence rationale, links to reference facts, and error type and severity. It separately graded coverage of every required reference fact, assigning zero to omissions. Optional facts were excluded from the coverage denominator but assessed for correctness when asserted. Supported observations beyond the listed reference facts were permitted. Repeated propositions, including repetitions in Keywords, were counted once; novel keyword claims were assigned to their corresponding semantic field. Assessments were checked against a structured schema, and numerical scores were computed deterministically from the saved grades.

\textbf{Rubric dimensions and MouseGPT Benchmark Index.}
The rubric evaluated factual correctness ($P$), required-fact coverage ($R$), output structure ($S$) and instruction adherence ($A$). Claim correctness and fact coverage were graded on the scale $\{0,0.25,0.5,0.75,1\}$, with zero indicating no correctness or coverage credit and one indicating full credit. For frame $i$, let $c_{ij}$ denote the correctness grade of distinct claim $j$, and $r_{ik}$ the coverage grade of required reference fact $k$. Semantic-field weights were 2 for Overall, 1 for Head, Limb and Torso, and 0.5 for Others. Denoting the corresponding claim and fact weights by $w_{ij}$ and $v_{ik}$, respectively, we calculated
\begin{equation}
P_i=\frac{\sum_j w_{ij}c_{ij}}{\sum_j w_{ij}},\qquad
R_i=\frac{\sum_k v_{ik}r_{ik}}{\sum_k v_{ik}}.
\end{equation}
The correctness denominator included erroneous and unsupported claims; the coverage denominator included all required facts, including those omitted from the description. We set $P_i=0$ when no assessable claim was present. All benchmark references contained required facts. Factual quality was summarized by
\begin{equation}
F_i=\frac{2P_iR_i}{P_i+R_i},
\end{equation}
with $F_i=0$ when $P_i=R_i=0$.

For each description, structure and instruction adherence each received one overall grade on the same quarter-step scale. Structure assessed semantic organization, field placement, consistency and nonredundancy. Instruction adherence assessed the specified JSON format, six-field schema, language, focus on the mouse, evidence constraints and treatment of uncertainty. The MouseGPT Benchmark Index was calculated for each frame as
\begin{equation}
I_i=100F_i\left(0.8+0.1S_i+0.1A_i\right).
\end{equation}
Thus, structural and instruction contributions depended on factual quality and could not yield a positive index when $F_i=0$. Components ranged from 0 to 1 and the index from 0 to 100. We calculated each frame's index before averaging across frames. The weights and grade spacing were rubric-defined choices; the index is not an accuracy percentage, and $F$ is a graded factual score rather than binary-classification F1. Expert verification of references did not constitute a separate calibration of the automated judge against held-out expert scores.

\textbf{Summary statistics and challenging-frame analysis.}
Each frame contributed equally to the reported means. For the rubric scores and overall indices (Extended Fig.~\ref{fig:ef3}d,e), 95\% confidence intervals were obtained from 10,000 recording-cluster bootstrap resamples. In each resample, 25 recording identifiers were drawn uniformly with replacement, all frames from each selected recording were retained with the same multiplicity, and the mean was calculated over the resulting frame entries. The same recording selections were used across settings and metrics. Interval bounds were the 2.5th and 97.5th percentiles of the bootstrap means. Recordings were the resampling units and were not assumed to identify distinct animals. These intervals describe variation under resampling of the included recordings, conditional on the saved descriptions and judgments; they do not capture variability from repeated model generation or grading.

For the challenging-frame analysis (Extended Fig.~\ref{fig:ef3}f), we first averaged each frame's factual $F$ scores across settings within each model, then averaged across the eight models with equal model weights. We ranked frames by this score and selected the lowest 10\%, rounded up to 44 frames, using frame identifier to resolve ties. Every setting was evaluated on this same subset and on all 435 frames. The subset was selected post hoc using the same outputs being summarized, and the all-frame group included the selected frames. Accordingly, this comparison was descriptive rather than an independent validation of performance on a predefined difficult set.

\begin{figure}[!htbp]
\setcounter{figure}{2}
\renewcommand{\thefigure}{\arabic{figure}}
\renewcommand{\theHfigure}{extended.\arabic{figure}}
\renewcommand{\figurename}{Extended Fig.}
\centering
\includegraphics[width=\textwidth,height=0.8\textheight,keepaspectratio]{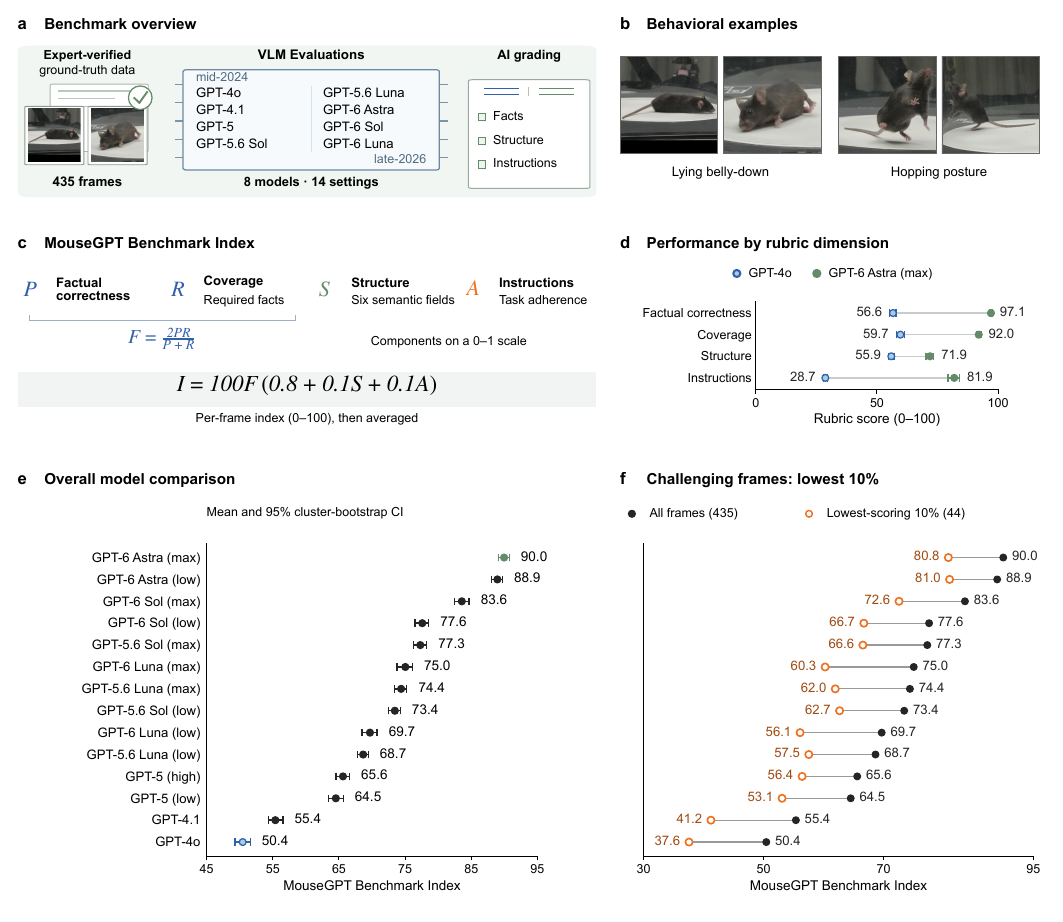}
\caption{\textbf{Benchmarking visual descriptions of mouse behavior.}
\textbf{a}, Benchmark overview: 435 manually selected frames from 25 recordings with expert-verified, AI-assisted references and automated grading. Eight vision--language models were evaluated across 14 settings using the same camera views and velocity-bearing prompts. A fixed GPT-6.1 Sol judge configuration graded all 6,090 descriptions.
\textbf{b}, Examples illustrating the benchmark's behavioral diversity: synchronized views of less-frequent, challenging lying belly-down and hopping postures.
\textbf{c}, Four rubric dimensions: factual correctness ($P$), required-fact coverage ($R$), structure ($S$) and instruction adherence ($A$). Semantic-field-weighted correctness and coverage combine as $F=2PR/(P+R)$, with $F=0$ when $P=R=0$. The MouseGPT Benchmark Index is $I=100F(0.8+0.1S+0.1A)$. Components range from 0 to 1; indices are calculated per frame before averaging. Structure and instruction contributions depend on factual quality.
\textbf{d}, Rubric scores ($\times100$) for the conditions with the lowest (GPT-4o) and highest (GPT-6 Astra, max) mean indices.
\textbf{e}, Mean index across all 435 frames for each condition. Parentheses indicate requested reasoning effort; output-token limits also varied by setting.
\textbf{f}, Mean index for all 435 frames and a shared subset of 44 challenging frames (lowest 10\%, rounded up). Frames were ranked by factual $F$, first averaging settings within each model and then averaging the eight models equally. The same subset was used for every condition. Selection used these same outputs post hoc; the overlapping-group comparison is descriptive, not independent validation.
Dots and numerical labels show means. Whiskers in \textbf{d,e} show 95\% percentile confidence intervals from 10,000 recording-cluster bootstrap resamples: 25 recordings were sampled with replacement, retaining all their frames and calculating frame-weighted means.}
\label{fig:ef3}
\label{fig:mousegpt-benchmark}
\end{figure}

\paragraph{Evaluation on MABe 2022}
\label{sec:eva_mabe}

To evaluate the MouseGPT framework for social behavior analysis, we adapted our pipeline to the MABe 2022 benchmark, which contains low-resolution, top-down recordings of social interactions among three mice (Extended Fig.~\ref{fig:ef5}d). We used GPT-6 Astra, GPT-5.6-Luna, GPT-5.1 and Doubao-1.6-lite for description generation and retained the downstream representation-extraction and behavior-analysis pipeline.

We analyzed video clips, each comprising a continuous sequence of frames from a source recording. For each clip, we sampled up to 15 images from its beginning at six-frame intervals, corresponding to 5 fps for 30-fps input. The images were provided together with a task-specific prompt asking the model to locate the mice, describe each individual and assess nine social behaviors. The model returned a JSON record containing spatial layout, individual descriptions, a scene summary, and a Boolean detection and visual evidence for each behavior. The full JSON records, including the stored source-video and clip identifiers, were serialized as text and embedded using text-embedding-3-large.

Training and test sets were separated by source video using custom split lists. GPT-5.6-Luna and Doubao-1.6-lite shared the full-data video split; GPT-6 Astra and GPT-5.1 were evaluated on the same one-fifth subset of the MABe 2022 dataset. Data coverage for each model is reported in Supplementary Table~2. For each behavior, a clip embedding was paired with positive and negative labels, weighted by the respective numbers of annotated frames within the clip; zero-count entries were omitted. StandardScaler and principal component analysis were fitted on the resulting training data, retaining up to 128 principal components, and applied to the test embeddings. A separate RidgeClassifier was trained for each behavior with alpha = 10, balanced class weights and the frame-count sample weights. At test time, each clip received one prediction per behavior, which was evaluated against the actual positive and negative frame counts in its recorded interval. We calculated the positive-class F1 score for each of the nine behaviors and report their arithmetic mean as Avg F1 (Extended Fig.~\ref{fig:ef5}e,f).

For comparison, we included the published per-behavior F1 scores for AmadeusGPT, BAMS and PCA from Table 1 of Ye et al.~\cite{ye2023amadeusGPT}. AmadeusGPT uses keypoints and natural-language behavior definitions for rule-based task programming, whereas BAMS and PCA provide keypoint-based behavioral representations. These scores were taken from the published evaluations, and their Avg F1 values were calculated as the arithmetic mean across the same nine behavior classes.

\paragraph{Evaluation on Agent Framework}
\label{sec:eva_agent}

As a practical application layer built on top of MouseGPT's semantic representation, rather than the central methodological contribution itself, we quantified the reliability of the natural-language agent using a standardized benchmark (Extended Fig.~\ref{fig:ef9}; Supplementary Table~3). We evaluated 20 user tasks spanning 10 functional categories on five independent datasets, each comprising an approximately 20-min recording, for a total of 100 trials. A trial was counted as successful only when the agent correctly interpreted the request, invoked the appropriate tools, and produced a valid final output. Under this criterion, the agent achieved an overall success rate of 87\% (87/100), and 15 of the 20 tasks succeeded on all five datasets. Failures were concentrated in a small subset of code-heavy tasks, particularly those requiring multi-step quantitative computation, dynamic code execution, or brittle plotting and file input/output, where occasional runtime issues could prevent completion even when the user intent was correctly understood. These results indicate that the agentic interface is already practically useful for routine laboratory workflows, while also highlighting clear directions for future engineering guardrails, including stricter file-schema validation, more robust code templates, and automatic checking and retry for plotting and file export.

\begin{figure}[!htbp]
\setcounter{figure}{8}
\renewcommand{\thefigure}{\arabic{figure}}
\renewcommand{\theHfigure}{extended.\arabic{figure}}
\renewcommand{\figurename}{Extended Fig.}
\centering
\includegraphics[width=\textwidth,height=0.8\textheight,keepaspectratio]{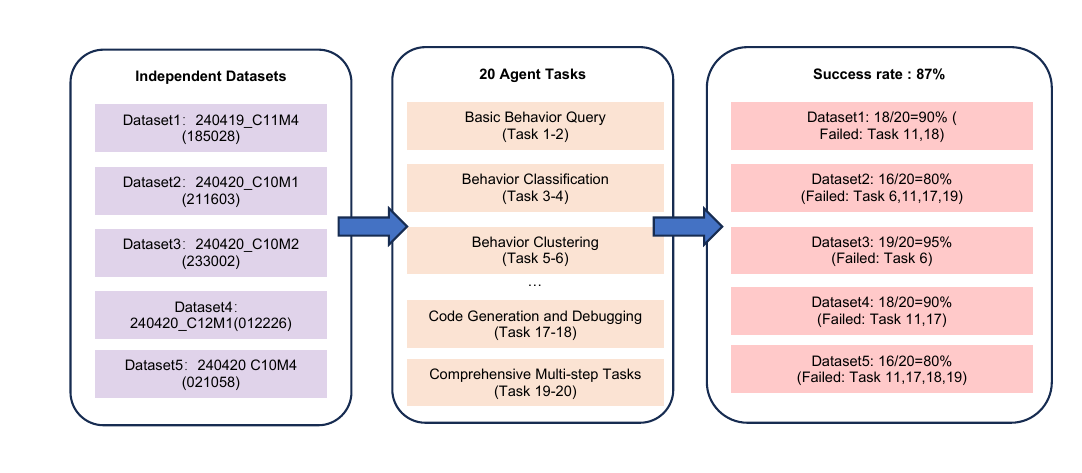}
\caption{\textbf{Evaluation of task execution performance in the MouseGPT agent.}
Schematic of the standardized agent benchmark. Left, five independent evaluation datasets, with dataset identifiers shown. Middle, 20 benchmark tasks spanning 10 functional categories; full prompts are provided in Supplementary Table~3. Right, dataset-level success rates and failed task identifiers. A trial was judged successful upon manual verification that the agent correctly understood the request, invoked appropriate tools and produced a valid final output. Twenty tasks were performed on each of the five datasets, yielding 100 trials and an overall success rate of 87\% (87/100); dataset-level success rates were 90\%, 80\%, 95\%, 90\% and 80\%, respectively.}
\label{fig:ef9}
\label{fig:reb_fig_sup5}
\end{figure}

\paragraph{Validation of Single-Camera Generalizability on Single-Camera Datasets}
\label{sec:eva_single_cam}

To illustrate that multi-camera acquisition is not required for inference, we first performed a qualitative single-camera test using MouseGPT-Lite on representative frames from our in-house dataset. For this demonstration, only one side-view image was used as the visual input, together with the corresponding kinematic summary for the same time point used by the standard inference prompt. No additional fine-tuning, retraining, or architecture modification was introduced for this setting; we used the same inference procedure and the same structured output schema as in the main model, except that the visual input was restricted to a single view (Supplementary Fig.~6). 

To further demonstrate MouseGPT's applicability to standard laboratory workflows, we evaluated MouseGPT-Lite on the public CalMS21 benchmark (NeurIPS 2021) without any fine-tuning (zero-shot). This dataset features top-down, single-camera recordings of social interactions, presenting challenges such as low contrast and frequent occlusions. We processed 10 video sequences (over 30,000 frames) recordings of social interactions. Because the top-down perspective frequently occludes fine limb articulations, we slightly adjusted the output schema, replacing the limb-specific field with an explicit `Interaction' field appropriate for social behavior. A blind panel of 5 independent experts rated the generated descriptions on a stratified random sample of 390 frames. The model achieved a global mean score of 7.93/10 across diverse categories (Attack, Investigation, Mount, Other), confirming strong descriptive accuracy from a single view (Extended Fig.~\ref{fig:ef5}b). 

Furthermore, we conducted a supervised classification benchmark to test the downstream utility of MouseGPT's semantic embeddings compared to standard keypoint-derived kinematic features. Using the official CalMS21 evaluation protocol, a classifier trained on MouseGPT embeddings from only a 10-video subset achieved a held-out test F1 score of 0.8534 (with per-class F1 for Attack: 0.83, Investigation: 0.87, Mount: 0.86). This substantially outperformed the baseline kinematic feature approach (F1 $\approx$ 0.79), which required the full 70-video dataset for training (Extended Fig.~\ref{fig:ef5}c).

\begin{figure}[!htbp]
\setcounter{figure}{4}
\renewcommand{\thefigure}{\arabic{figure}}
\renewcommand{\theHfigure}{extended.\arabic{figure}}
\renewcommand{\figurename}{Extended Fig.}
\centering
\includegraphics[width=\textwidth,height=0.8\textheight,keepaspectratio]{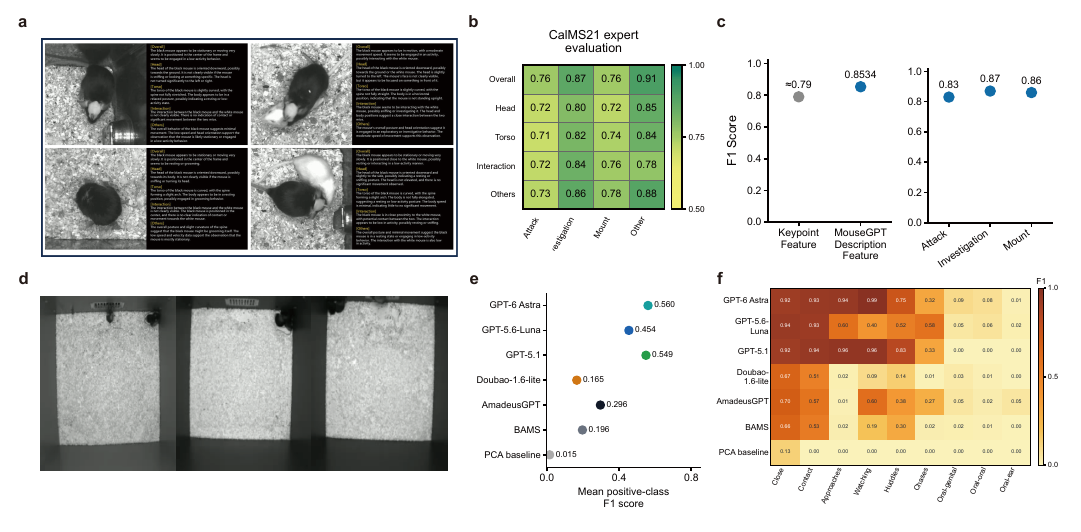}
\caption{\textbf{Evaluation of MouseGPT language-based representations on public social behavior datasets.}
\textbf{a}, Example single-camera CalMS21 images and MouseGPT-Lite descriptions of single-animal activity (left) and social interaction (right).
\textbf{b}, Expert evaluation of description quality on CalMS21. Five independent experts rated descriptions of 390 frames sampled using existing labels (Attack, Investigation, Mount and Other). Evaluation dimensions were Overall, Head, Torso, Interaction and Others, with Interaction added for this social setting. Heatmap values are mean 1--10 ratings normalized to [0,1]; the overall mean was 7.93/10.
\textbf{c}, Supervised classification evaluation of semantic representations on CalMS21. The CalMS21 supervised classification protocol compares keypoint features with semantic embeddings of MouseGPT descriptions. Left, test-set macro F1; right, behavior-specific F1 for MouseGPT (Methods). Semantic embeddings yield a higher macro F1.
\textbf{d}, Example video frames from MABe 2022 showing interactions among multiple mice, recorded from above at lower resolution, providing less visual detail.
\textbf{e}, Mean F1 scores across methods on MABe 2022. The MABe 2022 evaluation compares semantic representations generated within MouseGPT using different general-purpose vision-language models with published results for AmadeusGPT (language-based analysis) and BAMS/PCA (behavior representations). Mean positive-class F1 across nine behaviors is shown; settings are detailed in Methods and Supplementary Table~2.
\textbf{f}, F1 scores for individual behavior classes on MABe 2022. Behavior-specific F1 for the methods in e (rows, methods; columns, behaviors). GPT-6 Astra and GPT-5.1 settings score higher on Close, Contact, Approaches and Watching, and lower on Oral-genital, Oral-oral and Oral-ear.}
\label{fig:ef5}
\label{fig:reb_fig_sup1}
\label{fig:reb_fig_sup8}
\end{figure}

\paragraph{Validation of Pose-Estimator Compatibility with DeepLabCut (DLC) and SLEAP}
\label{sec:eva_pose_est}

To directly address the concern that MouseGPT might depend on a proprietary upstream estimator or multi-camera setups, we tested whether the kinematics module could be replaced by community-standard pose trackers without materially changing downstream interpretation. MouseGPT was designed from the outset to accept time-stamped keypoint coordinates as a modular input; accordingly, we implemented native support for DeepLabCut (DLC) and SLEAP as interchangeable upstream pose sources.

We fine-tuned DLC and SLEAP on our data and evaluated them on the same held-out test set used for our native pose pipeline, treating the native multi-view estimate as a silver-standard reference. We then performed a two-stage validation. First, we quantified signal fidelity by comparing per-keypoint instantaneous-speed signals derived from DLC or SLEAP against the silver standard using Pearson correlation (agreement in temporal dynamics) and RMSE (absolute magnitude error). DLC showed strong kinematic agreement overall (mean correlation $=0.9435$, mean RMSE $=2.47$ cm/s), whereas SLEAP also produced compatible signals (mean correlation $=0.7738$, mean RMSE $=6.24$ cm/s), with several individual keypoints reaching very high agreement. Importantly, the keypoints used by the MouseGPT kinematics module represent only a small, anatomically standard subset of the full skeleton, indicating that the pipeline does not depend on an unusually dense or proprietary keypoint definition (Extended Fig.~\ref{fig:ef2}a,b).

Second, because practical adoption depends on whether behavior interpretation changes when the pose source changes, we quantified semantic stability under pose-source swapping. For each frame, we generated MouseGPT descriptions using kinematics derived from DLC or SLEAP and compared them against descriptions generated from the native pose source. Paired descriptions were then assigned to one of four discrepancy levels using an LLM-based semantic comparison: L0, semantically equivalent; L1, minor mismatch in speed or intensity wording; L2, major speed contradiction; and L3, action or behavior mismatch. For DLC versus native pose, the semantic stability score (L0+L1) was $90.4\%$ ($N=5{,}992$), with L0 $=51.0\%$ (3057/5992), L1 $=39.4\%$ (2359/5992), L2 $=8.2\%$ (491/5992), and L3 $=1.4\%$ (85/5992). For SLEAP versus native pose, the semantic stability score was $87.5\%$ ($N=5{,}988$), with L0 $=49.2\%$ (2944/5988), L1 $=38.3\%$ (2295/5988), L2 $=11.2\%$ (673/5988), and L3 $=1.3\%$ (76/5988) (Extended Fig.~\ref{fig:ef2}c).

\begin{figure}[!htbp]
\setcounter{figure}{1}
\renewcommand{\thefigure}{\arabic{figure}}
\renewcommand{\theHfigure}{extended.\arabic{figure}}
\renewcommand{\figurename}{Extended Fig.}
\centering
\includegraphics[width=\textwidth,height=0.8\textheight,keepaspectratio]{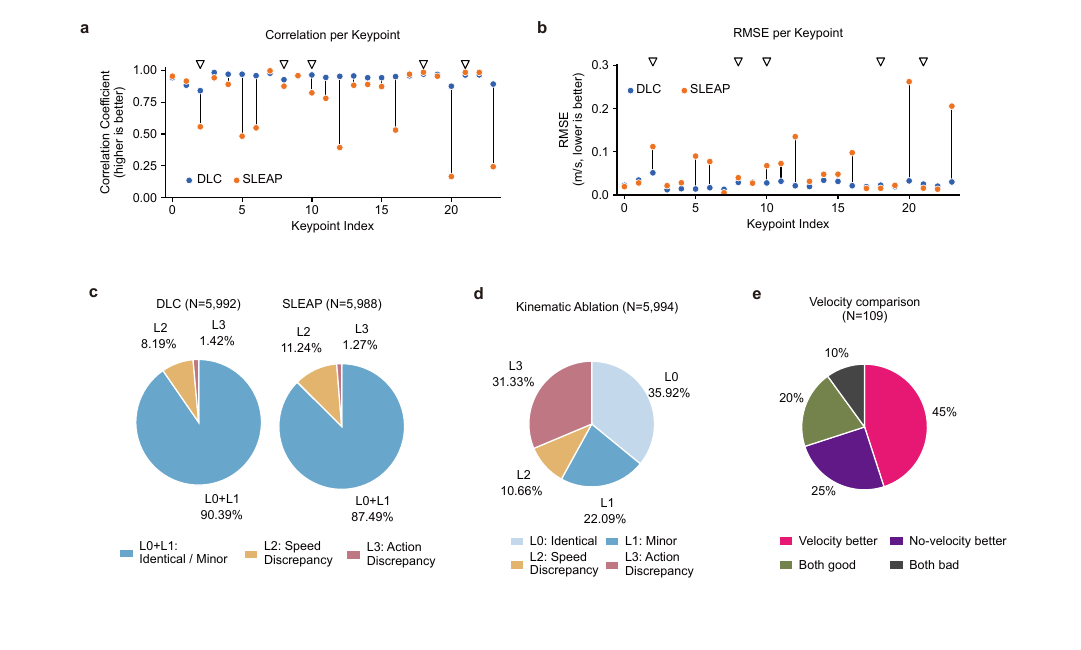}
\caption{\textbf{Evaluation of cross-tool compatibility and the role of kinematic input in MouseGPT.}
\textbf{a}, Pearson correlations between instantaneous keypoint speeds derived from DeepLabCut (DLC, blue) or SLEAP (orange) and those from MouseGPT’s built-in pose estimator. Black line segments indicate the difference between the two tools’ correlation coefficients for each keypoint. The keypoints used to calculate MouseGPT’s kinematic inputs—the snout tip, mid-tail, left forepaw tip and both hindpaw tips—are marked by open inverted triangles (see Supplementary Table 1 for keypoint definitions).
\textbf{b}, Root mean square error (RMSE, m/s) of DLC- and SLEAP-derived keypoint speeds relative to those from the built-in pose estimator. Colors and symbols are as in a.
\textbf{c}, Kinematic inputs derived from the built-in pose estimator were replaced with those derived from DLC or SLEAP. A large language model (LLM) judge evaluated differences between behavioral descriptions generated for the same samples before and after replacement (N = 5,992 and 5,988, respectively). Differences were classified as L0 (semantically equivalent), L1 (minor differences without changing the core behavior), L2 (speed or intensity differences) or L3 (action differences). Descriptions remained semantically equivalent or differed only slightly in 90.39\% and 87.49\% of samples following replacement with DLC and SLEAP inputs, respectively (L0+L1, shown together).
\textbf{d}, Behavioral descriptions generated by MouseGPT-Lite with and without kinematic information in the prompt were compared on the same test set (N = 5,994). The LLM judge classified differences using the L0–L3 categories defined in c. Removing kinematic information changed speed or action descriptions in 41.99\% of samples (L2+L3), indicating that kinematic input influences how the model describes movement speed and actions.
\textbf{e}, Human evaluation of 109 samples randomly selected from the L2/L3 cases in d. Raters viewed the original images and judged whether descriptions with speed information were better (pink), descriptions without speed information were better (purple), both were good (olive green) or both were poor (dark gray).}
\label{fig:ef2}
\label{fig:reb_fig_sup9}
\label{fig:reb_fig_sup4}
\end{figure}

Thus, for both DLC and SLEAP, the vast majority of frames preserved the same action-level interpretation, whereas genuine behavior mismatches were rare. Together, these results show that MouseGPT is pose-source agnostic in practice: the kinematics module is explicitly swappable, and the downstream semantic analysis remains robust when the input pose trajectories are provided by community-standard tools rather than by our native pipeline. This compatibility is important for adoption, because it means MouseGPT can function as a practical add-on to existing single-camera analysis workflows instead of requiring a custom capture or estimation stack.

\FloatBarrier

\backmatter

\section*{Declarations}

\bmhead{Acknowledgements}

This work was supported by The National Natural Science Foundation of China grant 32192410, 32192414, 32330043, STI2030-Major Projects (2021ZD0203900), The National Natural Science Foundation of China 82371522, The National Natural Science Foundation of China grant 32171025, ShanghaiTech AI4S Initiative SHTAI4S202404, National Key R\&D Program of China (2022YFF0902301), NSFC programs (61976138, 61977047), STCSM (2015F0203-000-06), and SHMEC(2019-01-07-00-01-E00003). We also acknowledge support from Shanghai Frontiers Science Center of Human-centered Artificial Intelligence (ShangHAI), Shanghai clinical research and trial center, Shanghai Frontiers Science Center for Bomacromolecules and Precision Medicine, HPC Platform of ShanghaiTech University, ShanghaiTech GenAI Project, The Library and Office of IT Services of ShanghaiTech University, and MoE Key Lab of Intelligent Perception and Human-Machine Collaboration (ShanghaiTech University).

\bmhead{Competing interests}
The authors declare no competing interests.

\bmhead{Ethics approval and consent to participate}
All experiments procedures were approved by the Institutional Animal Care and Use Committee (IACUC) of ShanghaiTech University.

\bmhead{Data Availability}
Example image--prompt--response pairs and a prepared analysis session containing behavioral descriptions and cached embeddings will be available at \url{https://huggingface.co/datasets/Ten1o/MouseGPT-examples}. The 435-frame benchmark images, expert-verified reference annotations and saved model scores will be released on Hugging Face, with links provided in the code repository. MouseGPT-Large, MouseGPT-Lite and kinematic-estimation model checkpoints and their loading instructions will be available at \url{https://huggingface.co/Ten1o/MouseGPT}. These resources will be made publicly accessible. Example data are licensed under CC BY-NC-ND 4.0; checkpoint-specific license terms will accompany the model releases. Full-resolution multiview recordings and associated frame-level data, including prompts, kinematic information and behavioral descriptions are available from the corresponding authors upon reasonable request because their large volume requires coordinated transfer.

\bmhead{Code Availability}
MouseGPT source code for preprocessing, training, inference and downstream analysis, together with installation and usage documentation and the benchmark evaluation, scoring and figure-generation code, will be available at \url{https://github.com/Ten1o/MouseGPT}. License terms and third-party notices are provided in the repository: original MouseGPT-authored material is distributed under CC BY-NC-ND 4.0, subject to component-specific exclusions and upstream licenses. We will maintain the software and provide user support for at least three years after publication; the published code and documentation will remain publicly accessible beyond this period. Installation, usage and compatibility questions and reproducible bug reports can be submitted at \url{https://github.com/Ten1o/MouseGPT/issues}.

\bmhead{Author Contribution}
J.Y. and J.H. conceived the project and T.X., Y.W., P.Y. designed the experiments. T.X., Y.W., and T.Z. designed and implemented the algorithm and trained the models. P.Y., S.T. collected and checked data and Z.T., X.C., H.L. assisted. K.S., T.Z. processed the raw videos and computed the coordinate of 3D skeleton. H.W. and X.W. support the psychedelics in this study. S.T., P.Y., and T.Z. performed the analysis for hallucination behavior. Y.L., Y.W., Z.T., and P.Y. evaluate the performance of different VLMs. Z.X., Y.L., and B.X. conducted evaluation and comparison studies. T.X., Y.W., P.Y., T.Z., K.S., S.T., and Y.L. participated in the writing of the corresponding content in the manuscript and J.W., J.H. and J.Y. revised the manuscript. J.H. and J.Y.supervised the project.

\clearpage
\section*{Supplementary Information}
\pdfbookmark[0]{Supplementary Information}{sec:supplementary}
\begingroup
\renewcommand{\figurename}{Supplementary Fig.}
\renewcommand{\tablename}{Supplementary Table}
\renewcommand{\thefigure}{\arabic{figure}}
\renewcommand{\thetable}{\arabic{table}}
\renewcommand{\theHfigure}{supplementary.\arabic{figure}}
\renewcommand{\theHtable}{supplementary.\arabic{table}}
\setcounter{figure}{0}
\setcounter{table}{0}
\begin{figure}[!htbp]
\setcounter{figure}{0}
\renewcommand{\thefigure}{\arabic{figure}}
\renewcommand{\theHfigure}{supplementary.\arabic{figure}}
\renewcommand{\figurename}{Supplementary Fig.}
\centering
\includegraphics[width=\textwidth,height=0.8\textheight,keepaspectratio]{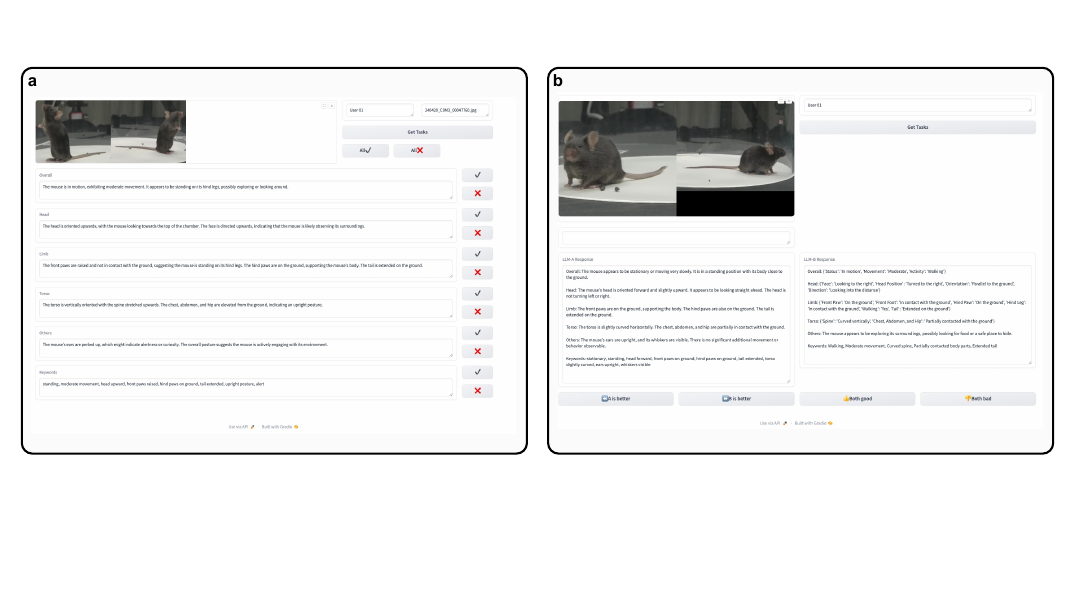}
\caption{\textbf{Expert evaluation interface for behavioral descriptions.} An interface designed for expert evaluation of MouseGPT's performance. \textbf{a}, Interface for human evaluation of description correctness: The interface allows experts to assess the correctness of model-generated behavioral descriptions. \textbf{b}, Interface for human comparison of descriptions from two models: Expert evaluation interface for paired comparisons between MouseGPT and other models.}
\label{fig:sf1}
\end{figure}
\clearpage

\begin{figure}[!htbp]
\setcounter{figure}{1}
\renewcommand{\thefigure}{\arabic{figure}}
\renewcommand{\theHfigure}{supplementary.\arabic{figure}}
\renewcommand{\figurename}{Supplementary Fig.}
\centering
\includegraphics[width=\textwidth,height=0.8\textheight,keepaspectratio]{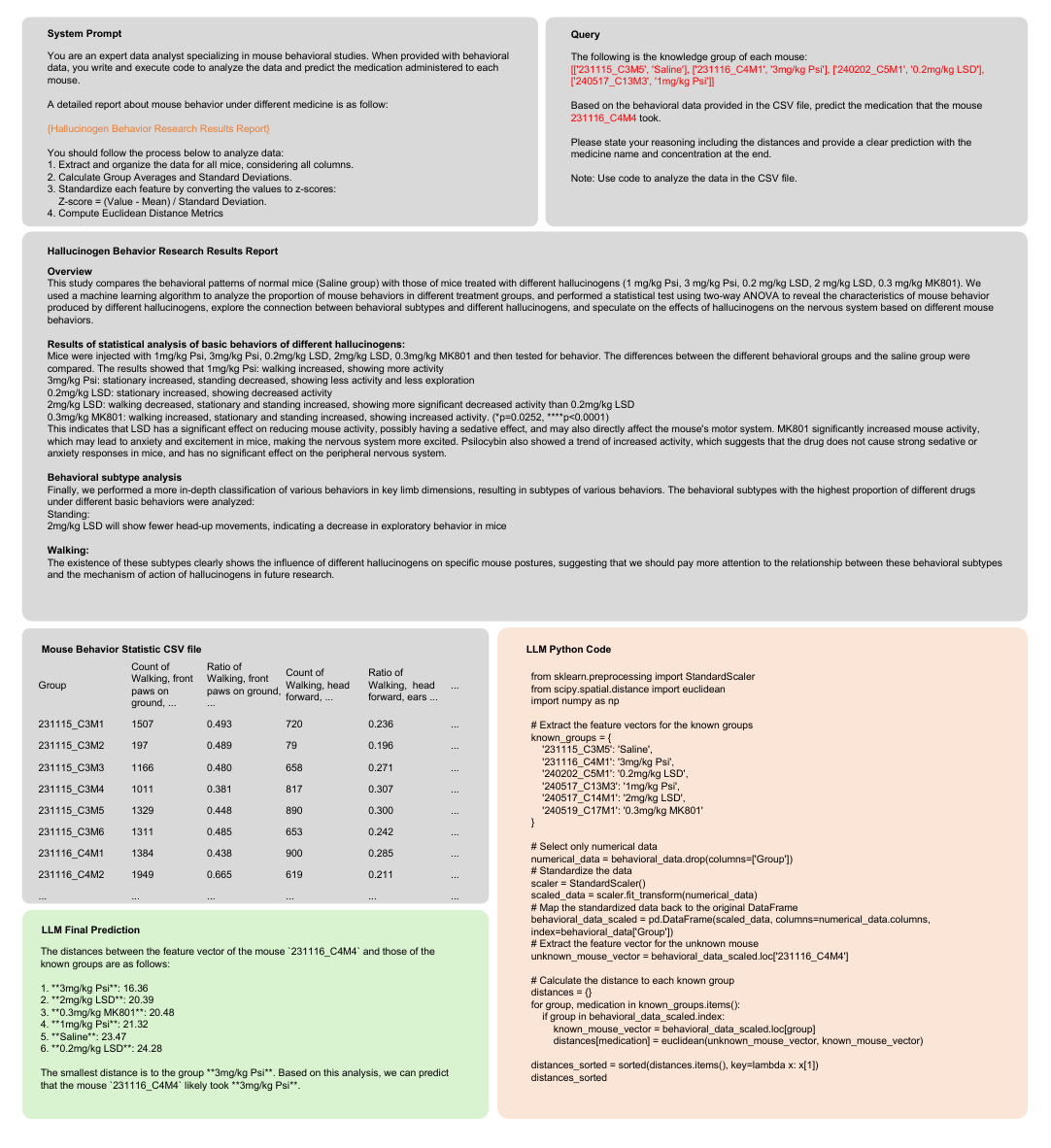}
\caption{\textbf{Interactive workflow for drug treatment prediction from behavioral phenotypes.} An example of the step-by-step process used by MouseGPT to predict the drug treatment administered to a mouse from behavioral phenotype data. The procedure involves standardizing behavioral data, calculating Euclidean distances between feature vectors and identifying the closest match among known treatment groups. The workflow combines computational tools and statistical analysis to generate a prediction of drug treatment.}
\label{fig:sf2}
\end{figure}
\clearpage

\begin{figure}[!htbp]
\setcounter{figure}{2}
\renewcommand{\thefigure}{\arabic{figure}}
\renewcommand{\theHfigure}{supplementary.\arabic{figure}}
\renewcommand{\figurename}{Supplementary Fig.}
\centering
\includegraphics[width=\textwidth,height=0.8\textheight,keepaspectratio]{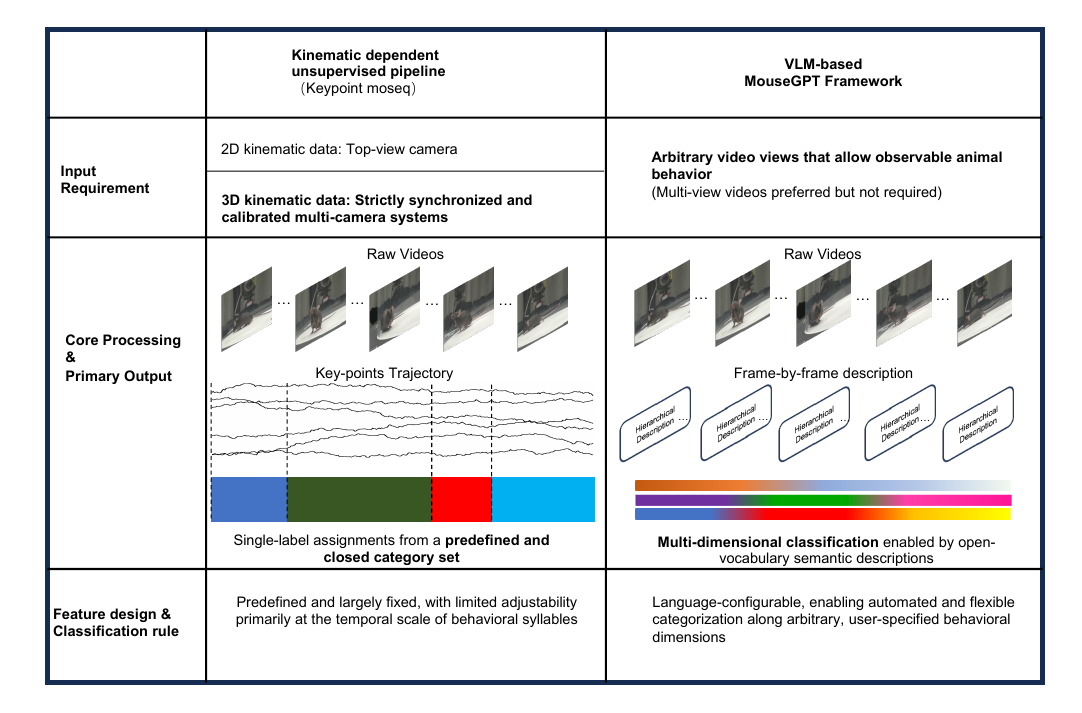}
\caption{\textbf{Overview of processing pipelines in MouseGPT and conventional kinematics-based unsupervised methods.} Schematic comparison of input requirements, intermediate representations and primary outputs. Keypoint-MoSeq operates on kinematic trajectories and produces behavioral syllables that have not yet been assigned semantic names and require subsequent semantic interpretation. MouseGPT operates directly on raw videos (multiview input is optional) and generates open-vocabulary semantic descriptions or labels, supporting flexible, language-configurable behavioral categorization.}
\label{fig:sf3}
\end{figure}
\clearpage

\begin{figure}[!htbp]
\setcounter{figure}{3}
\renewcommand{\thefigure}{\arabic{figure}}
\renewcommand{\theHfigure}{supplementary.\arabic{figure}}
\renewcommand{\figurename}{Supplementary Fig.}
\centering
\includegraphics[width=\textwidth,height=0.8\textheight,keepaspectratio]{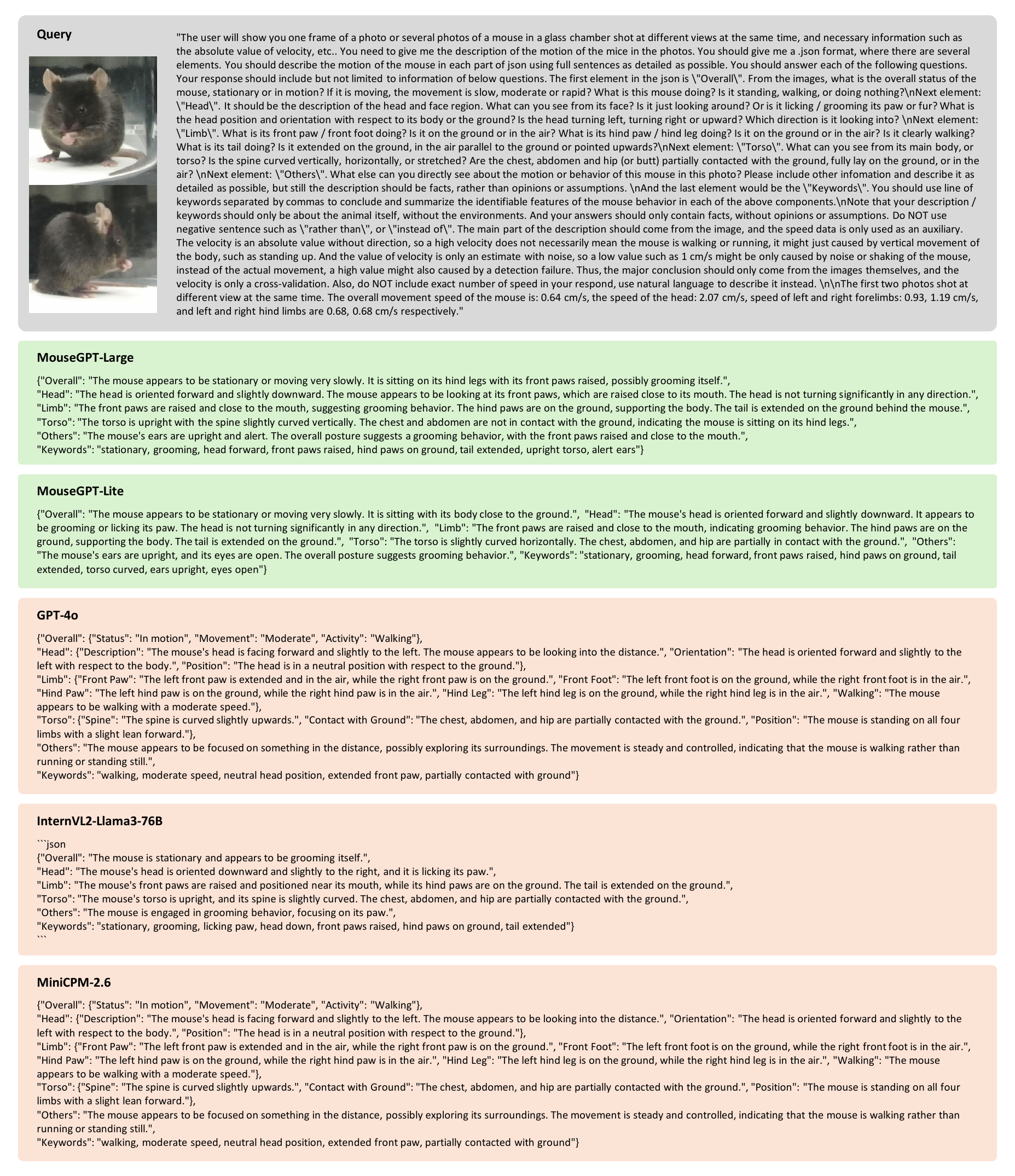}
\caption{\textbf{Comparison of MouseGPT Behavior Understanding Model and Other Vision-Language Models.} A comparative result of behavior descriptions generated by MouseGPT-Large, MouseGPT-Lite, GPT-4o, InternVL2-Llama3-76B, and MiniCPM-2.6. The figure demonstrates the differences in descriptive detail, accuracy, and interpretability across models for key behavioral components such as overall activity, head orientation, limb positioning, and torso posture.}
\label{fig:sf4}
\end{figure}
\clearpage

\begin{figure}[!htbp]
\setcounter{figure}{4}
\renewcommand{\thefigure}{\arabic{figure}}
\renewcommand{\theHfigure}{supplementary.\arabic{figure}}
\renewcommand{\figurename}{Supplementary Fig.}
\centering
\includegraphics[width=\textwidth,height=0.8\textheight,keepaspectratio]{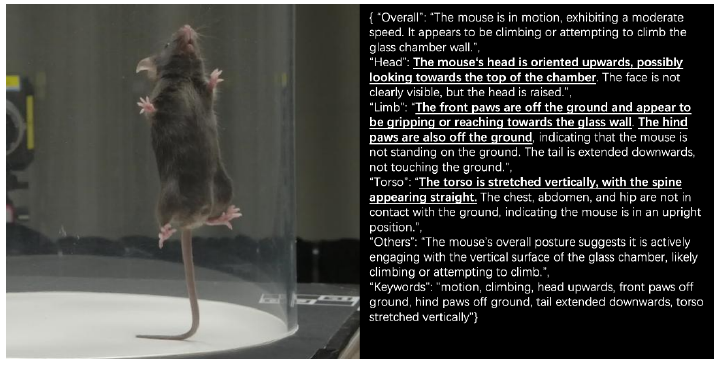}
\caption{\textbf{Zero-shot rare-event example: MouseGPT surfaces an unusual frame as a semantic outlier.} The model successfully flags a rare event (wall-climbing) in the continuous embedding space and generates a structured open-vocabulary, body-part-resolved description without relying on predefined labels (verbatim output shown).}
\label{fig:sf5}
\end{figure}
\clearpage

\begin{figure}[!htbp]
\setcounter{figure}{5}
\renewcommand{\thefigure}{\arabic{figure}}
\renewcommand{\theHfigure}{supplementary.\arabic{figure}}
\renewcommand{\figurename}{Supplementary Fig.}
\centering
\includegraphics[width=\textwidth,height=0.8\textheight,keepaspectratio]{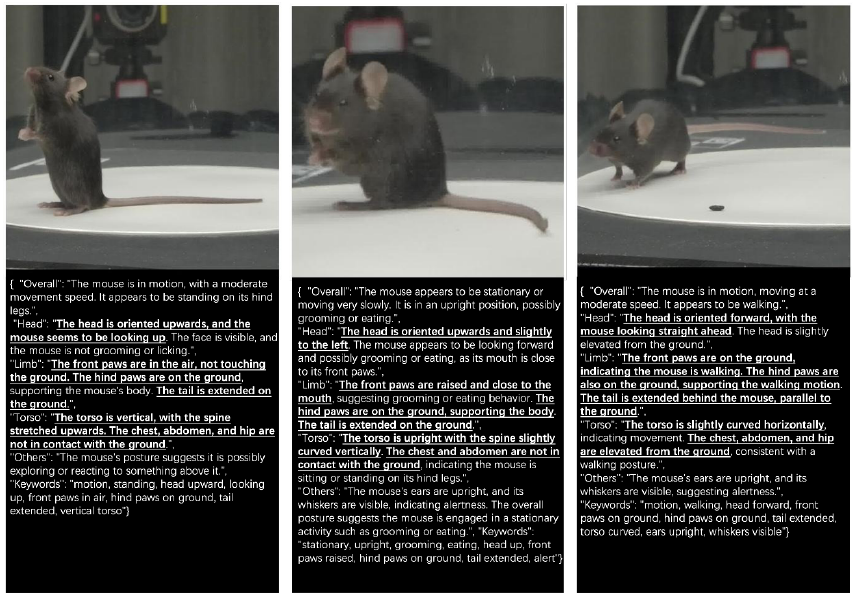}
\caption{\textbf{Single-camera inference example of MouseGPT-Lite.} Representative single-camera frames paired with verbatim MouseGPT-Lite structured descriptions, including head orientation, paw contact state and torso posture. For this qualitative demonstration, one side-view image was supplied as visual input together with the corresponding kinematic summary used by the standard inference prompt. No additional fine-tuning, retraining or architecture modification was introduced.}
\label{fig:sf6}
\end{figure}
\clearpage

\clearpage
\begingroup
\renewcommand{\tablename}{Supplementary Table}
\renewcommand{\thetable}{\arabic{table}}
\renewcommand{\theHtable}{supplementary.\arabic{table}}
\setcounter{table}{0}
\newlength{\MouseGPTTableWidth}
\setlength{\LTleft}{0pt}
\setlength{\LTright}{0pt}
\setlength{\LTpre}{8pt}
\setlength{\LTpost}{8pt}
\makeatletter
\long\def\LT@makecaption#1#2#3{%
  \LT@mcol\LT@cols c{%
    \hbox to\z@{\hss\parbox[t]{\LTcapwidth}{%
      \@tablecaption{#1{#2}}{#3}\vspace{6pt}%
    }\hss}%
  }%
}
\makeatother

\clearpage
\begingroup
\tablebodyfont
\setlength{\tabcolsep}{4pt}
\renewcommand{\arraystretch}{1.12}
\setlength{\LTcapwidth}{\linewidth}
\setlength{\MouseGPTTableWidth}{\dimexpr\linewidth-4\tabcolsep\relax}
\begin{longtable}{@{}>{\centering\arraybackslash}p{0.1\MouseGPTTableWidth}>{\raggedright\arraybackslash}p{0.24\MouseGPTTableWidth}>{\raggedright\arraybackslash}p{0.66\MouseGPTTableWidth}@{}}
\caption{\textbf{Mouse body keypoint labels and anatomical definitions.}}\label{tab:st1}\\
\toprule
\textbf{Index} & \textbf{Label} & \textbf{Detailed Description} \\
\midrule
\endfirsthead
\caption[]{\textbf{Mouse body keypoint labels and anatomical definitions.} (continued)}\\
\toprule
\textbf{Index} & \textbf{Label} & \textbf{Detailed Description} \\
\midrule
\endhead
\midrule
\multicolumn{3}{r}{Continued on next page}\\
\endfoot
\bottomrule
\endlastfoot
1 & EarL & Tip of the Left Ear \\
2 & EarR & Tip of the Right Ear \\
3 & Snout & Tip of the Snout \\
4 & SpineF & Beginning of the Spine \\
5 & SpineG & First Tripartite Point of the Spine \\
6 & SpineH & Second Tripartite Point of the Spine \\
7 & Hip & Hip Joint \\
8 & Tail (base) & Base of the Tail \\
9 & Tail (mid) & Middle Section of the Tail \\
10 & Tail (end) & Tip of the Tail \\
11 & ForepawL & Tip of the Left Forepaw \\
12 & WristL & Left Wrist Joint \\
13 & ElbowL & Left Elbow Joint \\
14 & ShoulderL & Left Shoulder Joint \\
15 & ForepawR & Tip of the Right Forepaw \\
16 & WristR & Right Wrist Joint \\
17 & ElbowR & Right Elbow Joint \\
18 & ShoulderR & Right Shoulder Joint \\
19 & HindpawL & Tip of the Left Hindpaw \\
20 & AnkleL & Left Ankle Joint \\
21 & KneeL & Left Knee Joint \\
22 & HindpawR & Tip of the Right Hindpaw \\
23 & AnkleR & Right Ankle Joint \\
24 & KneeR & Right Knee Joint \\
\end{longtable}
\endgroup

\clearpage
\begingroup
\setlength{\topmargin}{\dimexpr\paperheight/2-\textheight/2-1in-\headheight-\headsep\relax}
\begin{landscape}
\makeatletter
\let\@oddhead\@empty\let\@evenhead\@empty
\makeatother
\begingroup
\tablebodyfont
\setlength{\tabcolsep}{3pt}
\renewcommand{\arraystretch}{1.12}
\setlength{\LTcapwidth}{\linewidth}
\setlength{\MouseGPTTableWidth}{\dimexpr\linewidth-14\tabcolsep\relax}
\begin{longtable}{@{}>{\raggedright\arraybackslash}p{0.16\MouseGPTTableWidth}>{\raggedright\arraybackslash}p{0.12\MouseGPTTableWidth}>{\raggedright\arraybackslash}p{0.12\MouseGPTTableWidth}>{\raggedright\arraybackslash}p{0.12\MouseGPTTableWidth}>{\raggedright\arraybackslash}p{0.12\MouseGPTTableWidth}>{\raggedright\arraybackslash}p{0.12\MouseGPTTableWidth}>{\raggedright\arraybackslash}p{0.12\MouseGPTTableWidth}>{\raggedright\arraybackslash}p{0.12\MouseGPTTableWidth}@{}}
\caption{\textbf{Experimental settings and results for the MABE 2022 benchmark.}}\label{tab:st2}\\
\toprule
\textbf{Metric / Setup} & \textbf{Our framework\newline w/ GPT-6 Astra} & \textbf{Our framework\newline w/ GPT-5.6-Luna} & \textbf{Our framework\newline w/ GPT-5.1} & \textbf{Our framework\newline w/ Doubao-1.6-lite} & \textbf{AmadeusGPT} & \textbf{BAMS} & \textbf{PCA baseline} \\
\midrule
\endfirsthead
\caption[]{\textbf{Experimental settings and results for the MABE 2022 benchmark.} (continued)}\\
\toprule
\textbf{Metric / Setup} & \textbf{Our framework\newline w/ GPT-6 Astra} & \textbf{Our framework\newline w/ GPT-5.6-Luna} & \textbf{Our framework\newline w/ GPT-5.1} & \textbf{Our framework\newline w/ Doubao-1.6-lite} & \textbf{AmadeusGPT} & \textbf{BAMS} & \textbf{PCA baseline} \\
\midrule
\endhead
\midrule
\multicolumn{8}{r}{Continued on next page}\\
\endfoot
\bottomrule
\endlastfoot
Model type & Generalist VLM perception module & Generalist VLM perception module & Generalist VLM perception module & Generalist VLM perception module & Language-guided, rule-based task programming & Learned behavioral representation & PCA representation baseline \\
\addlinespace[3pt]
Data coverage & Same one-fifth subset of MABe 2022 & Full local evaluation condition & Same one-fifth subset of MABe 2022 & Full local evaluation condition & Published MABe Round 1 evaluation split & Published evaluation split & Published evaluation split \\
\addlinespace[3pt]
Train / test split & Custom source-video-level train/test split & Shared custom 80/20 video-level split & Custom source-video-level train/test split & Shared custom 80/20 video-level split & As reported in the reference & As reported in the reference & As reported in the reference \\
\addlinespace[3pt]
Input / clip setting & Video clips; archived segment lengths: 15--1,800 frames & Video clips; archived segment lengths: 15--1,800 frames & Video clips; archived segment lengths: 15--1,800 frames & Video clips; archived segment lengths: 15--1,800 frames & Pre-computed keypoints; natural-language behavior definitions & Keypoint trajectories & Keypoint trajectories \\
\addlinespace[3pt]
Representation for scoring & Description JSON \(\rightarrow\) text-embedding-3-large \(\rightarrow\) PCA (128D) & Description JSON \(\rightarrow\) text-embedding-3-large \(\rightarrow\) PCA (128D) & Description JSON \(\rightarrow\) text-embedding-3-large \(\rightarrow\) PCA (128D) & Description JSON \(\rightarrow\) text-embedding-3-large \(\rightarrow\) PCA (128D) & Rule-based behavior events generated by task programs & Learned representations, as reported & PCA features, as reported \\
\addlinespace[3pt]
Downstream evaluation & Per-behavior Ridge classification; archived frame-label scoring & Per-behavior Ridge classification; archived frame-label scoring & Per-behavior Ridge classification; archived frame-label scoring & Per-behavior Ridge classification; archived frame-label scoring & Rule-based detection; F1 reported in Table 1 & Benchmark evaluation reported in Table 1 & Benchmark evaluation reported in Table 1 \\
\addlinespace[3pt]
\noalign{\penalty -10000}
Training videos & 1,057 & 1,464 & 1,066 & 1,464 & n.r. & n.r. & n.r. \\
\addlinespace[3pt]
Test videos & 275 & 366 & 266 & 366 & n.r. & n.r. & n.r. \\
\addlinespace[3pt]
Training clips & 2,970 & 26,976 & 3,096 & 26,988 & n.a. & n.r. & n.r. \\
\addlinespace[3pt]
Test clips & 816 & 7,043 & 701 & 7,043 & n.r. & n.r. & n.r. \\
\addlinespace[3pt]
Avg F1 & 0.5600 & 0.4540 & 0.5486 & 0.1652 & 0.2960 & 0.1964 & 0.0145 \\
\addlinespace[3pt]
Result source & Current archived evaluation & Current archived evaluation & Current archived evaluation & Current archived evaluation & Published reference~\cite{ye2023amadeusGPT} & Published reference~\cite{ye2023amadeusGPT} & Published reference~\cite{ye2023amadeusGPT} \\
\end{longtable}
\noindent Avg F1 is the arithmetic mean of the positive-class F1 scores for the nine behaviors in the final figure.\par\smallskip
\noindent Video and clip counts reflect actual archived coverage. n.r., not reported in the extracted reference; n.a., not applicable.\par\smallskip
\noindent Published reference results were taken from Table 1 of Ye et al.~\cite{ye2023amadeusGPT}. Data coverage and per-behavior scores are provided in the following table blocks.\par\smallskip
\clearpage
\makeatletter
\noindent\@tablecaption{Supplementary Table 2}{\textbf{Experimental settings and results for the MABE 2022 benchmark.} (continued)}\par\medskip
\makeatother
\noindent\textbf{Archived data coverage}\par\smallskip
\setlength{\MouseGPTTableWidth}{\dimexpr\linewidth-16\tabcolsep\relax}
\noindent\begin{tabular}{@{}>{\raggedright\arraybackslash}p{0.18\MouseGPTTableWidth}>{\centering\arraybackslash}p{0.1025\MouseGPTTableWidth}>{\centering\arraybackslash}p{0.1025\MouseGPTTableWidth}>{\centering\arraybackslash}p{0.1025\MouseGPTTableWidth}>{\centering\arraybackslash}p{0.1025\MouseGPTTableWidth}>{\centering\arraybackslash}p{0.1025\MouseGPTTableWidth}>{\centering\arraybackslash}p{0.1025\MouseGPTTableWidth}>{\centering\arraybackslash}p{0.1025\MouseGPTTableWidth}>{\centering\arraybackslash}p{0.1025\MouseGPTTableWidth}@{}}
\toprule
\textbf{Method} & \textbf{Split-list train videos} & \textbf{Split-list test videos} & \textbf{Actual train videos} & \textbf{Actual test videos} & \textbf{Train clips} & \textbf{Test clips} & \textbf{Train frames} & \textbf{Test frames} \\
\midrule
GPT-6 Astra & 1,464 & 366 & 1,057 & 275 & 2,970 & 816 & 334,494 & 87,340 \\
GPT-5.6-Luna & 1,464 & 366 & 1,464 & 366 & 26,976 & 7,043 & 2,634,231 & 658,800 \\
GPT-5.1 & 1,066 & 266 & 1,066 & 266 & 3,096 & 701 & 336,781 & 86,327 \\
Doubao-1.6-lite & 1,464 & 366 & 1,464 & 366 & 26,988 & 7,043 & 2,635,200 & 658,800 \\
\bottomrule
\end{tabular}\par\bigskip
\noindent\textbf{Positive-class F1 by behavior}\par\smallskip
\setlength{\MouseGPTTableWidth}{\dimexpr\linewidth-20\tabcolsep\relax}
\noindent\begin{tabular}{@{}>{\raggedright\arraybackslash}p{0.157\MouseGPTTableWidth}>{\centering\arraybackslash}p{0.082\MouseGPTTableWidth}>{\centering\arraybackslash}p{0.082\MouseGPTTableWidth}>{\centering\arraybackslash}p{0.105\MouseGPTTableWidth}>{\centering\arraybackslash}p{0.082\MouseGPTTableWidth}>{\centering\arraybackslash}p{0.082\MouseGPTTableWidth}>{\centering\arraybackslash}p{0.082\MouseGPTTableWidth}>{\centering\arraybackslash}p{0.082\MouseGPTTableWidth}>{\centering\arraybackslash}p{0.082\MouseGPTTableWidth}>{\centering\arraybackslash}p{0.082\MouseGPTTableWidth}>{\centering\arraybackslash}p{0.082\MouseGPTTableWidth}@{}}
\toprule
\textbf{Method} & \textbf{Close} & \textbf{Contact} & \textbf{Approaches} & \textbf{Watching} & \textbf{Huddles} & \textbf{Chases} & \textbf{Oral-genital contact} & \textbf{Oral-oral contact} & \textbf{Oral-ear contact} & \textbf{Mean F1} \\
\midrule
GPT-6 Astra & 0.9164 & 0.9347 & 0.9435 & 0.9913 & 0.7535 & 0.3184 & 0.0908 & 0.0764 & 0.0147 & 0.5600 \\
GPT-5.6-Luna & 0.9374 & 0.9264 & 0.5968 & 0.4005 & 0.5175 & 0.5806 & 0.0462 & 0.0618 & 0.0186 & 0.4540 \\
GPT-5.1 & 0.9161 & 0.9433 & 0.9590 & 0.9593 & 0.8340 & 0.3262 & 0.0000 & 0.0000 & 0.0000 & 0.5486 \\
Doubao-1.6-lite & 0.6668 & 0.5064 & 0.0194 & 0.0919 & 0.1428 & 0.0093 & 0.0350 & 0.0138 & 0.0015 & 0.1652 \\
AmadeusGPT & 0.7000 & 0.5720 & 0.0140 & 0.6000 & 0.3800 & 0.2740 & 0.0500 & 0.0240 & 0.0500 & 0.2960 \\
BAMS & 0.6640 & 0.5330 & 0.0200 & 0.1910 & 0.3020 & 0.0230 & 0.0165 & 0.0140 & 0.0045 & 0.1964 \\
PCA baseline & 0.1300 & 0.0008 & 0.0000 & 0.0000 & 0.0000 & 0.0000 & 0.0000 & 0.0000 & 0.0000 & 0.0145 \\
\bottomrule
\end{tabular}\par\medskip
\endgroup
\end{landscape}
\endgroup

\clearpage
\begingroup
\tablebodyfont
\setlength{\tabcolsep}{4pt}
\renewcommand{\arraystretch}{1.16}
\setlength{\LTcapwidth}{\linewidth}
\setlength{\MouseGPTTableWidth}{\dimexpr\linewidth-4\tabcolsep\relax}
\begin{longtable}{@{}>{\raggedright\arraybackslash}p{0.23\MouseGPTTableWidth}>{\raggedright\arraybackslash}p{0.105\MouseGPTTableWidth}>{\raggedright\arraybackslash}p{0.665\MouseGPTTableWidth}@{}}
\caption{\textbf{Tasks and full prompts for the MouseGPT agent benchmark.}}\label{tab:st3}\\
\toprule
\textbf{Category} & \textbf{Task ID} & \textbf{Prompt (verbatim)} \\
\midrule
\endfirsthead
\caption[]{\textbf{Tasks and full prompts for the MouseGPT agent benchmark.} (continued)}\\
\toprule
\textbf{Category} & \textbf{Task ID} & \textbf{Prompt (verbatim)} \\
\midrule
\endhead
\midrule
\multicolumn{3}{r}{Continued on next page}\\
\endfoot
\bottomrule
\endlastfoot
Basic Behavior Query & Task 1 & “What behavior is the mouse performing in the current frame? Please explain the reasoning.” \\
\addlinespace[2pt]
Basic Behavior Query & Task 2 & “Show me the five frames that are most similar to the current frame’s posture.” \\
\addlinespace[2pt]
Behavior Classification & Task 3 & “Please classify the mouse behavior across the entire video and summarize the percentage of each class.” \\
\addlinespace[2pt]
Behavior Classification & Task 4 & “Generate a detailed ethogram for the full recording and save it as an output file.” \\
\addlinespace[2pt]
Behavior Clustering & Task 5 & “Cluster all mouse behaviors.” \\
\addlinespace[2pt]
Behavior Clustering & Task 6 & “Compare cluster 2 with cluster 5 and describe how their postures differ.” \\
\addlinespace[2pt]
Behavior Search & Task 7 & “Search for all behavior instances similar to ‘sniffing’ and return the top five matching frames.” \\
\addlinespace[2pt]
Behavior Search & Task 8 & “Find frames where the mouse appears to stretch its body upward.” \\
\addlinespace[2pt]
Novel Behavior\newline \& Outlier Detection & Task 9 & “Detect all outlier behaviors and list the frame IDs where unusual postures appear.” \\
\addlinespace[2pt]
Novel Behavior\newline \& Outlier Detection & Task 10 & “Are there any behaviors in this video that the model has never seen before? Please identify them and describe the top 5 frames.” \\
\addlinespace[2pt]
Behavior Transition\newline \& Statistics & Task 11 & “Calculate how many times the mouse transitions from ‘walking’ to ‘rearing’.” \\
\addlinespace[2pt]
Behavior Transition\newline \& Statistics & Task 12 & “Generate a behavior transition matrix for the entire video and save it as a CSV file.” \\
\addlinespace[2pt]
Multi-behavior\newline Comparative Analysis & Task 13 & “Compare the frequency of crouching versus grooming across the full recording.” \\
\addlinespace[2pt]
Multi-behavior\newline Comparative Analysis & Task 14 & “Analyze how the mouse’s behavior distribution changes over time (early/middle/late).” \\
\addlinespace[2pt]
Frame-Based\newline Trend Prediction & Task 15 & “Based on the nearby frames, what do you predict the mouse will do next?” \\
\addlinespace[2pt]
Frame-Based\newline Trend Prediction & Task 16 & “Based on the last 30 frames, predict the mouse’s behavior in the next 10 frames.” \\
\addlinespace[2pt]
Code Generation\newline \& Debugging & Task 17 & “Load the cluster.csv file, print the column names, and calculate the average duration of each behavior.” \\
\addlinespace[2pt]
Code Generation\newline \& Debugging & Task 18 & “Draw the frequency of each behavior over time and save the picture but don’t show me in UI. If any errors occur, correct them automatically.” \\
\addlinespace[2pt]
Comprehensive\newline Multi-step Tasks & Task 19 & “Find all frames similar to ‘looking upward’, save results as search\_looking\_upward.csv calculate the overall frequency of this behavior, and analyze how it changes every 5 minutes.” \\
\addlinespace[2pt]
Comprehensive\newline Multi-step Tasks & Task 20 & “Summarize the key behavioral patterns of this mouse and highlight anything unusual or noteworthy.” \\
\end{longtable}
\endgroup

\endgroup

\endgroup

\end{document}